\documentclass[a4paper,fleqn]{cas-sc}
\usepackage{threeparttable}
\usepackage{bbding}
\usepackage{color}
\usepackage{multirow}
\makeatletter
\renewcommand{\fnum@figure}{Fig. \thefigure.\@gobble}
\makeatother

\usepackage[numbers]{natbib}

\def\tsc#1{\csdef{#1}{\textsc{\lowercase{#1}}\xspace}}
\tsc{WGM}
\tsc{QE}

\begin{document}
\let\WriteBookmarks\relax
\def\floatpagepagefraction{1}
\def\textpagefraction{.001}

\shorttitle{A Comprehensive Survey of Large Language Models and Multimodal Large Language Models in Medicine}   

\shortauthors{Xiao et al.}

\title[mode = title]{A Comprehensive Survey of Large Language Models and Multimodal Large Language Models in Medicine}  

\author[1]{Hanguang\ Xiao}[orcid=0000-0002-4359-7455]
\cormark[1] 
\fnmark[1] 
\ead{simenxiao1211@163.com} 

\author[1]{Feizhong\ Zhou}[style=chinese]
\fnmark[1] 

\author[1]{Xingyue\ Liu}
\author[1]{Tianqi\ Liu}
\author[1]{Zhipeng\ Li}
\author[1]{Xin\ Liu}
\author[1]{Xiaoxuan\ Huang}

\address[1]{School of Artificial Intelligence, Chongqing University of Technology, Chongqing 401120, China}

\cortext[1]{Corresponding author} 

\begin{abstract}
Since the release of ChatGPT and GPT-4, large language models (LLMs) and multimodal large language models (MLLMs) have attracted widespread attention for their exceptional capabilities in understanding, reasoning, and generation, introducing transformative paradigms for integrating artificial intelligence into medicine. This survey provides a comprehensive overview of the development, principles, application scenarios, challenges, and future directions of LLMs and MLLMs in medicine. Specifically, it begins by examining the paradigm shift, tracing the transition from traditional models to LLMs and MLLMs, and highlighting the unique advantages of these LLMs and MLLMs in medical applications. Next, the survey reviews existing medical LLMs and MLLMs, providing detailed guidance on their construction and evaluation in a clear and systematic manner. Subsequently, to underscore the substantial value of LLMs and MLLMs in healthcare, the survey explores five promising applications in the field. Finally, the survey addresses the challenges confronting medical LLMs and MLLMs and proposes practical strategies and future directions for their integration into medicine. In summary, this survey offers a comprehensive analysis of the technical methodologies and practical clinical applications of medical LLMs and MLLMs, with the goal of bridging the gap between these advanced technologies and clinical practice, thereby fostering the evolution of the next generation of intelligent healthcare systems.
\end{abstract}




\maketitle

\section{Introduction}
\label{sec1}
The introduction of the Transformer~\cite{vaswani2017attention} has revolutionized the fields of Natural Language Processing (NLP) and Computer Vision (CV). The Transformer’s robust parallel computing capabilities and self-attention mechanism facilitate the integration of vast training datasets, forming the foundation for LLM and MLLM development. A variety of Transformer-based LLMs and MLLMs have emerged to date, with this survey primarily focusing on the vision-language modality. Notable examples include the PaLM series~\cite{chowdhery2023palm,anil2023palm}, GPT series \cite{brown2020language,ouyang2022training}, and LLaMA series \cite{touvron2023llama,touvron2023llama2, dubey2024llama} among LLMs, and Gemini \cite{team2023gemini}, GPT-4 \cite{achiam2023gpt}, and LLaVA~\cite{liu2024visual} among MLLMs. Their exceptional capabilities in understanding, reasoning, and generation have enabled them to achieve state-of-the-art performance across various downstream tasks, including text generation, machine translation, and visual question answering (VQA). LLMs and MLLMs exhibit increasingly robust generalization abilities, significantly impacting the medical domain and accelerating the convergence of artificial intelligence and medicine~\cite{thirunavukarasu2023large}. Notably, Google's Med-PaLM 2 \cite{singhal2023towards} scored 86.5 on the United States Medical Licensing Examination (USMLE) \cite{jin2021disease}, achieving expert-level performance \cite{zhou2023survey} and further highlighting the immense potential of LLMs in medicine. Additionally, emerging medical LLMs and MLLMs, such as ChatDoctor~\cite{li2023chatdoctor}, ChatCAD~\cite{wang2024interactive}, and LLaVA-Med \cite{li2024llava}, represent novel opportunities enabled by artificial intelligence in the medical field. These models offer promising solutions for medical report generation~\cite{van2023clinical,wang2023r2gengpt}, clinical diagnosis \cite{wang2024interactive,tu2024towards}, mental health services \cite{chen2023soulchat,liu2023chatcounselor}, and various other clinical applications.

\begin{figure}[ht]
  \centering
  \includegraphics[width=\linewidth]{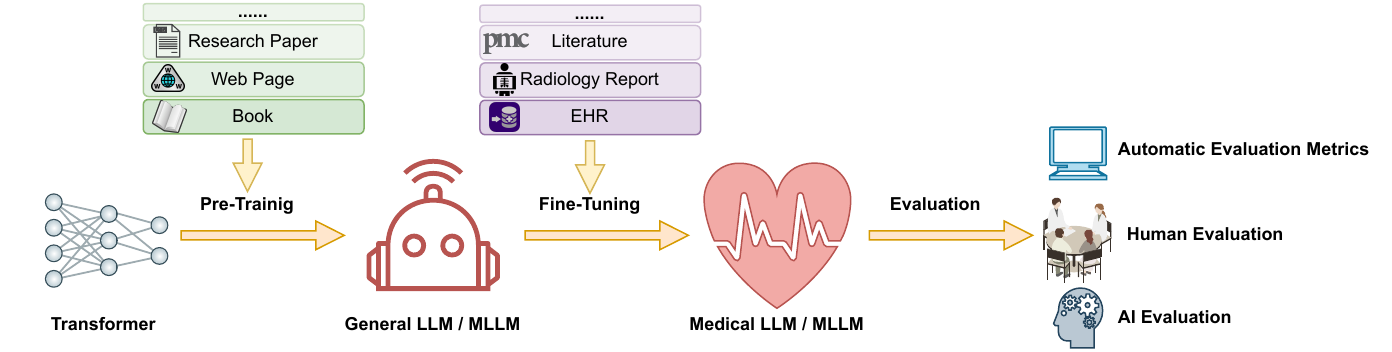}
  \caption{The process of constructing and evaluating medical LLMs and MLLMs.}
  \label{fig1}
\end{figure}

Despite the significant academic breakthroughs of LLMs and MLLMs in the medical field, hospitals still face notable challenges in training their own medical LLMs and MLLMs and deploying them in practical clinical applications. First, training medical LLMs and MLLMs requires a substantial amount of medical data, which is costly to obtain, demands annotation by medical experts, and raises significant concerns about data privacy~\cite{zhang2024data}. Second, the parameters and computational demands of LLMs and MLLMs are substantial, necessitating significant computational resources for their training and deployment~\cite{moor2023foundation,qiu2023large}, substantially increasing the adoption barrier for hospitals. Third, given the unique characteristics of the medical field, it is essential not only to evaluate the performance of LLMs and MLLMs on benchmarks but also to assess their instruction-following ability~\cite{zhang2023instruction,ouyang2022training,liu2024visual}, safety, and ethical considerations, necessitating additional training and evaluation strategies to enhance and measure the models' performance across multiple dimensions. Furthermore, the development of LLMs and MLLMs in the medical field is still in its infancy, with many of their potential application scenarios remaining undefined. Moreover, they face a range of challenges, including hallucinations \cite{umapathi2023med,rawte2023survey,ji2023survey} and a lack of up-to-date information \cite{thirunavukarasu2023large}, which significantly impede their practical use in clinical settings. 

To address the aforementioned challenges, this survey begins by tracing the evolution of LLMs and MLLMs through the lens of paradigm shifts. Subsequently, it reviews existing medical LLMs and MLLMs, summarizing their structural characteristics. The survey then gathers datasets suitable for training medical LLMs and MLLMs and elaborates on methods for training and evaluating these models, as shown in Fig.~\ref{fig1}. Furthermore, to highlight the significant potential impact of LLMs and MLLMs in medicine, this survey summarizes their applications in clinical practice and analyzes current limitations and potential solutions. Finally, the survey explores the future directions of medical LLMs and MLLMs, offering forward-looking and insightful perspectives.

Medicine is a multimodal field~\cite{LIN2024102795,KRONES2025102690}, making the study of medical MLLMs particularly important, as they can integrate and analyze information from various modalities to enhance clinical decision support, disease diagnosis, and treatment planning. However, the articles relevant to this survey mainly focus on medical LLMs and lack a detailed examination of medical MLLMs \cite{zhou2023survey,he2023survey,wang2023pre}. Additionally, most articles focus on the applications and impacts of LLMs in medicine but lack detailed discussions of technical aspects \cite{thirunavukarasu2023large,thapa2023chatgpt,qiu2023large,omiye2024large,bhayana2024chatbots}, such as datasets, model architectures, and construction methods. In contrast, this survey not only examines the background and principles of LLMs and MLLMs but also explores their applications and impacts in medicine, offering a clear logical structure and substantial depth and breadth. In summary, the contributions of this survey are as follows:

\begin{itemize}
\item This survey offers a thorough overview of medical LLMs and MLLMs, starting with an examination of their developmental background and architectural frameworks.	Building on this foundation, it catalogs existing medical LLMs and MLLMs while providing a detailed analysis of their structural variations and key components.		

\item This survey systematically elucidates the complete process of medical LLMs and MLLMs, from training to evaluation, covering fine-tuning methods, evaluation strategies, and relevant medical datasets. Additionally, it highlights how to select appropriate datasets, fine-tuning methods, and evaluation strategies to assist researchers in the rapid development of medical LLMs and MLLMs.

\item This survey summarizes the applications, challenges, and potential solutions of medical LLMs and MLLMs in clinical practice, while providing a forward-looking analysis of future developmental trajectories. It seeks to offer visionary perspectives that inspire advancements in the field, benefiting medical professionals and researchers alike.
\end{itemize}

\begin{figure}
  \centering
  \includegraphics[scale=0.5]{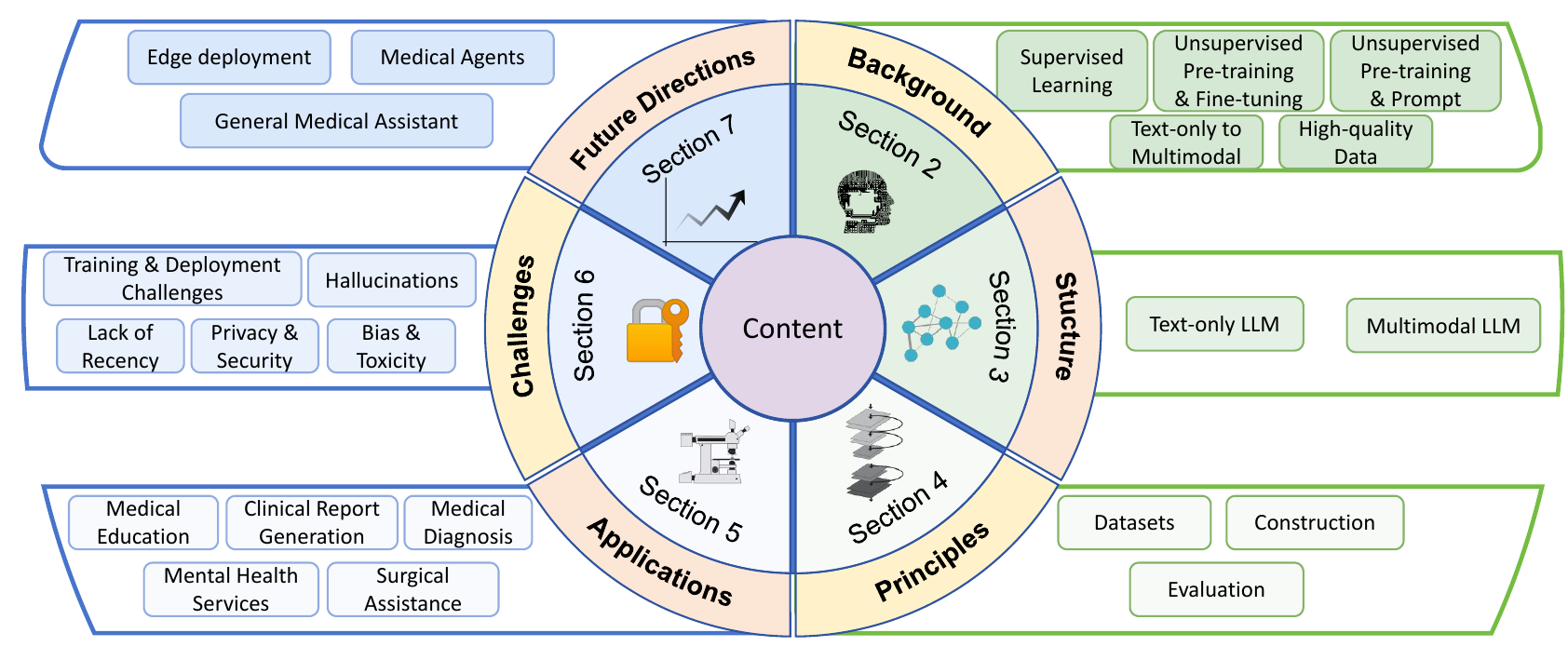}
  \caption{The overall structure of the survey. Section~\ref{sec2} to Section~\ref{sec4} are biased toward principles of medical LLMs and MLLMs; Section~\ref{Sec5} to Section~\ref{Sec7} are biased toward practical clinical applications.}
  \label{fig2}
\end{figure}

This survey aims to advance the development of LLMs and MLLMs for clinical medicine applications, thereby promoting deeper integration between artificial intelligence and healthcare. The structure of this survey is outlined in Fig.~\ref{fig2}: Section~\ref{sec2} reviews the development background of LLMs and MLLMs. Section~\ref{sec3} describes the architectures of current LLMs and MLLMs and highlights the structural differences among them. Section~\ref{sec4} covers the datasets for training medical LLMs and MLLMs, along with an overview of training and evaluation methods. Section~\ref{Sec5} examines current potential applications of medical LLMs and MLLMs. Section~\ref{Sec6} discusses the challenges and limitations of LLMs and MLLMs in clinical settings and proposes potential solutions. Section~\ref{Sec7} offers a forward-looking perspective on the future of medical LLMs and MLLMs. Finally, Section~\ref{Sec8} concludes the survey. In summary, readers interested in the foundational knowledge and principles of medical LLMs and MLLMs should refer to Section~\ref{sec2} to Section~\ref{sec4}, while those focused on applications, challenges, and future directions should consult Section~\ref{Sec5} and Section~\ref{Sec7}.

\section{Background of LLMs and MLLMs}
\label{sec2}
\begin{figure}[ht]
  \centering
  \includegraphics[width=\linewidth]{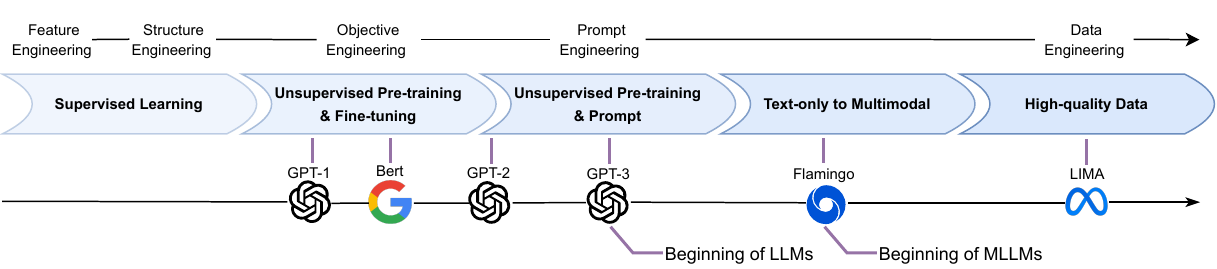}
  \caption{Evolution of LLMs and MLLMs. The Evolution of LLMs and MLLMs. The upper section illustrates the research focuses and paradigm shifts across the evolution of these models, while the lower section highlights key milestones achieved at each stage.}
  \label{fig3}
\end{figure}

This section  focuses on paradigm shifts, categorizing the development of NLP into four distinct stages, as illustrated in Fig.~\ref{fig3}: (1)~Supervised Learning; (2)~Unsupervised Pre-training and Fine-tuning; (3)~Unsupervised Pre-training and Prompt; (4)~Text-only to Multimodal. Recent research~\cite{zhou2024lima} underscores the importance of high-quality datasets for LLMs and MLLMs. Accordingly, we introduce (5) High-quality Data, which examines the shift from reliance on large-scale data to an emphasis on high-quality data.

\subsection{Supervised Learning}
\label{sec2.1}
Supervised learning is a fundamental paradigm in machine learning that focuses on minimizing a loss function. The objective can be expressed as follows:
\begin{flalign}
    &&
    \underset{\boldsymbol{\theta}}{\arg \min } \frac{1}{n} \sum_{i=1}^{n} \mathcal{L}\left(f\left(\boldsymbol{x}_{i} ; \boldsymbol{\theta}\right), y_{i}\right)+\lambda \Omega(\boldsymbol{\theta})
    &&
\end{flalign}
where the first term represents the empirical risk, and the second term represents the regularization term. In supervised learning, a model is trained to map input variables $x$ to output variables $y$ by minimizing the discrepancy between $f\left(\boldsymbol{x} ; \boldsymbol{\theta}\right)$ and $y$, where $\theta$ denotes model parameters, $x$ may consist of manually extracted features or raw text, and $y$ represents supervision signals such as category labels, text, or other forms.

Before the advent of pre-training methods, the supervised learning paradigm dominated the NLP field. Early NLP relied heavily on feature engineering \cite{liu2023pre}, requiring researchers to extract and select features from datasets to perform tasks like text classification \cite{liu2005toward} and machine translation \cite{och2004smorgasbord}. The advent of deep learning~\cite{lecun2015deep} enabled end-to-end model training, shifting research focus from feature engineering to model architecture design, with CNN and LSTM models emerging as prominent approaches. The supervised learning era in NLP marked a shift from feature selection to model architecture design, signifying a transition from feature engineering to structure engineering.

\subsection{Unsupervised Pre-training and Fine-tuning}
\label{sec2.2}
Supervised learning depends on annotated datasets, which establish clear standards for model optimization. However, acquiring sufficient annotated data is challenging for certain tasks, particularly in medical domains, due to the scarcity of specialized annotators and the complexity of the annotation process \cite{zhang2024data}. The introduction of Transformer~\cite{vaswani2017attention} revolutionized the NLP learning paradigm, rendering supervised learning increasingly marginalized~\cite{liu2023pre}.

Transformer-based models like GPT \cite{radford2018improving} and BERT \cite{devlin2018bert} achieved state-of-the-art results through unsupervised pre-training on large-scale unlabeled text, followed by supervised fine-tuning (SFT) using task-specific objective functions. The emergence of GPT and BERT introduced a new NLP paradigm: unsupervised pre-training $\&$ fine-tuning fine-tuning. This paradigm also revolutionized language model development in the medical field, giving rise to prominent models like PubMedBERT~\cite{gu2021domain} and BioBERT~\cite{lee2020biobert}. Compared to earlier models, these approaches offer several advantages: (1) Pre-training data can be drawn from any unannotated text corpus, such as biomedical literature, mitigating challenges related to limited annotated data in medical domains; (2) Training on large-scale unlabeled data enables the acquisition of general and abstract language representations, improving generalization; (3) Fine-tuning for downstream tasks requires only task-specific objective functions, eliminating extensive architectural modifications and facilitating a transition from structural to objective engineering.

\subsection{Unsupervised Pre-training and Prompt}
\label{sec2.3}
While GPT and BERT achieved state-of-the-art results in tasks like machine translation, sentiment analysis, and question-answering (QA), they still required task-specific fine-tuning for each downstream task. To develop a general language model capable of handling diverse tasks without task-specific fine-tuning, Radford et al.~\cite{radford2019language} curated a dataset of over 8 million documents, totaling 40 GB of text data, encompassing examples from multiple domains and tasks, and used it to train GPT-2. GPT-2 set state-of-the-art benchmarks on 7 out of 8 language modeling tasks without requiring task-specific fine-tuning.

To further enhance the generalization capabilities of language models, Brown et al. \cite{brown2020language} scaled the model to 175 billion parameters and significantly expanded the training dataset. This resulted in GPT-3, which demonstrated a qualitative leap in performance, showcasing remarkable few-shot capabilities without requiring fine-tuning. GPT-3 could handle unfamiliar tasks based solely on provided examples, often achieving performance on par with fine-tuned state-of-the-art models. As a result, GPT-3 is widely regarded as the beginning of LLMs \cite{thapa2023chatgpt}. The proposal of GPT-3 further revolutionized NLP by shifting the paradigm from unsupervised pre-training $\&$ fine-tuning to unsupervised pre-training $\&$ prompt \cite{liu2023pre}. Such models can handle most tasks effectively using user prompts and contextual examples. For example, Flan-PaLM~\cite{singhal2023towards} achieved state-of-the-art performance on MedQA, MedMCQA, PubMedQA, and MMLU clinical benchmarks by employing advanced prompting strategies. However, these models are highly sensitive to user-provided prompts, which directly influence the quality of their responses. This sensitivity has spurred researchers to explore prompting strategies in depth \cite{mesko2023prompt}, initiating a shift from objective engineering to prompt engineering.

\subsection{Text-only to Multimodal}
\label{sec2.4}
Inspired by GPT-3, researchers have intensified efforts in developing LLMs, resulting in prominent works such as GLM-130B \cite{zeng2022glm}, PaLM \cite{chowdhery2023palm,anil2023palm}, and LLaMA \cite{touvron2023llama,touvron2023llama2,dubey2024llama}. However, these LLMs remain text-focused, and despite progress in multimodal research, they often require fine-tuning for new tasks \cite{zellers2021merlot,hendricks2021decoupling} or lack text generation capabilities \cite{radford2021learning,li2021align}, which restricts their application scope. Inspired by few-shot learners like GPT-3, Alayrac et al. \cite{alayrac2022flamingo} curated a large-scale multimodal dataset from the web, comprising primarily text-image pairs, to train an MLLM named Flamingo. Flamingo adapts seamlessly to visual tasks via few-shot learning, eliminating the need for task-specific fine-tuning. Flamingo’s robust multimodal in-context learning (ICL) and few-shot capabilities mark it as the GPT-3 moment for the multimodal domain, positioning it as the beginning of MLLMs. These MLLMs employ LLMs as cognitive engines, preserving their inherent capabilities while integrating powerful visual functionalities~\cite{zhang2024mm}. This advancement paves the way for the development of generalist medical AI systems. For example, Med-PaLM M~\cite{tu2024towards} achieved performance levels comparable to or exceeding the state-of-the-art expert models across 14 different medical tasks, showcasing the potential of MLLMs as general medical assistants.

\subsection{High-quality Data}
\label{sec2.5}
A key factor behind the success of LLMs and MLLMs is their use of large-scale training data, which enables the acquisition of universal representations transferable to diverse language understanding and generation tasks \cite{zhou2024lima}. However, much of this training data comes from web sources like WebText \cite{radford2019language} and Common Crawl, and it is inevitable that there are some toxicities and biases in these large amounts of web data, which are also carried over to LLMs and MLLMs \cite{moor2023foundation}. To mitigate the adverse effects of large-scale datasets and improve model performance, researchers often employ high-quality datasets for fine-tuning.	

For example, Li et al.~\cite{li2024llava} utilized GPT-4 to create open-ended instruction-following data derived from biomedical image-caption datasets, subsequently training LLaVA-Med on this data. LLaVA-Med exhibited remarkable multimodal conversational capabilities, adeptly addressing biomedical image queries based on user instructions. To further refine multimodal medical datasets, Xie et al.~\cite{xie2024medtrinity} introduced a dataset enriched with multi-grained annotations. These annotations encompass global context, modality details, and localized descriptions of medical images.	 Training LLaVA-Med on this dataset led to a nearly 10\% average performance boost across three biomedical VQA datasets, emphasizing the critical role of high-quality datasets.	Notably, LIMA \cite{zhou2024lima} fine-tuned using only 1,000 meticulously curated prompts and responses with standard supervised loss, outperformed both Alpaca\cite{alpaca} and Bard in human and GPT-4 preference scores. Ablation experiments on LIMA demonstrated that enhancing data quality provides greater benefits than merely increasing data quantity~\cite{zhou2024lima}. Consequently, it can be predicted that data engineering is emerging as a pivotal area of research focus.~\\

This section outlines the development trajectory of LLMs, emphasizing the shift from supervised learning to unsupervised pre-training $\&$ fine-tuning, and ultimately to unsupervised pre-training $\&$  prompting. The success of LLMs has catalyzed rapid growth in the multimodal domain, leading to the development of MLLMs built on LLM foundations. Notably, GPT-3 and Flamingo, with their robust few-shot capabilities, mark the beginning of LLMs and MLLMs, respectively.	 Recent studies highlighting the role of high-quality datasets in LLMs and MLLMs suggest that data engineering will emerge as a key research focus.	In summary, the evolution of LLMs and MLLMs reflects a progressive shift from initial feature engineering to structure engineering, objective engineering, and now, to prompt and data engineering.

\section{Structure of LLMs and MLLMs}
\label{sec3}
Existing LLMs are universally based on the Transformer architecture, which adopts an encoder-decoder framework.	Accordingly, these LLMs have evolved into three structural variants based on the Transformer architecture: (1) Encoder-only, represented by models such as BERT~\cite{devlin2018bert}; (2) Decoder-only, represented by models such as the GPT series~\cite{{brown2020language,ouyang2022training}}; (3) Encoder-Decoder, represented by models like T5 \cite{raffel2020exploring}. Current MLLMs extend LLMs by integrating a vision encoder to process visual information and a modality alignment module \cite{zhang2024mm,yin2023survey} to bridge the gap between vision and text modalities. This section provides a comprehensive overview of existing medical LLMs and MLLMs, focusing on their respective model architectures.	Section~\ref{sec3.1} reviews medical LLMs categorized by the three aforementioned structures. Section~\ref{sec3.2} discusses the common vision encoders, LLM backbones, and modality alignment techniques employed in medical MLLMs. For clarity, Table \ref{table1} and Table \ref{table2} detail and categorize existing medical LLMs and MLLMs.

\begin{table}
    \tiny
    \centering
    \begin{threeparttable}
    \caption{Detailed information on existing medical LLMs categorized by architecture type.}
    \label{table1}
    \begin{tabular}{llllllll}
    \toprule
    Category & Model Name & Base Model & Para.(B) & Training Data Source & Construction Method & Evaluation Method & Date \\
    \midrule
    \multirow{26}{*}{Decoder-Only} 
        & Med-PaLM \cite{singhal2023large} & PaLM & 540 & MultiMedQA & IFT  & AEM, Human  & 2022/12\\
        & ChatDoctor \cite{li2023chatdoctor} & LLaMA & 7 & Alpaca-52K, HealthCareMagic-100k & IFT & AI & 2023/03\\
        & Baize-Healthcare \cite{xu2023baize} & LLaMA & 7 & Quora, MedQuAD & SFT & AI & 2023/04\\
        & BenTsao \cite{wang2023huatuo} & LLaMA & 7 & CMeKG & SFT & Human & 2023/04\\
        & MedAlpaca \cite{han2023medalpaca} & LLaMA & 7 / 13  & Medical Meadow & IFT & AEM & 2023/04\\
        & PMC-LLaMA \cite{wu2024pmc} & LLaMA & 7 / 13  & MedC-K, MedC-I & CPT, IFT & AEM & 2023/04\\
        & Med-PaLM 2 \cite{singhal2023towards} & PaLM 2 & 340  & MultiMedQA & IFT & AEM, Human & 2023/05\\
        & Clinical Camel \cite{toma2023clinical} & LLaMA 2 & 13 / 70  & ShareGPT, PubMed, MedQA & SFT & AEM & 2023/05\\
        & HuatuoGPT \cite{zhang2023huatuogpt} & BLOOMZ & 7  & Hybrid Data & SFT, RLAIF & AEM, Human, AI & 2023/05\\
        & GatorTronGPT \cite{peng2023study} & GPT-3 & 5 / 20  & Clinical Text from UF Health, Pile & PT & AEM & 2023/06\\
        & ClinicalGPT \cite{wang2023clinicalgpt} & BLOOM & 7  & Three MedQA, MD-EHR, MedDialog & SFT, RLHF & AEM & 2023/06\\
        & Zhongjing \cite{yang2024zhongjing} & Ziya-LLaMA & 13 & CMtMedQA, ChatMed, CMeKG & CPT, SFT, RLHF & Human, AI & 2023/08\\
        & Radiology-Llama2 \cite{liu2023radiology} & LLaMA 2 & 7 & MIMIC-CXR, OpenI & IFT & AEM, Human & 2023/08\\
        & MedChatZH \cite{tan2023medchatzh} & Baichuan & 7 & Books, med-mix-2M & CPT, IFT & AEM, AI & 2023/09 \\
        & CPLLM \cite{shoham2023cpllm} & LLaMA 2 & 13 & eICU-CRD, MIMIC-IV & IFT & AEM & 2023/09 \\
        & ChatCounselor \cite{liu2023chatcounselor} & Vicuna & 7 & Psych8k & IFT & AI & 2023/09\\
        & Qilin-Med \cite{ye2023qilin} & Baichuan & 7 & ChiMed & CPT, SFT, DPO & AEM & 2023/10\\
        & AlpaCare \cite{zhang2023alpacare} & LLaMA & 7 / 13 & MedInstruct-52k & IFT & AI & 2023/10\\
        & TCM-GPT \cite{yang2024tcm} & BLOOM & 7  & TCM-Corpus-1B, TCM-EXAM, TCM-EHR & CPT, SFT & AEM & 2023/11 \\
        & HuatuoGPT-II \cite{chen2023huatuogpt} & Baichuan 2 & 7 / 13 & Web Corpus, Books, Literature, Encyclopedia & IFT & AEM, Human, AI & 2023/11 \\
        & MEDITRON \cite{chen2023meditron} & LLaMA 2 & 7 / 70  & GAP-Replay, MedMCQA, PubMedQA, MedQA & CPT, SFT & AEM & 2023/11 \\
        & AMIE \cite{tu2024towards} & PaLM 2 & 340  & MedQA, MultiMedBench, MIMIC-III, RealWorld Dialogue & IFT & Human, AI & 2024/01 \\
        & BioMistral \cite{labrak2024biomistral} & Mistral & 7  & PubMed Central & CPT, SFT & AEM & 2024/02 \\
        & Me-LLaMA \cite{xie2024me} & LLaMA 2 & 13 / 70  & Pile, MIMIC-III, MIMIC-IV, MIMIC-CXR, RedPajama & CPT, IFT & AEM & 2024/02 \\
        & Apollo \cite{wang2024apollo} & Qwen & 7 & ApolloCorpora & CPT, SFT & AEM & 2024/03 \\
        & BioMedLM \cite{bolton2024biomedlm} & Transformer & 2.7 & PubMed Center, Pile & PT, SFT & AEM & 2024/03 \\
        & PediatricsGPT \cite{yang2024pediatricsgpt} & Baichuan 2 & 7 / 13 & PedCorpus & CPT, SFT, DPO & AEM, Human, AI & 2024/05 \\
        
    \midrule
    \multirow{0}{*}{Encoder-Decoder} 
        & DoctorGLM \cite{xiong2023doctorglm} & ChatGLM & 6  & ChatDoctor, HealthcareMagic, MedDialog, CMD. & IFT & Human & 2023/04\\
        & BianQue \cite{chen2023bianque} & ChatGLM & 6  & BianQueCorpus & IFT & AEM & 2023/10\\
        & SoulChat \cite{chen2023soulchat} & ChatGLM & 6  & SoulChatCorpus & IFT & AEM, Human & 2023/11\\
    \bottomrule
    \end{tabular}
    \begin{tablenotes}
        \item[1] Encoder-only models are not included as they typically belong to the PLM, not LLM.
        \item[2] "CPT" means continuous pre-training, "IFT" means instruction fine-tuning, "SFT" means supervised fine-tuning, "RLHF" means reinforcement learning from human feedback, "RLAIF" means reinforcement learning from AI feedback, "DPO" means direct preference optimization.
        \item[3] "AEM" means automatic evaluation metrics.
    \end{tablenotes}
    \end{threeparttable}
\end{table}

\subsection{Structure of LLMs}
\label{sec3.1}
\subsubsection{Encoder-only}
Encoder-only language models (LMs) consist of multiple encoder layers within the Transformer architecture, with BERT being the earliest and most representative example. Inspired by BERT, additional encoder-only LMs such as DeBERTa DeBERTa \cite{he2020deberta}, ALBERT \cite{lan2019albert}, and RoBERTa \cite{liu2019roberta} have been developed. Encoder-only LMs commonly utilize the masked language modeling (MLM) task during pre-training, where random tokens in sentences are masked, and the model is trained to predict these tokens accurately.	This pre-training approach equips encoder-only LMs with exceptional natural language understanding capabilities, allowing them to effectively encode and comprehend medical knowledge, thereby improving performance in various medical tasks. Consequently, researchers have focused on developing dedicated encoder-only LMs tailored specifically for the medical domain \cite{lee2020biobert,ji2022mentalbert,gu2021domain}. For example, BioBERT \cite{lee2020biobert}, pre-trained on biomedical corpora, achieved state-of-the-art results in tasks such as biomedical named entity recognition, relation extraction, and QA. MentalBERT \cite{ji2022mentalbert} was trained on datasets of mental health disorders (e.g., depression, anxiety, suicidal ideation) sourced from social platforms like Reddit and Twitter, facilitating its application in mental health research.

Despite the presence of numerous encoder-only LMs in the medical domain, these models are better classified as pre-trained language models (PLMs)~\cite{he2023survey,wang2023pre} rather than LLMs. This distinction arises because they require fine-tuning for downstream tasks, lacking the robust ICL and few-shot capabilities of models like GPT-3. Therefore, these PLMs will not be further addressed in the following sections. 

\subsubsection{Decoder-only}
Decoder-only models are the dominant architecture for LLMs, comprising multiple decoder layers within the Transformer. The first decoder-only language model was GPT, and GPT-3 later marked a beginning of LLMs, paving the way for numerous other notable decoder-only models~\cite{chowdhery2023palm,anil2023palm,ouyang2022training,touvron2023llama,touvron2023llama2}. Decoder-only LLMs primarily use next-token prediction (NTP) as their pre-training objective. In this process, the model learns to predict the next token in a sequence based on all preceding tokens. This training paradigm equips decoder-only LLMs with remarkable generative capabilities, enabling them to convert discriminative tasks into generative ones. This unification of task formats enhances both their generalization and adaptability across application scenarios. For example, Med-PaLM M~\cite{tu2024towards} excels in a range of tasks, such as text-based QA, VQA, image classification, radiology report generation, and summarization, achieving or exceeding state-of-the-art performance.

Compared to encoder-only LMs, these decoder-only LLMs utilize NTP as the pre-training task, which enhances their proficiency in text generation. Additionally, studies~\cite{wang2022language,dai2022can} have demonstrated that decoder-only LLMs exhibit the best few-shot and zero-shot performance across diverse downstream tasks, which is one of the reasons why decoder-only has become the predominant framework for LLMs today.

\subsubsection{Encoder-Decoder}
Encoder-decoder LLMs leverage the Transformer architecture, combining stacks of encoders and decoders. The encoder processes input sequence and outputs representations with contextual information, which the decoder uses for text generation. Prominent examples of encoder-decoder LLMs T5 \cite{raffel2020exploring} and GLM \cite{du2021glm}. Similar to encoder-only and decoder-only architectures, encoder-decoder LLMs have also been adapted for medical applications. For example, Chen et al. \cite{chen2023soulchat} fine-tuned ChatGLM using the empathetic dialogue dataset SoulChatCorpus. The resulting model exhibited robust empathetic capabilities, assisting users in articulating their thoughts and offering suitable suggestions during psychological counseling.

While encoder-decoder LLMs integrate the strengths of encoder-only and decoder-only architectures, balancing text understanding and generation, Wang et al. \cite{wang2022language} showed that decoder-only LLMs excel in zero-shot scenarios without fine-tuning. In contrast, encoder-decoder LLMs require multitask fine-tuning with annotated data to reach optimal performance. Since the prevailing LLM training paradigm relies on unsupervised learning on large-scale corpora, decoder-only architectures, with their superior zero-shot performance, are better suited to exploit unlabeled data. As a result, decoder-only architectures remain the predominant choice for LLMs.

\subsection{Structure of MLLMs}
\label{sec3.2}
\begin{figure}[ht!]
  \centering
  \includegraphics[width=\linewidth]{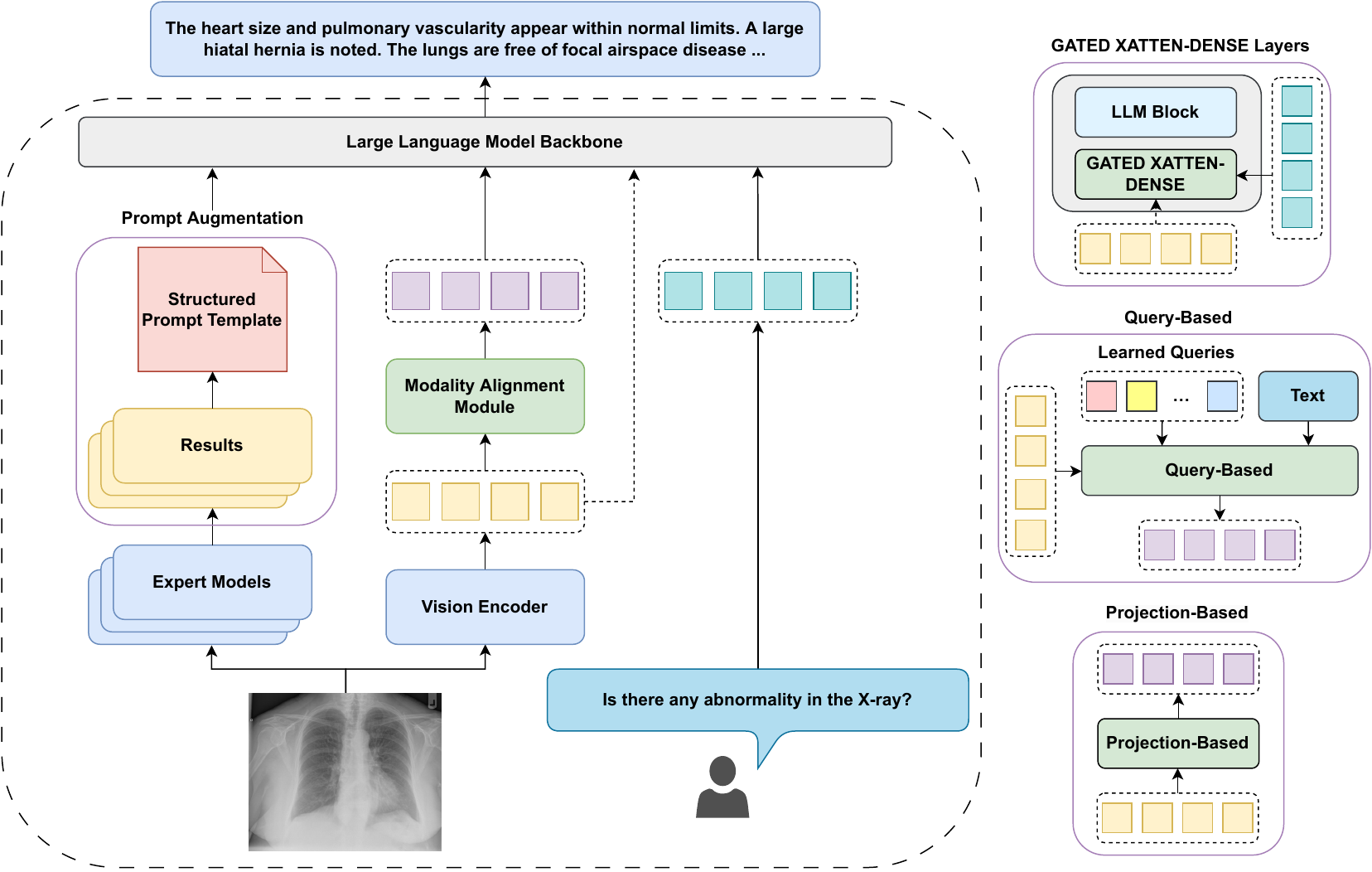}
  \caption{The core modules and pipeline of MLLMs. On the far right are three types of modality alignment modules. The approach of utilizing expert models to construct MLLMs is regarded as a type of prompt augmentation method, classified under modality alignment modules for further elaboration.}
  \label{fig4}
\end{figure}

As shown in Fig.~\ref{fig4}, this section provides a detailed discussion of three critical modules in MLLMs: the Vision Encoder, the LLM Backbone, and the Modality Alignment Module. The method of leveraging expert models to construct MLLMs is treated as a form of prompt augmentation method~\cite{shu2023visual} and is discussed alongside other modality alignment modules.

\subsubsection{Vision Encoder}
\label{sec3.2.1}
MLLMs extend LLMs by incorporating a vision encoder, thereby equipping them with visual capabilities. Specifically, the vision encoder $V$ encodes visual input $I_{x}$ into visual features $Z_{x}$, illustrated below:
\begin{flalign}
    &&
    Z_{x}=V(I_{x})
    &&
\end{flalign}

Various options exist for the vision encoder $V$, such as ResNet \cite{he2016deep}, the ViT\cite{dosovitskiy2020image}, and CLIP-ViT\cite{radford2021learning}. Here, we provide several insights regarding different vision encoders. (1) Transformer-based vision encoders typically outperform CNN-based counterparts due to their superior scalability. Pre-training on large-scale datasets enables these encoders to extract more effective visual representations. Furthermore, as most LLMs are based on the Transformer architecture, Transformer-based vision encoders integrate more seamlessly with these models.	(2) Chen et al.~\cite{chen2023pali3} observed that contrastive-learning-based vision encoders often outperform classification-based ones, particularly in localization and visual-text understanding tasks. This advantage likely arises from the pre-training tasks of contrastive-learning-based encoders, which inherently involve vision-language alignment, enabling the extraction of visual features more aligned with semantic spaces for MLLMs.	(3) In the medical domain, vision encoders pre-trained on medical imaging datasets outperform those trained on natural scene datasets. Models such as LLaVA-Med~\cite{li2024llava}, MAIRA-1~\cite{hyland2023maira}, and PathChat~\cite{lu2024multimodal} provide evidence supporting this claim.

In summary, while ResNet remains a strong choice among CNNs, Transformer-based ViT models are increasingly preferred by researchers. Additionally, contrastive-learning-based ViT models, such as CLIP-ViT and EVA-CLIP ViT~\cite{fang2023eva}, often outperform classification-pretrained ViT models when used as vision encoders for MLLMs. Consequently, ViT models trained using contrastive learning represent the mainstream choice for vision encoders.Furthermore, vision encoders pre-trained on medical imaging datasets significantly enhance the performance of medical MLLMs.	

\subsubsection{LLM Backbone}
\label{sec3.2.2}
As the cognitive engine of MLLMs, the LLM backbone is the most critical module among the three primary components and contains the largest number of parameters. It equips MLLMs with capabilities such as text interaction, ICL, and reasoning. The operational principle of the LLM backbone in MLLMs is illustrated below:
\begin{flalign}
    &&
    R=L(H_{x},T_{x})
    &&
\end{flalign}

Where $L$ represents the LLM backbone, $R$ denotes the response output of the LLM, $T_{x}$ indicates the embedded tokens of the text input, and $H_{x}$ are visual representations that LLM can understand. The specific meaning of $H_{x}$ is explained in Equation (\ref{equation4}).

Although powerful LLMs like ChatGPT have not yet been open-sourced, numerous high-quality open-source LLMs are available for researchers. Among these, the LLaMA series\cite{touvron2023llama,touvron2023llama2, dubey2024llama} developed by Meta, stands out as one of the most popular open-source LLMs and is frequently used as the backbone for MLLMs. Additionally, fine-tuned versions of LLaMA, such as Vicuna-13B~\cite{vicuna2023}, achieve performance comparable to 90\% of ChatGPT and Bard. Notably, different models demonstrate varying levels of performance across languages. For example, Mistral~\cite{jiang2023mistral} excels in French, Qwen~\cite{bai2023qwen} is optimized for Chinese, and GPT-4 offers robust support for a broader range of languages. Consequently, researchers can choose LLM backbones based on specific linguistic requirements.

\begin{table}
    \tiny
    \centering
    \caption{Detailed information on existing medical MLLMs.}
    \label{table2}
    \begin{tabular}{lllllll}
    \toprule
    Modality Alignment Method & Model Name & Vision Encoder  & LLM Backbone & Data Source & Evaluation Method & Date \\
    \midrule
    \multirow{-1}{*}{GATED XATTN-DENSE Layers} 
    & Med-Flamingo \cite{moor2023med} & CLIP-ViT & LLaMA & MTB, PMC-OA & AEM, Human & 2023/07 \\
    \midrule
    \multirow{5}{*}{Query-Based} 
    & MedBLIP \cite{chen2023medblip} & EVA-CLIP-ViT & BioMedLM & ADNI, NACC, OASIS & AEM & 2023/05 \\
    & XrayGLM \cite{wang2023XrayGLM} & ViT-G & ChatGLM & MIMIC-CXR, OpenI & / & 2023/05 \\
    & PCLmed \cite{yang2023customizing} & EVA-CLIP-ViT & ChatGLM & ImageCLEF 2023 caption prediction & AEM & 2023/06 \\
    & RadFM \cite{wu2023towards} & 3D ViT & MedLLaMA-13B & MedMD, RadMD & AEM, Human & 2023/08 \\
    & CheXagent \cite{chen2024chexagent} & EVA-CLIP-ViT & Mistral & CheXinstruct & AEM, Human & 2024/01 \\
    \midrule
    \multirow{15}{*}{Projection-Based} 
    & CLIP-ViT w/ GPT2 \cite{van2023open} & CLIP-ViT & GPT2-XL & Slake, PathVQA, OVQA & AEM & 2023/05 \\
    & MedVInt \cite{zhang2023pmc} & PMC-CLIP-ViT & PMC-LLaMA & PMC-VQA & AEM & 2023/05 \\
    & PathAsst \cite{sun2024pathasst} & PathCLIP-ViT & Vicuna & PathCap, PathInstruct & / & 2023/05 \\
    & LLaVA-Med \cite{li2024llava} & CLIP-ViT & LLaMA & PMC-15M, VQA-RAD, SLAKE, PathVQA & AEM, AI & 2023/06 \\
    & XrayGPT \cite{thawkar2023xraygpt} & MedCLIP-ViT & Vicuna & MIMIC-CXR, OpenI & AEM, AI & 2023/06 \\
    & Med-PaLM M \cite{tu2024towards} & ViT-e, ViT-22B & PaLM & MultiMedBench & AEM, Human & 2023/07 \\
    & R2GenGPT \cite{wang2023r2gengpt} & Swin Transformer & LLaMA 2 & IU-Xray, MIMIC-CXR & AEM &2023/09 \\
    & Qilin-Med-VL \cite{liu2023qilin} & ViT & LLaMA-2-Chinese & ChiMed-VL & / & 2023/10 \\
    & MAIRA-1 \cite{hyland2023maira} & RAD-DINO & Vicuna & MIMIC-CXR & AEM, Human & 2023/11 \\
    & PeFoM-Med \cite{he2024pefomed} & EVA-CLIP-ViT & LLaMA 2 & ROCO, VQA-RAD & AEM, Human & 2024/01 \\
    & M3D-LaMed \cite{bai2024m3d} & 3D ViT & LLaMA-2 & M3D-Data & AEM, AI & 2024/03 \\
    & MoE-TinyMed \cite{jiang2024moe} & CLIP-ViT & Phi-2 & LLaVA-Med, VQA-RAD, SLAKE, PathVQA & AEM & 2024/04 \\
    & MAIRA-2 \cite{bannur2024maira} & Rad-DINO & Vicuna & MIMIC-CXR, PadChest, USMix & AEM & 2024/06 \\
    & PathChat \cite{lu2024multimodal} & UNI & LLaMA 2 & PubMed, WSIs & AEM, Human & 2024/06 \\
    & HuatuoGPT-Vision \cite{chen2024huatuogpt} & CLIP-ViT & Yi-1.5 & PubMedVision, HuatuoGPT-II & AEM & 2024/06 \\
    & miniGPT-Med \cite{alkhaldi2024minigpt} & EVA-CLIP-ViT & LLaMA 2 & MIMIC, NLST, SLAKE, RSNA, RadVQA & AEM & 2024/07 \\
    & SkinGPT-4 \cite{zhou2024pre} & ViT & LLaMA 2 & SKINCON, Dermnet & Human & 2024/07 \\
    & LLaVA-Med++ \cite{xie2024medtrinity} & CLIP-ViT & LLaMA & MedTrinity-25M, VQA-RAD, SLAKE, PathVQA & AEM & 2024/08 \\
    & SigPhi-Med \cite{zhou4988925sigphi} & SigLIP & Phi-2 & LLaVA-Med, VQA-RAD, SLAKE, PathVQA & AEM & 2024/10 \\
    \midrule
    \multirow{4}{*}{Prompt Augmentation} 
    & ChatCAD \cite{wang2024interactive} & Expert models & ChatGPT & MIMIC-CXR, CheXpert & AEM  & 2023/02 \\
    & Visual Med-Alpaca \cite{shu2023visual} & Expert models & Med-Alpaca & ROCO, BigBIO & / & 2023/04 \\
    & ChatCAD+ \cite{zhao2023chatcad+} & Expert models & ChatGPT & CheXpert, MIMIC-CXR & AEM & 2023/05 \\
    & OphGLM \cite{gao2023ophglm} & Expert models & ChatGLM & Web data, MedDialog  & AEM & 2023/06 \\
    \bottomrule
    \end{tabular}
\end{table}

\subsubsection{Modality Alignment}
\label{sec3.2.3}
Although integrating a vision encoder into LLMs enables them to process visual inputs, LLMs trained exclusively on text datasets cannot interpret the output features ${Z}_{x}$ produced by the vision encoder. Consequently, modality alignment is required to transform ${Z}_{x}$ into a format that LLMs can understand, as shown in Equation~(\ref{equation4}):
\begin{flalign}
    &&
    H_{x}=f(Z_{x})
    &&
    \label{equation4}
\end{flalign}

Where $f$ represents the modality alignment method, and $H_{x}$ refers to visual representations that LLMs can understand. Modality alignment plays a critical role in enabling MLLMs to interpret visual information and significantly enhances their multimodal capabilities. The subsequent sections introduce four established modality alignment methods: GATED XATTN-DENSE Layers, Query-Based method, Projection-Based method, and Prompt Augmentation.

\textbf{GATED XATTN-DENSE Layers}, introduced in Flamingo \cite{alayrac2022flamingo}, incorporate dense cross-attention layers into a frozen pre-trained LLM. These cross-attention layers receive information from the vision encoder's output, which is typically processed through a Perceiver Resampler~\cite{jaegle2021perceiver} to reduce the computational complexity of vision-text cross-attention. Using additional cross-attention layers, the LLM generates text responses based on visual representations. Subsequent works such as Med-Flamingo \cite{moor2023med}, which are based on Flamingo, also utilize these cross-attention layers for modality alignment.

\textbf{Query-Based} method, often considered a multimodal perceiver \cite{song2023bridge}, extracts information from visual representations through a set of learnable query vectors. For example, the Q-Former introduced in BLIP-2 \cite{li2023blip} extracts visual features from a frozen vision encoder, enabling LLMs to generate text responses aligned with visual information.  
This query-based approach can be effectively extended to 3D spaces, as demonstrated by Chen et al. \cite{chen2023medblip} with MedBLIP, which adapts the querying mechanism for 3D medical imaging. Although these methods can represent images with only a few queries, thus reducing training costs, they risk losing critical visual information.	 Moreover, Yao et al. \cite{yao2024deco} demonstrated that the Q-Former functions merely as an inefficient visual token compressor. For cost-effective token reduction, adaptive average pooling outperforms it.

\textbf{Projection-Based} method can be regarded as a type of multimodal converter \cite{song2023bridge}. It is simpler than the query-based method, as it maps visual representations from the vision encoder’s output to the word embedding space using a simple projection layer, allowing LLMs to interpret images. For example, LLaVA-Med, Qilin-Med-VL, and XrayGPT \cite{thawkar2023xraygpt} employ a simple linear layer to map visual representations, and MedVIntTE \cite{zhang2023pmc} and LLaVA-1.5 \cite{liu2023improved} rely on MLPs for this purpose. The mapped visual representations, combined with textual representations, act as inputs to the LLM backbone. Compared to query-based methods, projection-based methods preserve more visual information because they do not reduce the number of vision tokens, though this advantage comes at the cost of increased training complexity.

\textbf{Prompt Augmentation} typically processes images using expert models, integrates their results with text prompt templates to serve as input for LLMs, thereby linking visual information with text. For example, OphGLM \cite{gao2023ophglm} extracts information from fundus images through classification and segmentation models, embeds this information into structured text templates, and forms diagnostic reports that act as inputs to LLMs. Similarly, in ChatCAD \cite{wang2024interactive}, X-ray images are processed by trained computer-aided diagnosis (CAD) models to generate outputs, which are subsequently transformed into natural language via prompt templates and used as inputs to LLMs. Compared to query-based and projection-based methods, prompt augmentation leverages expert models without requiring additional modality alignment training. However, this method heavily depends on the quality of prompt templates and the performance of expert models.

Despite their differences, all four approaches share a text-centered design philosophy, leveraging text as a modality space to convert visual information into textual space, thereby enabling LLMs to interpret visual input.

\section{Principles of Medical LLMs and MLLMs}
\label{sec4}
To assist researchers and medical professionals in developing their own medical LLMs and MLLMs, this section summarizes the medical datasets available for training, explains the methods for fine-tuning medical LLMs and MLLMs, and finally discusses three approaches for evaluating the performance of medical LLMs and MLLMs.

\subsection{Training Datasets}
\label{sec4.1}
The currently available datasets are categorized into six main types: electronic health records (EHRs), scientific literature, QA, dialogue, medical image-text pairs, and instruction-following data. Table 3 provides detailed information about these datasets.

\begin{table}
    \scriptsize
    \centering
    \resizebox{\textwidth}{!}{
    \begin{threeparttable}
    \caption{Summary of medical datasets for pre-training and fine-tuning.}
    \label{table3}
    \begin{tabular}{lllc}
    \toprule
    Datasets & Type & Description & AI Synthesis \\
    \midrule
    \multicolumn{4}{c}{\textit{Datasets For Medical LLMs}} \\
    \midrule
    MIMIC-III \cite{johnson2016mimic}& EHR & Approximately 2M de-identified notes. & \textcolor{red}{\XSolidBrush} \\
    MIMIC-IV \cite{johnson2023mimic}& EHR & About 300K patients, 430K admissions. & \textcolor{red}{\XSolidBrush} \\
    CPRD \cite{herrett2015data}& EHR & Anonymized medical records for over 11.3M patients. & \textcolor{red}{\XSolidBrush} \\
    PubMed & Literature & Over 34M citations and abstracts of biomedical literature, about 4.5B words. & \textcolor{red}{\XSolidBrush} \\
    PMC & Literature & Provides free full-text access to PubMed, about 13.5B words. & \textcolor{red}{\XSolidBrush} \\
    CORD-19 \cite{wang2020cord}& Literature & More than 140K papers, with more than 72K full text. & \textcolor{red}{\XSolidBrush} \\
    PubMedQA \cite{jin2019pubmedqa}& QA & 1K labeled, 612K unlabeled and 211.3K manually generated QA. & \textcolor{red}{\XSolidBrush} \\
    MedQA (USMLE) \cite{jin2021disease}& QA & 61,097 multiple-choice QA pairs. & \textcolor{red}{\XSolidBrush} \\
    MedMCQA \cite{pal2022medmcqa}& QA & 194K multiple-choice QA pairs. & \textcolor{red}{\XSolidBrush} \\
    cMedQA2 \cite{zhang2018multi}& QA & 100K questions and 200k answers. & \textcolor{red}{\XSolidBrush} \\
    MultiMedQA \cite{singhal2023large}& QA & Includes six existing datasets and one new dataset. & \textcolor{red}{\XSolidBrush} \\
    MedQuAD \cite{ben2019question}& QA & 47,457 question-answer pairs from trusted medical sources. & \textcolor{red}{\XSolidBrush} \\
    Medical Meadow \cite{han2023medalpaca}& QA & Over 160K QA pairs. & \textcolor{green}{\Checkmark} \\
    Huatuo-26M \cite{li2023huatuo}& QA & 26M QA pairs. & \textcolor{red}{\XSolidBrush} \\
    Psych8k \cite{liu2023chatcounselor}& QA & 8,187 query-answer pairs. & \textcolor{green}{\Checkmark} \\
    CMtMedQA \cite{yang2024zhongjing}& Dialogue & 70K multi-round conversation datasets from real doctor-patient conversations. & \textcolor{green}{\Checkmark} \\
    MedDialog \cite{zeng2020meddialog}& Dialogue & 3.4M Chinese conversations and 0.6 million English conversations. & \textcolor{red}{\XSolidBrush} \\
    HealthCareMagic-100k \cite{li2023chatdoctor}& Dialogue & 100K authentic patient-doctor conversations. & \textcolor{red}{\XSolidBrush} \\
    GenMedGPT-5k \cite{li2023chatdoctor}& Dialogue & 5K generated conversations between patients and physicians from ChatGPT. & \textcolor{green}{\Checkmark} \\
    MedC-I \cite{wu2024pmc}& Instruction-Following Data & 202M tokens. & \textcolor{green}{\Checkmark} \\
    MedInstruct-52k \cite{zhang2023alpacare}& Instruction-Following Data & 52K instruction-response pairs generated by GPT-4. & \textcolor{green}{\Checkmark} \\
    UMLS \cite{bodenreider2004unified}& Knowledge Base & 2M entities for 900K concepts. & \textcolor{red}{\XSolidBrush} \\
    CMeKG \cite{byambasuren2019preliminary}& Knowledge Base & Chinese medical knowledge graph. & \textcolor{red}{\XSolidBrush} \\
    COMETA \cite{basaldella2020cometa}& Web Data & Consisting of 20K English biomedical entity mentions. & \textcolor{red}{\XSolidBrush} \\
    TCM-Corpus-1B \cite{yang2024tcm}& Web Data & 20GB dataset collected from Baidu Baike, Wikipedia and other sources. & \textcolor{red}{\XSolidBrush} \\
    ChiMed \cite{ye2023qilin}& Hybrid & Composed of various data such as QA, books, dialogues, etc. & \textcolor{red}{\XSolidBrush} \\
    GAP-REPLAY \cite{chen2023meditron}& Hybrid & Includes data from clinical practice guidelines, abstracts, and original articles. & \textcolor{red}{\XSolidBrush} \\
    
    \midrule
    \multicolumn{4}{c}{\textit{Datasets For Medical MLLMs}} \\
    \midrule
    PMC-VQA \cite{zhang2023pmc}& QA & Contains 149K images, 227K VQA pairs. & \textcolor{green}{\Checkmark} \\
    VQA-RAD \cite{lau2018dataset}& QA & 315 radiology images and 3515 QA pairs generated by clinicians. & \textcolor{red}{\XSolidBrush} \\
    Slake \cite{liu2021slake}& QA & 642 radiology images and over 7000 diverse QA pairs. & \textcolor{red}{\XSolidBrush} \\
    PathVQA \cite{he2020pathvqa}& QA & 4,998 pathology images with 32,799 QA pairs. & \textcolor{red}{\XSolidBrush} \\
    MIMIC-CXR \cite{johnson2019mimic}& Image-Report  & 227,835 imaging studies for 65,379 patients. & \textcolor{red}{\XSolidBrush} \\
    OpenI \cite{demner2016preparing}& Image-Report & 7,470 images and 3,955 reports. & \textcolor{red}{\XSolidBrush} \\
    CheXpert \cite{irvin2019chexpert}& Image-Report  & 224,316 chest X-rays with reports. & \textcolor{red}{\XSolidBrush} \\
    ROCO \cite{pelka2018radiology}& Image-Caption  & Contains more than 81K radiologic images, each with a corresponding title, keywords. & \textcolor{red}{\XSolidBrush} \\
    OpenPath \cite{huang2023visual}& Image-Caption & 208,414 pathology images paired with natural language descriptions. & \textcolor{red}{\XSolidBrush} \\
    MedICaT \cite{subramanian2020medicat}& Image-Caption  & 160K images with captions and inline references. & \textcolor{red}{\XSolidBrush} \\
    PathCap \cite{sun2024pathasst}& Image-Caption  & 142K high quality pathology image-caption pairs. & \textcolor{green}{\Checkmark} \\
    MedMD \cite{wu2023towards}& Image-Caption & 15.5M 2D scans, 180k 3D scans, with corresponding captions or diagnosis labels. & \textcolor{green}{\Checkmark} \\
    PMC-OA \cite{lin2023pmc}& Image-Caption & 1.6M image-caption pairs. & \textcolor{red}{\XSolidBrush} \\
    PMC-15M \cite{zhang2023large}& Image-Caption & 15M figure-caption pairs from over 3M articles. & \textcolor{red}{\XSolidBrush} \\
    LLaVA-Med-Alignment \cite{li2024llava}& Image-Caption & 600K image-caption paires from PMC-15M. & \textcolor{green}{\Checkmark} \\
    ChiMed-VL-Alignment \cite{liu2023qilin}& Image-Caption & 580,014 images and context information or descriptions. & \textcolor{green}{\Checkmark} \\
    PubMedVision-Alignment \cite{chen2024huatuogpt}& Image-Caption & 647,031 image-caption pairs. & \textcolor{green}{\Checkmark} \\
    MedTrinity-25M \cite{xie2024medtrinity}& Image-Annotation & Contains 25M samples along with their multigranular annotations. & \textcolor{green}{\Checkmark} \\ 
    LLaVA-Med-Instruct \cite{li2024llava}& Instruction-Following Data & 60K instruction-following data. & \textcolor{green}{\Checkmark} \\
    ChiMed-VL-Instruction \cite{liu2023qilin}& Instruction-Following Data & 469,441 question-answer pairs. & \textcolor{green}{\Checkmark} \\
    PubMedVision-Alignment \cite{chen2024huatuogpt}& Instruction-Following Data & 647,031 instruction-following data. & \textcolor{green}{\Checkmark} \\
    PathInstruct \cite{sun2024pathasst}& Instruction-Following Data & 180K instruction-following data. & \textcolor{green}{\Checkmark} \\
    CheXinstruct \cite{chen2024chexagent} & Instruction-Following Data & An instruction-tuning dataset curated from 28 publicly available datasets & \textcolor{green}{\Checkmark} \\
    ApolloCorpora \cite{wang2024apollo}& Hybrid & 2.5B tokens with data in multiple languages. & \textcolor{red}{\XSolidBrush} \\
    PedCorpus \cite{yang2024pediatricsgpt}& Hybrid & The corpus includes pediatric textbooks, clinical guidelines, and knowledge graphs. & \textcolor{green}{\Checkmark} \\
    M3D-Data \cite{bai2024m3d}& Hybrid & Comprising 120K image-text pairs and 662K instruction-response pairs. & \textcolor{green}{\Checkmark} \\
    
    \bottomrule
    \end{tabular}
    \begin{tablenotes}
        \item[1] "Hybrid" means that the dataset is a mixture of multiple types of data.
        \item[2] "AI Synthesis" indicates that generative AI such as chatGPT and GPT-4 were used during the development of the dataset to assist in generating the data.
    \end{tablenotes}
    \end{threeparttable}
    }
\end{table}

\textbf{Electronic Health Records}: EHRs include personal health records, such as demographic information, summaries of major diseases and health issues, and primary healthcare records. The Medical Information Mart for Intensive Care III (MIMIC-III) \cite{johnson2016mimic} is one of the largest and most widely used publicly available EHR datasets, containing approximately 2 million de-identified notes across 13 specialties, including cardiology, respiratory medicine, and radiology. The MIMIC-III dataset is a valuable resource for developing medical LLMs, as evidenced by models such as AMIE \cite{tu2024towards_conversational} and GatorTron \cite{yang2022large}, which were trained using this dataset. Besides MIMIC-III, other widely used EHR datasets include the Clinical Practice Research Datalink (CPRD) \cite{herrett2015data} and the updated version of MIMIC-III, MIMIC-IV \cite{johnson2023mimic}.

\textbf{Scientific Literature}: Scientific literature, which provides accurate and authoritative medical knowledge, serves as a key source for medical datasets. PubMed, the most widely used repository for biomedical and life science literature, provides access to key resources, including MEDLINE, PubMed Central (PMC), and the NCBI Bookshelf. It indexes citations from over 34 million biomedical research articles. PubMed abstracts include approximately 4.5 billion words, making it an excellent resource for medical training datasets. In addition to PubMed, PubMed Central (PMC) is a widely used repository offering free full-text access, with its articles collectively containing around 13.5 billion words. PubMed and PMC offer high-quality medical literature, often used as sources for other datasets. For instance, PMC-OA \cite{lin2023pmc}, PMC-VAQ \cite{zhang2023pmc}, and PMC-15M \cite{zhang2023large} are three biomedical multimodal datasets extracted from PMC, significantly facilitating the development of medical LLMs \cite{toma2023clinical,wu2024pmc,chen2023meditron} and MLLMs \cite{li2024llava,moor2023med}.

\textbf{Question-Answer}: QA datasets are categorized into two types: discriminative QA \cite{jin2021disease,pal2022medmcqa} and generative QA \cite{zhang2023pmc}. Discriminative QA datasets primarily include multiple-choice questions, whereas generative QA focuses on open-ended questions. Typical QA datasets include PubMedQA \cite{jin2019pubmedqa}, MedQA \cite{jin2021disease}, PMC-VQA \cite{zhang2023pmc}, and MultiMedQA \cite{singhal2023large}. MultiMedQA, in particular, is a comprehensive medical QA dataset encompassing seven sub-datasets that assess the authenticity, helpfulness, accuracy, and potential harm of LLMs' responses. Beyond text-based QA datasets, the medical domain also includes VQA datasets. For example, the classic VQA-RAD \cite{lau2018dataset} and SLAKE~\cite{liu2021slake} are prominent VQA datasets comprising radiology images, questions, and answers, which span various organs and anatomical regions. Such medical VQA datasets have greatly contributed to the advancement of medical MLLMs.	

\textbf{Dialogue}: While datasets like EHRs, scientific literature, and QA pairs enrich LLMs and MLLMs with medical knowledge, relying exclusively on these may result in insufficient long-dialogue interaction capabilities, limiting their practical clinical application. As a result, researchers are focused on developing high-quality dialogue datasets to enhance the models' capabilities in multi-turn conversations. For example, Li et al.~\cite{li2023chatdoctor} curated HealthCareMagic-100k by collecting around 100K authentic doctor-patient dialogues from the HealthCareMagic platform, followed by extensive filtering. To bypass the labor-intensive process of gathering authentic dialogue datasets—requiring extensive filtering and deduplication—Li et al.~\cite{li2023chatdoctor} simulated real dialogue scenarios using ChatGPT to create a synthetic dataset named GenMedGPT-5k.

\textbf{Image-Text Pairs}: In addition to the aforementioned VQA datasets, image-text pairs, including image-caption and image-report datasets, are essential for training medical MLLMs.	These datasets offer crucial context and annotations, enriching the ability of MLLMs to interpret visual information in medical scenarios. For example, PMC-OA~\cite{lin2023pmc} comprises 1.65 million medical image-text pairs sourced from PMC, and it has been utilized to train models like PMC-CLIP~\cite{lin2023pmc} and Med-Flamingo~\cite{moor2023med}. Zhang et al.~\cite{zhang2023pmc} built upon PMC-OA and utilized ChatGPT to generate a diverse set of high-quality QA pairs. After filtering, they created PMC-VQA, which provides 227K VQA pairs. PMC-15M~\cite{zhang2023large}, derived from PMC articles, includes 15 million figure-caption pairs, making it two orders of magnitude larger than MIMIC-CXR~\cite{johnson2019mimic}. Other notable medical image-text datasets, such as ChiMed-VL~\cite{liu2023qilin}, RadMD \cite{wu2023towards}, and Open-I \cite{demner2016preparing}, have also played a significant role in advancing medical MLLMs.

\textbf{Instruction-Following Data}:
The effectiveness of medical LLMs and MLLMs in downstream tasks depends not only on the medical knowledge they acquire but also on their capacity to follow user instructions. Enhanced instruction-following ability enable more accurate comprehension and execution of user directives, thereby improving downstream task performance. To strengthen the instruction-following ability of medical LLMs and MLLMs, training on instruction-following datasets has become widely accepted. These datasets often include instruction-response pairs or image-instruction-response, such as PathInstruct ~\cite{sun2024pathasst} with 180K image-instruction-response pairs, LLaVA-Med-Instruct~\cite{li2024llava} with 60K image-instruction-response pairs, and ChiMed-VL-Instruction~\cite{liu2023qilin} with 469K image-instruction-response pairs. Such datasets effectively enhance models’ instruction-following ability, leading to improved zero-shot performance.	Notably, instructions can include explicit commands or questions; as a result, some works~\cite{li2024llava, liu2023qilin} also considered QA data as instruction-following data.

In addition to the aforementioned data types, various medical knowledge bases \cite{bodenreider2004unified} and web data \cite{yang2024tcm} also serve as training sources for medical LLMs and MLLMs. Different types of data are typically utilized at various training stages. For medical LLMs, scientific literatures and web data are primarily used during pre-training and continual pre-training phases to incorporate medical knowledge and facilitate domain adaptation.	 In contrast, QA, dialogue, and instruction-following data are typically employed during the fine-tuning phase to enhance interaction capabilities and instruction-following performance.	For medical MLLMs, image-caption pairs are widely used during pre-training to align visual features with text representations, while instruction-following data is commonly applied for fine-tuning.	Furthermore, the extensive content of EHRs and scientific literature often makes them foundational sources for other data types. For example, PMC-15M~\cite{zhang2023large} is derived from 3 million PMC articles, yielding 15 million image-caption pairs. Finally, studies have demonstrated that fine-tuning models with substantial high-quality synthetic data generated by ChatGPT can significantly enhance downstream task performance~\cite{tang2023does}. As a result, AI-assisted data generation has emerged as a prevalent strategy in the data-scarce medical field.	For example, Liu et al.~\cite{liu2023chatcounselor} transformed 260 real psychological counseling audio recordings into text and utilized GPT-4 to extract question-answer pairs and generate key summaries, providing supplementary contextual information for constructing the Psych8k dataset.

\subsection{Fine-Tuning Methods}
\label{sec4.2}
\begin{figure}[ht!]
  \centering
  \includegraphics[scale=0.7]{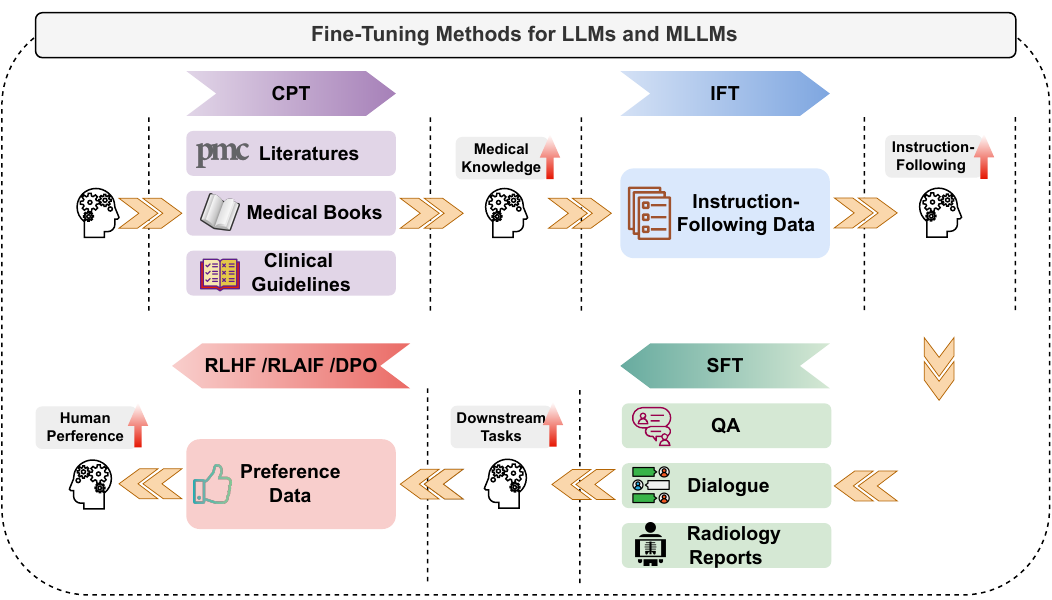}
  \caption{Overview of six fine-tuning methods. In our analysis of the related work on medical LLMs and MLLMs, we found that Continuous Pre-Training (CPT) is commonly used to inject medical knowledge into LLMs and MLLMs; Instruction Fine-Tuning (IFT) enhances the models' ability to follow instructions and their zero-shot performance; Supervised Fine-Tuning is frequently employed to improve model performance on specific tasks; and Reinforcement Learning from Human Feedback (RLHF), Reinforcement Learning from AI Feedback (RLAIF), and Direct Preference Optimization (DPO) are used to align model behavior with human preferences.}
  \label{fig5}
\end{figure}

The extensive parameters in LLMs and MLLMs make training medical LLMs and MLLMs from scratch computationally intensive. Consequently, the prevalent method for constructing medical LLMs and MLLMs involves fine-tuning general foundation models using medical datasets. This section outlines six fine-tuning methods, as illustrated in Fig.~\ref{fig5}, to aid researchers in developing medical LLMs and MLLMs. In addition, this section explores the characteristics of these fine-tuning methods, providing practical guidance for selecting the appropriate fine-tuning strategies.

\textbf{Continual Pre-Training}: CPT \cite{wang2023pre,wu2024continual} refers to the further pre-training of a general pre-trained model on large-scale medical datasets. Given that general LLMs and MLLMs often lack domain-specific medical knowledge, CPT aims to incorporate such knowledge into these models.	For LLMs, datasets comprising extensive medical texts, including scientific literature, books, and clinical guidelines, are typically used in CPT to ensure LLMs acquire sufficient medical knowledge. For example, MEDITRON-70B~\cite{chen2023meditron}, which is based on LLaMA 2, utilizes a medical mixed dataset composed of clinical guidelines, PubMed papers, and abstracts for CPT. For MLLMs, medical image-caption pairs are frequently utilized as CPT datasets. This method aligns visual features with the semantic space by training MLLMs to predict captions for medical images. Prominent models, including LLaVA-Med~\cite{li2024llava}, Qilin-Med-VL~\cite{liu2023qilin}, and HuatuoGPT-Vision~\cite{chen2024huatuogpt}, utilize large-scale medical image-caption datasets for CPT.

\textbf{Instruction Fine-Tuning}: While CPT using large-scale biomedical corpora integrates medical knowledge into LLMs and MLLMs, clinical performance also relies on their instruction-following capability. A lack of instruction-following ability in LLMs and MLLMs can result in unpredictable behavior, underscoring the necessity of fine-tuning these models with instruction-following datasets to improve their compliance with diverse human instructions~\cite{li2024llava,wei2021finetuned}. Instruction fine-tuning (IFT) enables models to accurately understand and execute human directives, thereby significantly enhancing their zero-shot performance. For example, Singhal et al.~\cite{singhal2023large} demonstrated substantial improvements in Flan-PaLM's performance on MedQA, MedMCQA, and PubMedQA after fine-tuning it with instruction data.

\textbf{Supervised Fine-Tuning}: To further improve the performance of LLMs and MLLMs in downstream tasks, SFT is typically performed on datasets tailored to those tasks. For instance, Chen et al.~\cite{chen2023meditron} fine-tuned MEDITRON on MedQA, PubMedQA, and MedMCQA datasets to enhance its performance in medical QA tasks. Similarly, Li et al.~\cite{li2024llava} improved LLaVA-Med's performance in medical VQA tasks by fine-tuning it on PathVQA, SLAKE, and VQA-RAD datasets. Hyland et al.~\cite{hyland2023maira} leveraged large-scale image-report pairs to enhance MAIRA-1's performance in radiology report generation tasks. While prior studies often do not strictly distinguish between SFT and IFT, we argue that IFT focuses on utilizing instruction datasets to enhance models' instruction-following ability and zero-shot performance. Conversely, SFT targets high-quality task-specific datasets to fine-tune LLMs and MLLMs, thereby boosting their performance in specialized downstream tasks.	

\begin{figure}
  \centering
  \includegraphics[width=\linewidth]{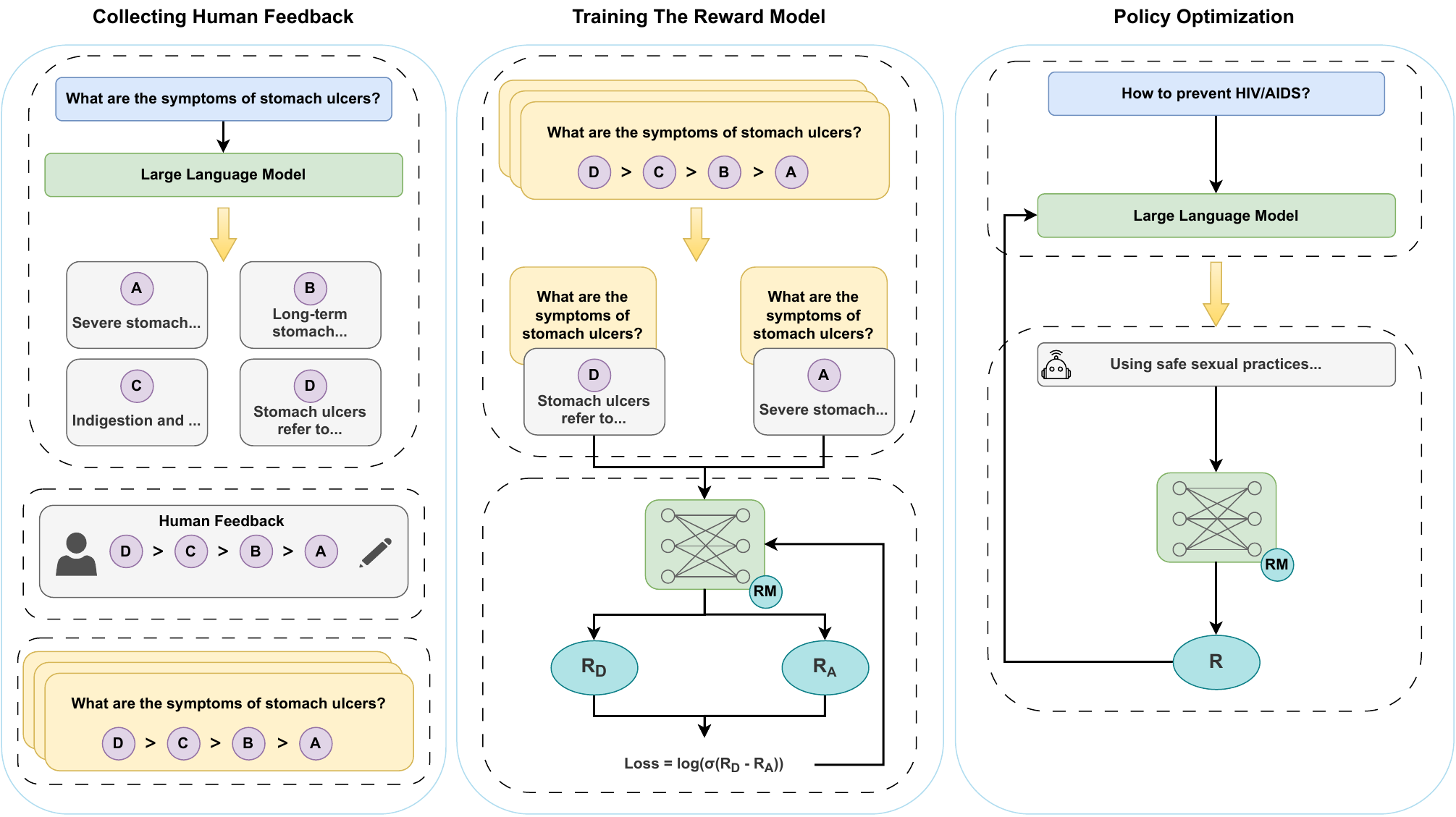}
  \caption{Pipeline of Reinforcement Learning from Human Feedback. \textbf{Left} illustrates the Collect Human Feedback phase: A labeler provides a single prompt to the model, ranks multiple responses, and collect the prompt along with the labeled responses. \textbf{Mid} shows the Reward Model Training phase: A prompt and two responses are randomly sampled from the dataset and used to train the reward model. \textbf{Right} represents the Policy Optimization phase: A new prompt is provided, and the reward model generates a scalar reward for the model's response, which is then used for policy optimization.}
  \label{fig6}
\end{figure}

\textbf{Reinforcement Learning from Human Feedback}: Reinforcement learning from human feedback (RLHF) \cite{ouyang2022training,casper2023open} is a technique for aligning model behavior more closely with human preferences and directives. Compared to the preceding three fine-tuning approaches, RLHF is more intricate and involves three distinct stages \cite{ouyang2022training,casper2023open,stiennon2020learning}: collecting human feedback, training the reward model, and policy optimization, as shown in Fig.~\ref{fig6}. During the human feedback collection stage, the primary task is gathering comparison data. Typically, an LLM is provided with a prompt, generating multiple outputs that experts annotate and score based on quality ~\cite{stiennon2020learning}. These annotated outputs, along with the prompts, form the comparison data. For example, Yang et al.~\cite{yang2024zhongjing} employed 6 medical graduate students or clinical doctors as labelers to rank the model's outputs based on dimensions such as safety, professionalism, and fluency, forming a comparison dataset. During reward model training, a reward model learns from the comparative data to produce scalar rewards that represent human preferences. During policy optimization, a new prompt is provided as input to the LLM, whose response is evaluated by the reward model, outputting a scalar reward. The LLM is then fine-tuned using Proximal Policy Optimization (PPO) based on these rewards. Notably, the reward model's data quality is typically lower than that used for SFT~\cite{he2023survey}, directly transitioning from pre-training to RLHF may lead to suboptimal fine-tuning results, so RLHF is often conducted after IFT and SFT \cite{touvron2023llama2,casper2023open}. 

\textbf{Reinforcement Learning from AI Feedback}: Reinforcement learning from AI feedback (RLAIF) is a cost-effective alternative to RLHF, where the reward model learns from AI feedback without requiring human annotation \cite{bai2022constitutional}. In the medical domain, Zhang et al. \cite{zhang2023huatuogpt} sampled multiple responses from the fine-tuned model after IFT and SFT and used ChatGPT to evaluate them across dimensions such as informativeness, coherence, adherence to human preferences, and accuracy. This comparison data was then used to train a reward model. Training reward models through AI feedback eliminates the need for manual data labeling in RLHF, significantly reducing labor costs.

\textbf{Direct Preference Optimization}: While RLHF and RLAIF align models with human preferences and ethical norms, they typically involve fitting a reward model that reflects human preferences and combining reinforcement learning to fine-tune LLMs and MLLMs. However, this  process is both complex and often unstable. Direct preference optimization (DPO) \cite{rafailov2024direct} is a simpler and more efficient paradigm for aligning models with human preferences, bypassing the need for a reward model by directly optimizing the model using preference data. The core idea of DPO is to use an analytical mapping from the reward function to the optimal policy, transforming the reward function loss into a policy loss and thereby eliminating the need for explicit reward modeling. For example, Qilin-Med \cite{ye2023qilin} uses two publicly available preference datasets after SFT to optimize the model through DPO, achieving stable and efficient training while aligning it with human preferences.

In summary, CPT injects medical knowledge into LLMs and MLLMs. IFT enhances instruction-following capabilities and zero-shot performance. SFT improves task-specific performance in downstream tasks. and RLHF, RLAIF, and DPO further align models with human preferences. In the current work on medical LLMs and MLLMs, CPT, STF, and IFT are the most commonly used fine-tuning methods. In contrast, RLHF, RLAIF, and DPO have been used relatively infrequently in medical LLMs and MLLMs, with no existing medical MLLMs reported to utilize RLHF, RLAIF or DPO.

\subsection{Evaluation Methods}
\label{sec4.3}
\begin{figure}
  \centering
  \includegraphics[scale=0.6]{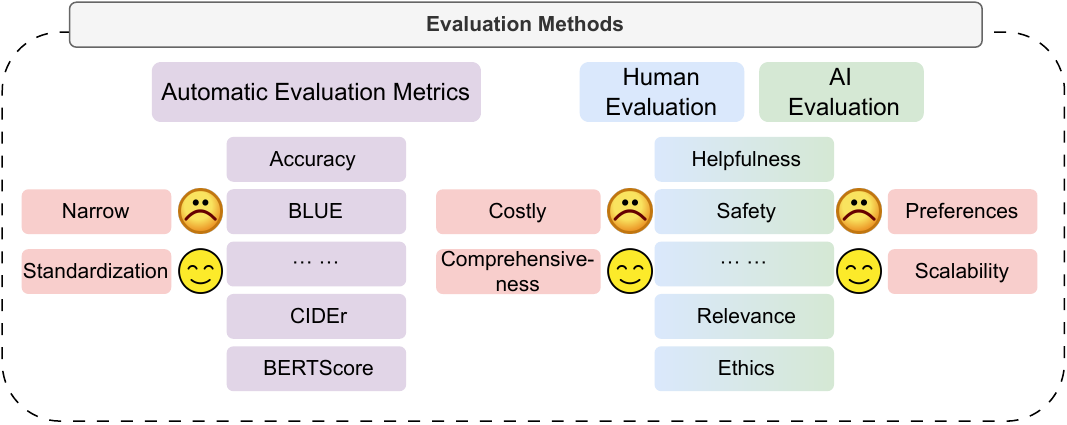}
  \caption{Overview of three evaluation methods. We summarize three methods for evaluating medical LLMs and MLLMs: Automatic Evaluation Metrics, Human Evaluation, and AI Evaluation, and discuss their respective advantages and disadvantages.}
  \label{fig7}
\end{figure}

With the continuous emergence of abilities in medical LLMs and MLLMs, coupled with growing ethical and safety concerns ~\cite{deshpande2023toxicity}, comprehensively evaluating their performance in medical tasks and ensuring their safety in clinical environments has become a pressing challenge. Therefore, this subsection summarizes three commonly used evaluation methods and examines their respective advantages and limitations, as shown in Fig.~\ref{fig7}.

\subsubsection{Automatic Evaluation Metrics}
Accuracy is commonly used to evaluate the performance of medical LLMs and MLLMs on multiple-choice question benchmarks, such as MedQA \cite{jin2021disease} and MedMCQA \cite{pal2022medmcqa}. However, accuracy alone is insufficient for evaluating tasks requiring longer text generation, such as medical report writing and summarization.	  Consequently, the evaluation of medical LLMs and MLLMs must incorporate additional metrics for a more comprehensive assessment.	

Bilingual Evaluation Understudy (BLEU) \cite{papineni2002bleu} metric assesses the quality of generated text by measuring the similarity of n-grams (consecutive word sequences of length $n$) between the generated and reference texts. BLEU is categorized into BLEU-1, BLEU-2, BLEU-3, and BLEU-4 based on the value of $n$, capturing n-gram similarity at different levels. For instance, BLEU-1 reflects word-level accuracy, while BLEU-4 emphasizes text continuity. Recall-Oriented Understudy for Gisting Evaluation (ROUGE) \cite{lin2004rouge} includes metrics such as ROUGE-N, ROUGE-L, ROUGE-W, and ROUGE-S. Similar to BLEU, ROUGE-N evaluates n-gram similarity between generated and reference texts; however, it emphasizes recall, whereas BLEU focuses on precision. ROUGE-L assesses textual coherence by calculating the longest common subsequence between the generated and reference texts. ROUGE-W extends ROUGE-L by applying weighted scoring, giving higher importance to continuous and accurate matching subsequences. ROUGE-S extends ROUGE-N by accommodating non-contiguous words in n-grams. Google BLEU (GLEU) \cite{wu2016google}, a BLEU variant, incorporates lexical overlap and word order, offering a more nuanced evaluation of fluency and naturalness. The Distinct-n \cite{li2015diversity} metric quantifies text diversity by calculating the ratio of unique n-grams to total n-grams. CIDEr \cite{vedantam2015cider}, tailored for image caption evaluation, combines n-gram recall and precision, assigning higher weights to rare n-grams to assess key information. BERTScore \cite{zhang2019bertscore} leverages pre-trained BERT embeddings to compute token-level similarity scores between generated and reference sentences. Compared to n-gram-based metrics, BERTScore more effectively evaluates vocabulary and compositional diversity.	

In the medical domain, models such as HuaTuoGPT\cite{zhang2023huatuogpt}, ClinicalGPT~\cite{wang2023clinicalgpt} \cite{wang2023clinicalgpt}, SoulChat~\cite{chen2023soulchat} and BianQue \cite{chen2023bianque} commonly use these metrics to assess generative performance. While these metrics partially capture the accuracy and fluency of generated text, they fall short in assessing clinical dialogue quality~\cite{tu2024towards_conversational} and alignment with human values, necessitating the inclusion of human evaluation.

\subsubsection{Human Evaluation}
Human evaluation is an essential approach to assess medical LLMs and MLLMs, as it addresses aspects that automatic evaluation metrics may fail to capture. For example, Tu et al. \cite{tu2024towards_conversational} highlighted that metrics such as BLEU and ROUGE fail to reflect the clinical quality of medical consultations. To address this, 23 medical experts from the United States, the United Kingdom, and India were invited to evaluate model-generated responses based on accuracy, appropriateness, and comprehensiveness. Similarly, Yang et al. \cite{yang2024zhongjing} engaged human experts to assess the safety, accuracy, and ethical implications of model-generated responses.

Clearly, human evaluation can encompass critical aspects such as safety and helpfulness, which are essential for medical LLMs and MLLMs. Despite its ability to evaluate diverse capabilities of medical LLMs and MLLMs, human evaluation remains inherently subjective due to the absence of standardized criteria. Moreover, hiring medical experts is costly, making AI evaluation a practical alternative.	

\subsubsection{AI Evaluation}
Employing advanced AI models, such as ChatGPT and GPT-4, which align with human values, has become the predominant method for evaluating medical LLMs and MLLMs \cite{wang2023chatgpt}. Wang et al. \cite{wang2023chatgpt} conducted experiments on five natural language generation datasets, showing that ChatGPT, as an evaluation tool, outperformed traditional metrics in most cases and matched human evaluation. In the medical domain, Li et al. \cite{li2024llava} asked GPT-4 to evaluate responses from itself and LLava-Med on criteria such as helpfulness, relevance, accuracy, and level of detail. Liu et al. \cite{liu2023chatcounselor} instructed GPT-4 to evaluate whether LLM responses were acceptable and whether their tone resembled that of human counselors.

Despite its scalability and reduced reliance on human input, AI evaluation has notable limitations.	Studies \cite{zheng2024judging,xu2023baize} have revealed that GPT-4, as an evaluation tool, tends to favor the first response when multiple answers are presented sequentially. Furthermore, GPT-4 tends to prefer longer responses and those it has generated itself \cite{liu2023chatcounselor}. To mitigate the limitations of the aforementioned methods, integrating multiple evaluation approaches may provide more reliable results. Additionally, training specialized LLMs or MLLMs through reinforcement learning or other methods to align with human judgment criteria could overcome AI evaluation's shortcomings.

\section{Applications of LLMs and MLLMs in Medicine}
\label{Sec5}
\begin{figure}[ht!]
  \centering
      \includegraphics[width=\linewidth]{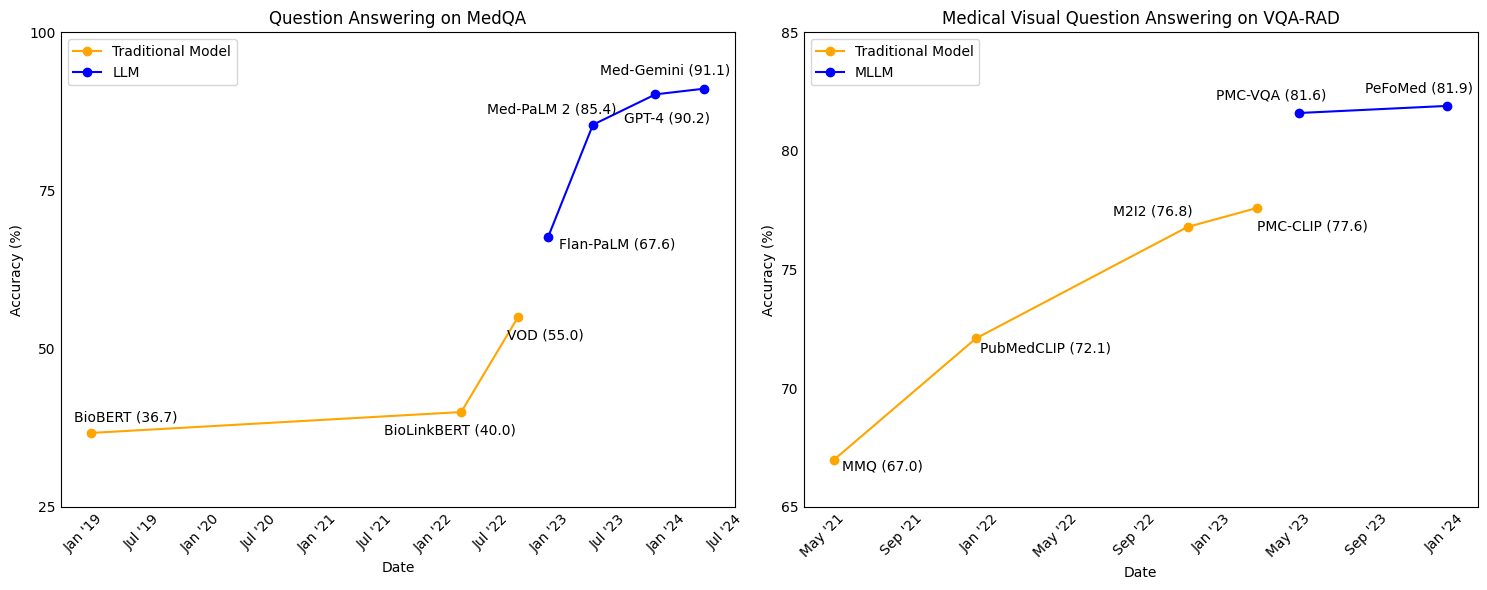}
  \caption{Performance comparison between traditional deep learning models and medical LLMs / MLLMs on QA and VQA benchmark datasets. Recent advancements in medical LLMs / MLLMs demonstrate their significant superiority over traditional deep learning models on QA and VQA benchmark datasets.}
  \label{fig8}
\end{figure}

\begin{figure}
  \centering
  \includegraphics[scale=0.6]{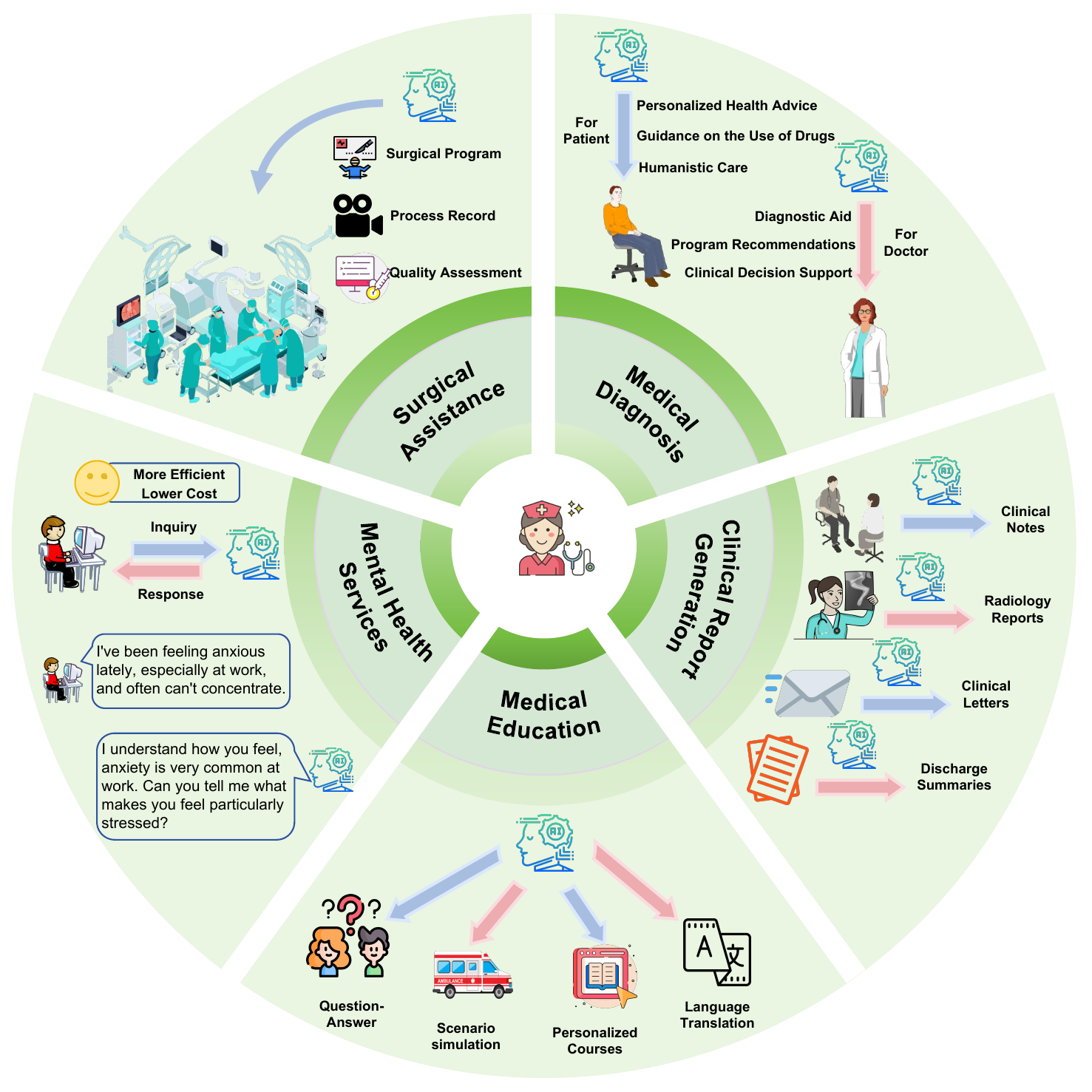}
  \caption{Overview of potential applications of LLMs and MLLMs in medicine.}
  \label{fig9}
\end{figure}

Traditional medical models are tailored for specific tasks, including medical named entity recognition, relation extraction, text classification, and semantic textual similarity. Although these models perform well in their respective tasks, they are limited in their ability to integrate multi-source data for complex clinical applications. In contrast, medical LLMs and MLLMs excel in navigating diverse medical contexts and performing a wide range of tasks, while also gradually surpassing traditional deep learning models in single-task performance, as shown in Fig.~\ref{fig8}. Consequently, medical LLMs and MLLMs act as versatile medical assistants with extensive application potential. To aid practitioners in comprehending the developmental trajectory of LLMs and MLLMs in medicine, this section highlights their potential applications in healthcare (Fig.~\ref{fig9}) and outlines strategies for employing these models in diverse medical tasks.	

\subsection{Medical Diagnosis}
AI has been under development in medical diagnosis for several decades \cite{szolovits1988artificial}. While some breakthroughs have been achieved, its role has largely been confined to assisting with some simple diagnostic tasks, such as medical image segmentation, lesion detection and classification. In recent years, with advancements in LLMs and MLLMs, doctors and patients are now expected to rely on these models for more comprehensive diagnosis. Specifically, these models can process subjective descriptions of disease symptoms or medical images, such as X-rays, and leverage embedded medical knowledge to directly generate diagnostic results. For physicians, medical LLMs and MLLMs shorten diagnostic times and reduces workload. For patients, they offers detailed information about their condition along with recovery suggestions.

Currently, Med-PaLM 2, one of the top-performing medical LLMs, generates responses to consumer medical and adversarial questions, outperforming physician-generated answers across multiple evaluation criteria~\cite{singhal2023towards}. This underscores the potential of LLMs as medical diagnostic assistants. Additionally, Yuan et al.~\cite{yuan2023advanced} showed that multi-turn dialogues with GPT-4 significantly improve its ability to accurately diagnose and recommend effective treatments for gastrointestinal cancers, achieving a performance level comparable to experienced physicians. In prostate cancer, Zhu et al.~\cite{zhu2023can} developed 22 questions based on patient education guidelines and clinical experience, addressing topics like screening, prevention, treatment options, and postoperative complications. Testing indicated that ChatGPT provided accurate and comprehensive responses, while demonstrating appropriate humanistic care toward patients. Yang et al.~\cite{yang2024tcm} further advanced the application of LLMs as diagnostic assistants by training TCM-GPT on traditional Chinese medicine datasets, showing that it outperformed other models in traditional Chinese medicine examination and diagnosis, contributing to the advancement of traditional Chinese medicine. Furthermore, inspired by general MLLMs \cite{achiam2023gpt,liu2024visual}, researchers have developed multimodal medical diagnostic assistants \cite{li2024llava,wang2023XrayGLM,wang2024interactive,tu2024towards,shu2023visual,liu2023medical,thawkar2023xraygpt}, extending diagnostic capabilities from text to medical images. For example, Zhou et al.~\cite{zhou2024pre} created SkinGPT-4, a model capable of autonomously analyzing images, identifying skin condition features and categories, conducting in-depth analyses, and offering interactive treatment recommendations.

Medical LLMs and MLLMs, as medical diagnostic assistants, are capable of providing diagnostic recommendations for doctors and offering consultation advice for patients. However, due to inherent limitations of LLMs and MLLMs~\cite{rawte2023survey,deshpande2023toxicity}, these models currently serve only as auxiliary tools for physicians, with their diagnostic outputs considered as references rather than definitive results.

\subsection{Clinical Report Generation}
Clinical reports are standardized documents prepared by doctors for patients. The manual drafting of clinical reports is often tedious and time-consuming, significantly increasing clinicians' workload and reducing overall efficiency. Medical LLMs and MLLMs, equipped with extensive medical knowledge and generative capabilities, serve as efficient tools for clinical report generation.

For example, during medical consultations, doctors typically record key information from patient interactions, which serves as a basis for evaluating conditions or informing other clinical reports. Medical LLMs can function as clinical note-taking tools, automating this task for doctors \cite{toma2023clinical}. Doctors can provide LLMs with doctor-patient interaction records, which the models process to generate detailed medical notes \cite{lee2023benefits}. Doctors can also prompt LLMs to simplify medical notes by removing complex details and generating concise summaries for easier review and analysis~\cite{van2023clinical}. Following a medical diagnosis, doctors often draft diagnostic documents such as radiology reports. Medical MLLMs, capable of processing visual inputs, are particularly effective in assisting with radiology report generation. For example, miniGPT-Med, developed by Alkhaldi et al.~\cite{alkhaldi2024minigpt}, achieved state-of-the-art performance in generating medical reports, surpassing prior models in accuracy by 19\%. This underscores the feasibility of using medical MLLMs for radiology report generation.	 During treatment, doctors explain the cause of the disease, the treatment process, and provide detailed clinical information to patients through clinic letters. By utilizing LLMs to generate clinic letters, clinicians can streamline this tedious process, with the resulting letters exhibiting coherence, accuracy, and empathy comparable to those created by humans \cite{ali2023using}. After patient recovery, clinicians allocate significant time to drafting discharge summaries, potentially delaying patient discharge. By employing LLMs, clinicians can generate complete discharge summaries in seconds by providing a template and necessary inputs \cite{patel2023chatgpt}. The quality of these summaries often surpasses those produced by junior doctors~\cite{clough2024transforming}.

Leveraging advanced LLMs and MLLMs, various clinical reports from patient admission to discharge can be automatically generated. These reports are more comprehensive and accurate than those produced by humans~\cite{van2023clinical,clough2024transforming}, significantly alleviating doctors' workloads and allowing them to dedicate more time to patient care~\cite{patel2023chatgpt}. However, these LLMs and MLLMs are intended to serve exclusively as auxiliary tools for generating clinical reports. They can draft, modify, and summarize reports, but the final versions must be reviewed, edited, and approved by clinicians, who remain accountable for their content \cite{thirunavukarasu2023large,moor2023foundation}.

\subsection{Medical Education}
GPT-4 and Med-PaLM 2 passed the USMLE with scores exceeding 86\% \cite{nori2023capabilities}, while GPT-4V \cite{yang2023dawn} achieved 90.7\%, outperforming most medical students on medical image-related questions \cite{yang2023performance}. This demonstrates that certain LLMs and MLLMs possess the capability to provide educational support for medical students. Moreover, LLMs and MLLMs can role-play in various contexts, enabling the simulation of diverse learning scenarios for users. Consequently, platforms like Khanmigo~\cite{khan2023harnessing} and Duolingo \cite{team2023introducing} have integrated tools like GPT-4 to enhance online teaching and learning.

Specifically, medical LLMs and MLLMs can simulate patients in scenarios such as accidents, emergency rooms, or operating rooms, offering simulation training to medical students prior to clinical practice to enhance their professional and practical skills—tasks unachievable by traditional deep learning~\cite{karabacak2023advent}. Additionally, medical LLMs and MLLMs can evaluate students’ performance in simulated exercises and create personalized learning plans, a process that is typically time-intensive for teachers but can be executed more cost-effectively and efficiently using these models~\cite{lee2024rise}. Finally, given their extensive training on large corpora, medical LLMs and MLLMs excel at translation, offering cross-language capabilities as well as the ability to simplify medical terminology into plain language~\cite{lyu2023translating},  greatly assisting medical students in reading and writing. In summary, powerful LLMs and MLLMs can enrich medical students’ learning experiences by offering comprehensive medical content, personalized curricula, and realistic, diverse scenarios, thereby broadening their horizons in the medical field and laying a strong foundation for clinical practice.

The potential of LLMs and MLLMs in medical education surpasses that of traditional training courses, as educators in these courses cannot always interact with students or offer tailored learning plans. While these models hold great promise in medical education, they should be regarded solely as auxiliary tools and not as replacements for educators, as their inherent biases and hallucinations make it challenging for students to evaluate the accuracy of the generated content \cite{han2023explorative,ahn2023impending}. If these models frequently deliver inaccurate content that is hard to identify, they risk misleading students over time.	Thus, LLMs and MLLMs can only serve a supportive role in medical education, requiring students to use them under the guidance and supervision of educators.

\subsection{Mental Health Services}
Growing societal pressures have led to an increased global demand for mental health services, yet many regions face a severe shortage of mental health specialists due to limited development and resources~\cite{van2023global}. Conversation-driven psychological counseling is a central aspect of mental health services, making chatbots powered by LLMs a potential solution for delivering such services in the future~\cite{liu2023chatcounselor,chen2023soulchat}.

Compared to professional mental health experts, LLM-based mental health chatbots are more accessible and can extend mental health services to remote or underserved areas lacking specialists. Additionally LLM-based chatbots can offer personalized interaction styles tailored to patients’ histories and interaction records, including specific emotional patterns, styles, or tones~\cite{de2023benefits}. Furthermore, the high cost of psychological counseling and therapy often deters individuals from seeking mental health services. However, LLM-based chatbots can substantially lower costs~\cite{zhou2023survey,stock2023tell}, making these services more accessible. Finally, studies suggest that individuals are more likely to share negative emotions with chatbots, as certain topics may feel awkward to discuss with humans but easier to disclose to a robot~\cite{chaves2021should}. In terms of convenience, cost, and acceptability, LLM-based mental health chatbots surpass mental health professionals, potentially encouraging more individuals with mental illnesses to seek help~\cite{de2023benefits}.

Given that patients with mental illnesses are often more vulnerable and psychologically sensitive, mental health service chatbots must exhibit empathy, trustworthiness, understanding, and comfort in conversations, beyond merely offering advice~\cite{chen2023soulchat}.  While research is advancing the empathetic capabilities of LLMs~\cite{chen2023soulchat}, they still fall short compared to humans. Moreover, despite efforts to align LLMs with human concepts and ethical norms using approaches like SFT and RLHF, they may still produce aggressive or psychologically harmful content~\cite{deshpande2023toxicity}, which is unacceptable for vulnerable mental health patients. Before deploying LLMs as mental health chatbots in real-world applications, significant work is required to address these challenges, along with stricter control measures for such products.

\subsection{Surgical Assistance}
Medical robots have undergone rapid development over the past few decades, significantly enhancing surgeons' capabilities~\cite{BAI2025102602} and expanding the potential for minimally invasive surgery~\cite{barua2024innovations}. Recently, medical robots have entered a new phase with the advent of MLLMs, which not only endow them with visual capabilities but also enhance interactivity and create a more user-friendly environment.

Efforts are currently underway to explore the application of MLLMs in surgical procedures \cite{seenivasan2023surgicalgpt}. Integrating MLLMs into surgical robots enables them to perform crucial auxiliary tasks, such as assisting in endoscopic examinations~\cite{moor2023foundation}. The robust visual capabilities and specialized knowledge of MLLMs can yield valuable diagnostic conclusions and feasible surgical solutions based on endoscopic images. Furthermore, during surgical procedures, MLLMs can combine video streams to annotate the surgical process, analyze and summarize the steps taken, and record non-compliant operations to assist in the surgeon's post-surgical review and examination.

While medical MLLMs exhibit promising potential for surgical assistance and may play a role in specific medical scenarios, they are not yet suitable for emergency surgeries. This is due to the fact that erroneous information from MLLMs could adversely impact the surgeon’s judgment, potentially leading to irreversible consequences. Additionally, current research on MLLMs predominantly focuses on vision-text modalities. We anticipate future investigations will explore other modalities, such as audio and time series, to enable surgical robots to perform more comprehensive and accurate auxiliary tasks while providing more flexible interaction methods.

\section{Challenges of LLMs and MLLMs in Medicine}
\label{Sec6}
Although LLMs and MLLMs have generated significant interest in the AI community and achieved initial successes in medicine, the unique characteristics of the medical field present numerous challenges and risks for their development and deployment. In this section, we will discuss and analyze the current challenges faced by LLMs and MLLMs in the medical field in detail, as well as propose possible solutions to these challenges.

\subsection{Hallucination Phenomenon}
Hallucinations refer to the generation of seemingly plausible yet unverified or incorrect information by LLMs and MLLMs \cite{rawte2023survey,ji2023survey}. This can result in issues such as misleading radiology reports and the dissemination of incorrect medical knowledge in medical education~\cite{lee2023benefits}. Therefore, addressing the hallucination problem in LLMs and MLLMs is crucial for accelerating the application of these technologies in medicine.

To tackle this challenge, some researchers have proposed a new benchmark dataset specifically for testing hallucinations in medical LLMs and MLLMs~\cite{umapathi2023med}. However, these benchmark datasets can only detect hallucination phenomena in models and do not directly mitigate the problem. Other studies indicate that LLMs primarily acquire knowledge during the pre-training phase \cite{zhou2024lima}, and the presence of noisy data, such as error messages in the training dataset, can contribute to hallucinations. Thus, the most fundamental approach to reducing hallucinations is to manually or automatically clean unreliable data from the pre-training corpus~\cite{ji2023survey}. However, the pre-training corpus of LLMs and MLLMs typically consists of vast amounts of data, including web-crawled information, which is challenging to clean and necessitates effective selection and filtering strategies. Consequently, fine-tuning LLMs and MLLMs with high-quality medical datasets is an effective strategy for mitigating hallucinations~\cite{cao2023instruction,chen2023alpagasus}. To further reduce the cost of addressing hallucinations, existing efforts have focused on solutions during the inference stage. For instance, prompting LLMs or MLLMs to verify their own responses has proven effective in alleviating hallucinations~\cite{lee2023benefits}. One such method, Chain-of-Verification (CoVe)~\cite{dhuliawala2023chain}, involves the model drafting an initial response, planning verification questions based on that response, answering these questions to validate the draft, and ultimately generating an optimized answer. Experiments have demonstrated that self-verification methods, such as CoVe, can effectively reduce hallucinations across various tasks. Additionally, retrieval-augmented generation has proven effective in reducing hallucinations~\cite{shuster2021retrieval}. This approach allows the model to retrieve relevant knowledge from external webpages or knowledge bases during the response generation phase \cite{li2023chatdoctor,sun2024pathasst}, significantly addressing the hallucination problem.

Particularly, for MLLMs, the occurrence of hallucinations may be attributed to limited visual perception capabilities. For example, the Q-former uses only 32 learnable vectors to represent an image, which inevitably leads to a loss of visual information. Furthermore, the resolution of vision encoders in current medical MLLMs typically ranges from 224 to 336, hindering their ability to capture the complexities of biomedical images and contributing to hallucinations.	To tackle this challenge, integrating more advanced and higher-resolution vision encoders, such as DINOv2~\cite{oquab2023dinov2} and SAM~\cite{kirillov2023segment}, into medical MLLMs represents a viable solution~\cite{zong2024mova}.

\subsection{Training and Deployment Challenges}
Due to the need for real-time data processing and enhanced privacy protection, local deployment of private LLMs and MLLMs is essential, particularly in medical settings. However, the substantial increase in parameters for LLMs and MLLMs significantly escalates the demand for computational resources, resulting in high operational costs.For example, MEDITRON-70B requires 128 A100 GPUs for training, and the smaller LLaVA-Med 7B necessitates 8 A100 GPUs. Furthermore, even after training and fine-tuning, the large sizes of medical LLMs and MLLMs result in high deployment and inference costs, presenting considerable challenges for most hospitals in locally deploying these models and applying them in practical settings. To facilitate the training and deployment of medical LLMs and MLLMs in hospitals with limited computational resources, this subsection proposes three solutions: optimizing the training process, reducing model parameters, and modifying model architectures.	

To achieve efficient training of LLMs and MLLMs while minimizing computational overhead, researchers have proposed a series of parameter-efficient fine-tuning methods~\cite{li2021prefix,hu2021lora,houlsby2019parameter}. These approaches involve freezing the majority of parameters in LLMs and MLLMs and updating only a small subset to enable effective training. For example, Hu et al.~\cite{hu2021lora} introduced LoRA, which freezes the pre-trained model weights and injects trainable rank decomposition matrices into each layer of the Transformer architecture to facilitate efficient training. However, while PEFT methods facilitate efficient training, they do not address the challenges associated with deployment. To address this issue, reducing model parameters and designing lightweight models are viable solutions~\cite{chu2023mobilevlm,yuan2023tinygpt, zhou4988925sigphi}. For example, MobileVLM \cite{chu2023mobilevlm} is a customized MLLM for mobile scenarios, which reduces the training and inference budget by downsizing LLaMA and designing an efficient projector. It is capable of running on mobile devices while remaining competitive with other MLLMs across most tasks.

Currently, most LLMs and MLLMs are based on the Transformer architecture, which leads to a quadratic increase in computational complexity with sequence length, resulting in low efficiency for long sequences. To fundamentally address the challenges of training and deploying medical LLMs and MLLMs, selecting model architectures that are more efficient in computation and inference is a viable option \cite{peng2023rwkv,gu2023mamba}. For example, RWKV \cite{peng2023rwkv} combines the efficient parallel training of Transformers with effective inference from RNNs, ensuring constant computational and memory complexity during inference while maintaining comparable performance to similarly scaled Transformer models. Furthermore, Mamba \cite{gu2023mamba}, based on the State Space Model, outperforms Transformer models in both performance and inference speed, achieving five times the inference speed of Transformers while remaining comparable in scale. Extending these computationally and inference-efficient model architectures to medical LLMs and MLLMs will help address the current training and deployment challenges faced by these models.	

\subsection{Lack of Recency}
Once medical LLMs and MLLMs are trained, the knowledge they acquire becomes static.	However, as medical knowledge is continuously updated, the absence of new concepts will exacerbate the models' inaccuracies and hallucination problems. This is particularly evident when the models encounter new terms that are not present in the training corpus, rendering them unable to comprehend this knowledge~\cite{thirunavukarasu2023large}. For example, if medical LLMs and MLLMs are trained exclusively on data prior to 2020, they will lack information regarding COVID-19. This limitation may prevent the models from understanding terms such as "COVID-19" or "Long COVID," or they may incorrectly classify COVID-19 as a known viral variant (e.g., SARS or MERS), leading to the provision of misleading advice. Consequently, the lack of recent knowledge will significantly hinder the real-world application of medical LLMs and MLLMs.

To address the lack of recency resulting from the offline learning of medical LLMs and MLLMs, continual parameter updates through fine-tuning methods to synchronize them with current human knowledge is a feasible solution \cite{wu2024continual}. While fine-tuning can inject new medical concepts and knowledge into the model, it also introduces two challenges: catastrophic forgetting, in which the model forgets previously learned knowledge upon acquiring new information~\cite{zhai2023investigating}. The second challenge is negative forward transfer, wherein performance on unseen tasks deteriorates when learning new tasks \cite{zheng2024beyond}. To address these issues, researchers have proposed model editing \cite{yao2023editing}, one is introducing additional trainable parameters to correct erroneous responses stemming from outdated knowledge while preserving the original parameters of the model \cite{huang2023transformer,hartvigsen2024aging}. Another approach involves identifying parameters related to specific knowledge and updating them accordingly to integrate relevant new information \cite{meng2022locating,meng2022mass,li2024pmet}. In addition to model editing, retrieval-augmented generation can be employed to update the knowledge of medical LLMs and MLLMs by linking the model to an information retrieval component. This allows the model to retrieve relevant content from external knowledge bases as references  \cite{li2023chatdoctor,sun2024pathasst}, thereby generating more reliable responses.

\subsection{Privacy and Security}
Medical LLMs and MLLMs are trained on a large-scale medical corpus that includes data such as EHRs, doctor-patient dialogues, and other information that may involve patient privacy, including names, phone numbers, and email addresses. This information can potentially be directly extracted from medical LLMs or MLLMs using specific prompting methods~\cite{carlini2021extracting, li2023multi}, raising significant privacy and security concerns.

Currently, a common practice to enhance patient privacy protection is data de-identification~\cite{li2023chatdoctor,liu2023chatcounselor}. This process involves removing or anonymizing sensitive information from datasets, including names, phone numbers, addresses, and medical record numbers. Additionally, sensitive terms in clinician-patient dialogues and medical records are replaced or removed to dissociate these terms from specific patients. Moreover, differential privacy methods can effectively mitigate the risk of privacy breaches by adding noise to obscure individual information in the training data, thus preventing the inference of specific details while still enabling meaningful data analysis~\cite{turgay2023perturbation}. Furthermore, utilizing high-quality synthetic data generated by models such as ChatGPT or GPT-4 during training ensures both the controllability and diversity of the datasets while mitigating the risk of privacy leaks. It is also advisable to monitor and filter the model’s outputs. For instance, if the output contains sensitive data such as names or contact information, this information should be removed or modified during post-processing. Finally, we urge developers to adhere to ethical standards, ensuring that models respect privacy rights when handling patient data and provide patients with the right to access and delete their information.	

\begin{figure}[ht]
  \centering
  \includegraphics[width=\linewidth]{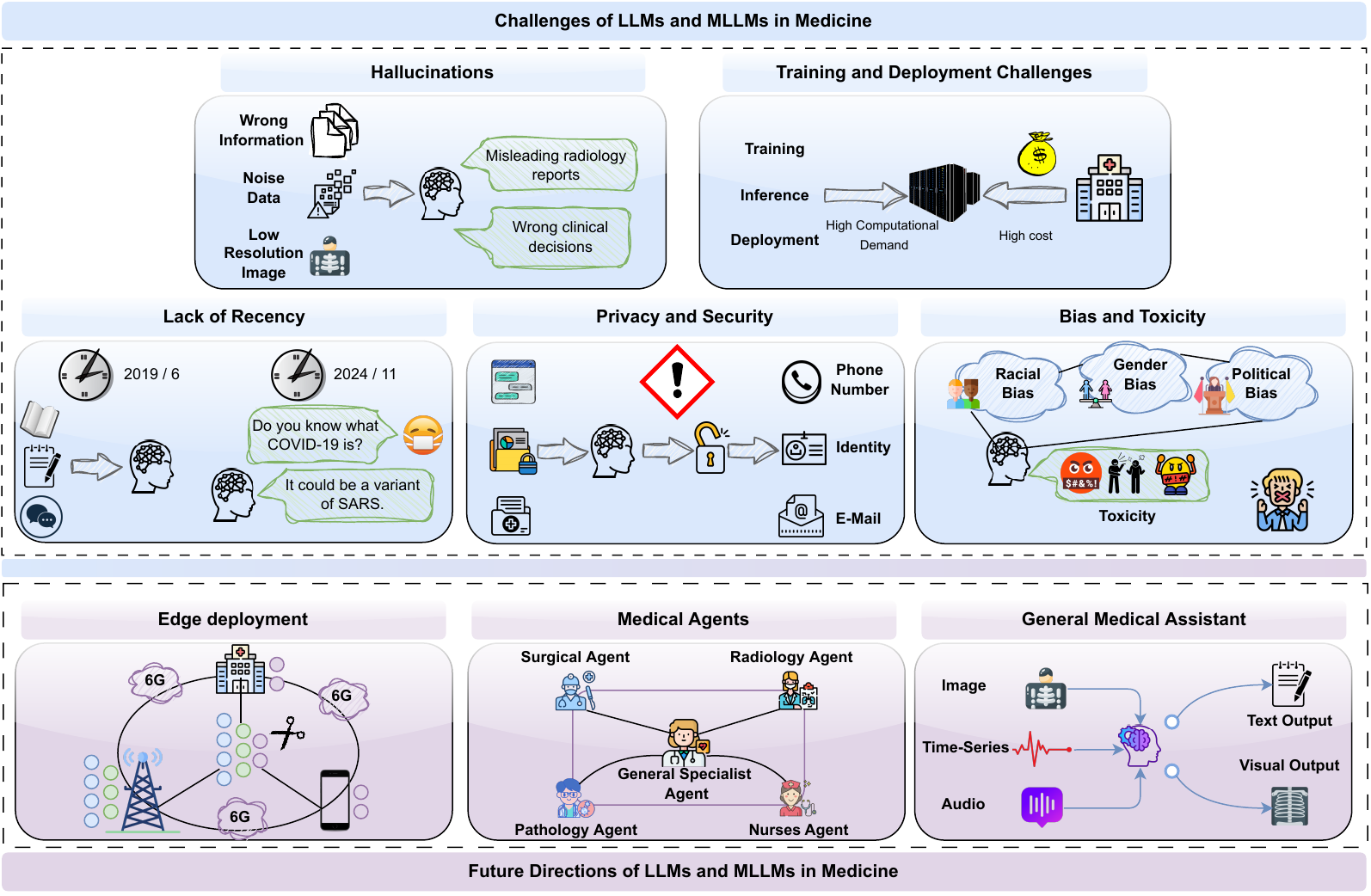}
  \caption{Overview of challenges and future directions for medical LLMs and MLLMs in clinical settings.}
  \label{fig10}
\end{figure}

\subsection{Bias and Toxicity}
Large-scale corpora, particularly data sourced from the internet, inevitably contain various biased viewpoints, which LLMs and MLLMs may learn \cite{qiu2023large,ferrara2023should}, including biases related to race \cite{yang2024unmasking}, gender \cite{kotek2023gender}, and politics \cite{liu2022quantifying}. Additionally, language models may generate toxic responses, such as aggressive and hurtful remarks, with certain groups being more likely to be targeted due to these biases~\cite{deshpande2023toxicity}. These biases and toxicities extend to LLMs and MLLMs, posing potential implications and threats to patients, and may have serious consequences for individuals with mental illness.	

Reducing bias in training data is a fundamental approach to mitigating bias in models. Specifically, careful curation and screening of diverse, balanced, and representative training data ensure that models learn from a broader range of perspectives and experiences, leading to a more comprehensive understanding and reduced biases across various dimensions \cite{ferrara2023should}. For addressing model toxicity, utilizing empathetic data has been shown to decrease the output of toxic content from models~\cite{lahnala2022mitigating}. However, re-screening pre-training datasets and retraining models to minimize biases and toxicities can be costly. Therefore, screening high-quality datasets with anti-bias and anti-toxicity for fine-tuning is a more cost-effective approach. In addition to training, further enhancement is needed in evaluating model bias and toxicity. Designing a comprehensive benchmark for evaluating model bias and toxicity facilitates the detection of these issues, allowing developers to conduct regular reviews of the models~\cite{ferrara2023should}.

\section{Future Directions of LLMs and MLLMs in Medicine}
This section examines future development trends in medical LLMs and MLLMs, envisioning their improved integration into clinical settings while offering forward-looking and insightful perspectives.	

\label{Sec7}
\subsection{Edge Deployment}
Due to their substantial computational and storage demands, existing LLMs and MLLMs are predominantly deployed on cloud servers. However, cloud-based deployment faces notable limitations in medical environments. First, in regions with poor network connectivity, communication latency with the cloud hinders the use of medical LLMs and MLLMs in critical clinical scenarios, such as surgical assistance. Second, transmitting data to cloud servers for cloud-based LLMs and MLLMs raises significant concerns about data privacy breaches. To address these challenges, deploying medical LLMs and MLLMs on edge devices is gaining traction as a promising future direction.

Edge deployment offers enhanced patient privacy protection and enables real-time responses~\cite{xu2024unleashing}. Nevertheless, the limited computational capacity and memory of edge devices present significant challenges for deploying medical LLMs and MLLMs in such environments. Lin et al.~\cite{lin2023pushing} proposed a Mobile Edge Computing (MEC) architecture leveraging 6G networks to coordinate tasks between edge devices and servers as a solution to this challenge. For example, in scenarios demanding high privacy, complete device-side inference can be utilized to ensure full patient privacy protection.	In more complex scenarios, this architecture supports dynamic adjustments of model splitting and task allocation. For instance, in surgical assistance, the MEC architecture can deploy lightweight model components on edge devices to enable real-time responses, while offloading computationally intensive tasks to nearby edge servers. This setup supports distributed collaborative inference through 6G communication, reducing the burden on edge devices.

\subsection{Medical Agents}
Clinical diagnosis is a multidisciplinary process requiring collaboration among experts from fields such as pathology, radiology, and surgery. LLM- and MLLM-based agents can simulate departmental experts, facilitating team collaboration to deliver comprehensive and accurate diagnostic support~\cite{zhou2023survey}.

For example, when handling a complex medical case, a general practitioner agent can decompose the task into clear sub-tasks based on the patient’s specific conditions and requirements, dynamically assigning them to the most suitable expert agents. Each expert agent provides analysis specific to their assigned sub-tasks. A radiology agent analyzes imaging studies to generate a radiology report, a surgical agent develops a surgical plan, and a pathology agent examines tissue samples or blood test results to provide cellular- or molecular-level diagnostic insights. After each agent completes their assigned sub-tasks, a collaborative mechanism consolidates their diagnostic inputs into a comprehensive medical opinion, which is then communicated back to the general practitioner agent~\cite{kim2024adaptive,tang2023medagents}. Additionally, during agent collaboration, shared data and insights contribute to improved overall diagnostic recommendations. However, effective collaboration among multiple agents requires low-latency communication, especially in urgent scenarios like surgical assistance, where real-time interaction is essential. Consequently, 6G communication technology can also offer vital technical support for enabling seamless multi-agent collaboration.	

\subsection{General Medical Assistant}
Although medical MLLMs currently support visual and text modalities, enabling tasks such as medical image report generation and visual question answering, clinical environments demand these models to process additional input modalities, including time series and audio data~\cite{chan2024medtsllm} and audio data. These additional input modalities can provide more comprehensive patient information, particularly playing a crucial role in dynamic monitoring and real-time diagnosis.

At present, medical LLMs and MLLMs primarily rely on static data, such as medical records and imaging studies, limiting their capacity to account for patients' dynamic data. Incorporating time series data allows medical MLLMs to analyze health trends and detect early signs of disease progression, leading to more precise alerts and predictions. For example, in an ICU, in an ICU, medical MLLMs can integrate continuous ECG monitoring and blood pressure variations to predict cardiac arrest risks and recommend timely interventions. Similarly, audio data, such as cough recordings, can aid in diagnosing lung diseases or predicting respiratory failure risks. Voice analysis by medical MLLMs can also assess a patient’s emotional state, supporting psychiatrists in diagnosing mental health conditions like depression and anxiety~\cite{hu2024parallel, CHEN2024102017,HE202256}.

Beyond additional input modalities, medical MLLMs are also expected to support a broader range of output modalities. Currently, most medical MLLMs rely on LLM components as output modules, restricting outputs to text. In clinical practice, medical MLLMs require visual output modules for region- and pixel-level outputs, such as segmenting pathological areas in images according to physician instructions to deliver more detailed diagnostic evidence.	

\section{Conclusion}
\label{Sec8}
In recent years, advancements in LLMs have driven significant breakthroughs in NLP, enabling researchers to make substantial progress toward artificial general intelligence by extending LLMs into the multimodal domain, resulting in the creation of MLLMs. Concurrently, the rapid development and impressive performance of general LLMs and MLLMs have spurred the emergence of numerous medical LLMs and MLLMs. This survey aims to help researchers and medical practitioners understand the technological advancements and developmental status of medical LLMs and MLLMs. It focuses on the paradigm shift of LLMs and MLLMs, highlighting their evolution from feature engineering to structure engineering, objective engineering, and now to prompt engineering and data engineering. The survey summarizes the mainstream architectures of current LLMs and MLLMs, compiles a list of existing medical LLMs and MLLMs, and provides insights into various architectures and model components. Additionally, it presents a comprehensive guide on existing medical datasets, model fine-tuning approaches, and evaluation methods, aiding researchers and practitioners in building medical LLMs and MLLMs. The survey also examines the broad potential of LLMs and MLLMs in diverse clinical applications, such as medical diagnosis, clinical report generation, medical education, mental health services, and surgical assistance. Despite their remarkable achievements, medical LLMs and MLLMs face several critical challenges and limitations that hinder their practical deployment in clinical settings. Consequently, this survey addresses these challenges, including hallucinations, training and deployment issues, lack of recency, privacy and security, bias and toxicity, while offering potential solutions to support the practical application of future medical LLMs and MLLMs. Lastly, the survey explores future directions for medical LLMs and MLLMs, such as edge deployment, medical agents, and general medical assistant, offering forward-looking and insightful analysis. In conclusion, this survey delivers a comprehensive analysis of medical LLMs and MLLMs, covering their background, principles, applications, challenges, and future directions, with the aim of advancing their development in clinical medicine and fostering AI integration in healthcare.


\bibliographystyle{unsrt}

\bibliography{cas-refs}

\begin{thebibliography}{100}

\bibitem{vaswani2017attention}
Ashish Vaswani, Noam Shazeer, Niki Parmar, Jakob Uszkoreit, Llion Jones, Aidan~N Gomez, {\L}ukasz Kaiser, and Illia Polosukhin.
\newblock Attention is all you need.
\newblock {\em Advances in Neural Information Processing Systems}, 30, 2017.

\bibitem{chowdhery2023palm}
Aakanksha Chowdhery, Sharan Narang, Jacob Devlin, Maarten Bosma, Gaurav Mishra, Adam Roberts, Paul Barham, Hyung~Won Chung, Charles Sutton, Sebastian Gehrmann, et~al.
\newblock Palm: Scaling language modeling with pathways.
\newblock {\em Journal of Machine Learning Research}, 24(240):1--113, 2023.

\bibitem{anil2023palm}
Rohan Anil, Andrew~M Dai, Orhan Firat, Melvin Johnson, Dmitry Lepikhin, Alexandre Passos, Siamak Shakeri, Emanuel Taropa, Paige Bailey, Zhifeng Chen, et~al.
\newblock Palm 2 technical report.
\newblock {\em arXiv preprint arXiv:2305.10403}, 2023.

\bibitem{brown2020language}
Tom Brown, Benjamin Mann, Nick Ryder, Melanie Subbiah, Jared~D Kaplan, Prafulla Dhariwal, Arvind Neelakantan, Pranav Shyam, Girish Sastry, Amanda Askell, et~al.
\newblock Language models are few-shot learners.
\newblock {\em Advances in Neural Information Processing Systems}, 33:1877--1901, 2020.

\bibitem{ouyang2022training}
Long Ouyang, Jeffrey Wu, Xu~Jiang, Diogo Almeida, Carroll Wainwright, Pamela Mishkin, Chong Zhang, Sandhini Agarwal, Katarina Slama, Alex Ray, et~al.
\newblock Training language models to follow instructions with human feedback.
\newblock {\em Advances in Neural Information Processing Systems}, 35:27730--27744, 2022.

\bibitem{touvron2023llama}
Hugo Touvron, Thibaut Lavril, Gautier Izacard, Xavier Martinet, Marie-Anne Lachaux, Timoth{\'e}e Lacroix, Baptiste Rozi{\`e}re, Naman Goyal, Eric Hambro, Faisal Azhar, et~al.
\newblock Llama: Open and efficient foundation language models.
\newblock {\em arXiv preprint arXiv:2302.13971}, 2023.

\bibitem{touvron2023llama2}
Hugo Touvron, Louis Martin, Kevin Stone, Peter Albert, Amjad Almahairi, Yasmine Babaei, Nikolay Bashlykov, Soumya Batra, Prajjwal Bhargava, Shruti Bhosale, et~al.
\newblock Llama 2: Open foundation and fine-tuned chat models.
\newblock {\em arXiv preprint arXiv:2307.09288}, 2023.

\bibitem{dubey2024llama}
Abhimanyu Dubey, Abhinav Jauhri, Abhinav Pandey, Abhishek Kadian, Ahmad Al-Dahle, Aiesha Letman, Akhil Mathur, Alan Schelten, Amy Yang, Angela Fan, et~al.
\newblock The llama 3 herd of models.
\newblock {\em arXiv preprint arXiv:2407.21783}, 2024.

\bibitem{team2023gemini}
Gemini Team, Rohan Anil, Sebastian Borgeaud, Yonghui Wu, Jean-Baptiste Alayrac, Jiahui Yu, Radu Soricut, Johan Schalkwyk, Andrew~M Dai, Anja Hauth, et~al.
\newblock Gemini: a family of highly capable multimodal models.
\newblock {\em arXiv preprint arXiv:2312.11805}, 2023.

\bibitem{achiam2023gpt}
Josh Achiam, Steven Adler, Sandhini Agarwal, Lama Ahmad, Ilge Akkaya, Florencia~Leoni Aleman, Diogo Almeida, Janko Altenschmidt, Sam Altman, Shyamal Anadkat, et~al.
\newblock Gpt-4 technical report.
\newblock {\em arXiv preprint arXiv:2303.08774}, 2023.

\bibitem{liu2024visual}
Haotian Liu, Chunyuan Li, Qingyang Wu, and Yong~Jae Lee.
\newblock Visual instruction tuning.
\newblock {\em Advances in Neural Information Processing Systems}, 36:34892--34916, 2024.

\bibitem{thirunavukarasu2023large}
Arun~James Thirunavukarasu, Darren Shu~Jeng Ting, Kabilan Elangovan, Laura Gutierrez, Ting~Fang Tan, and Daniel Shu~Wei Ting.
\newblock Large language models in medicine.
\newblock {\em Nature Medicine}, 29(8):1930--1940, 2023.

\bibitem{singhal2023towards}
Karan Singhal, Tao Tu, Juraj Gottweis, Rory Sayres, Ellery Wulczyn, Le~Hou, Kevin Clark, Stephen Pfohl, Heather Cole-Lewis, Darlene Neal, et~al.
\newblock Towards expert-level medical question answering with large language models.
\newblock {\em arXiv preprint arXiv:2305.09617}, 2023.

\bibitem{jin2021disease}
Di~Jin, Eileen Pan, Nassim Oufattole, Wei-Hung Weng, Hanyi Fang, and Peter Szolovits.
\newblock What disease does this patient have? a large-scale open domain question answering dataset from medical exams.
\newblock {\em Applied Sciences}, 11(14):6421, 2021.

\bibitem{zhou2023survey}
Hongjian Zhou, Boyang Gu, Xinyu Zou, Yiru Li, Sam~S Chen, Peilin Zhou, Junling Liu, Yining Hua, Chengfeng Mao, Xian Wu, et~al.
\newblock A survey of large language models in medicine: Progress, application, and challenge.
\newblock {\em arXiv preprint arXiv:2311.05112}, 2023.

\bibitem{li2023chatdoctor}
Yunxiang Li, Zihan Li, Kai Zhang, Ruilong Dan, Steve Jiang, and You Zhang.
\newblock Chatdoctor: A medical chat model fine-tuned on a large language model meta-ai (llama) using medical domain knowledge.
\newblock {\em Cureus}, 15(6), 2023.

\bibitem{wang2024interactive}
Sheng Wang, Zihao Zhao, Xi~Ouyang, Tianming Liu, Qian Wang, and Dinggang Shen.
\newblock Interactive computer-aided diagnosis on medical image using large language models.
\newblock {\em Communications Engineering}, 3(1):133, 2024.

\bibitem{li2024llava}
Chunyuan Li, Cliff Wong, Sheng Zhang, Naoto Usuyama, Haotian Liu, Jianwei Yang, Tristan Naumann, Hoifung Poon, and Jianfeng Gao.
\newblock Llava-med: Training a large language-and-vision assistant for biomedicine in one day.
\newblock {\em Advances in Neural Information Processing Systems}, 36:28541--28564, 2024.

\bibitem{van2023clinical}
Dave Van~Veen, Cara Van~Uden, Louis Blankemeier, Jean-Benoit Delbrouck, Asad Aali, Christian Bluethgen, Anuj Pareek, Malgorzata Polacin, Eduardo~Pontes Reis, Anna Seehofnerova, et~al.
\newblock Clinical text summarization: Adapting large language models can outperform human experts.
\newblock {\em Research Square}, 2023.

\bibitem{wang2023r2gengpt}
Zhanyu Wang, Lingqiao Liu, Lei Wang, and Luping Zhou.
\newblock R2gengpt: Radiology report generation with frozen llms.
\newblock {\em Meta-Radiology}, 1(3):100033, 2023.

\bibitem{tu2024towards}
Tao Tu, Shekoofeh Azizi, Danny Driess, Mike Schaekermann, Mohamed Amin, Pi-Chuan Chang, Andrew Carroll, Charles Lau, Ryutaro Tanno, Ira Ktena, et~al.
\newblock Towards generalist biomedical ai.
\newblock {\em NEJM AI}, 1(3):AIoa2300138, 2024.

\bibitem{chen2023soulchat}
Yirong Chen, Xiaofen Xing, Jingkai Lin, Huimin Zheng, Zhenyu Wang, Qi~Liu, and Xiangmin Xu.
\newblock Soulchat: Improving llms’ empathy, listening, and comfort abilities through fine-tuning with multi-turn empathy conversations.
\newblock In {\em Findings of the Association for Computational Linguistics: EMNLP 2023}, pages 1170--1183, dec 2023.

\bibitem{liu2023chatcounselor}
June~M Liu, Donghao Li, He~Cao, Tianhe Ren, Zeyi Liao, and Jiamin Wu.
\newblock Chatcounselor: A large language models for mental health support.
\newblock {\em arXiv preprint arXiv:2309.15461}, 2023.

\bibitem{zhang2024data}
Yunkun Zhang, Jin Gao, Zheling Tan, Lingfeng Zhou, Kexin Ding, Mu~Zhou, Shaoting Zhang, and Dequan Wang.
\newblock Data-centric foundation models in computational healthcare: A survey.
\newblock {\em arXiv preprint arXiv:2401.02458}, 2024.

\bibitem{moor2023foundation}
Michael Moor, Oishi Banerjee, Zahra Shakeri~Hossein Abad, Harlan~M Krumholz, Jure Leskovec, Eric~J Topol, and Pranav Rajpurkar.
\newblock Foundation models for generalist medical artificial intelligence.
\newblock {\em Nature}, 616(7956):259--265, 2023.

\bibitem{qiu2023large}
Jianing Qiu, Lin Li, Jiankai Sun, Jiachuan Peng, Peilun Shi, Ruiyang Zhang, Yinzhao Dong, Kyle Lam, Frank P-W Lo, Bo~Xiao, et~al.
\newblock Large ai models in health informatics: Applications, challenges, and the future.
\newblock {\em IEEE Journal of Biomedical and Health Informatics}, 27(12):6074--6087, 2023.

\bibitem{zhang2023instruction}
Shengyu Zhang, Linfeng Dong, Xiaoya Li, Sen Zhang, Xiaofei Sun, Shuhe Wang, Jiwei Li, Runyi Hu, Tianwei Zhang, Fei Wu, et~al.
\newblock Instruction tuning for large language models: A survey.
\newblock {\em arXiv preprint arXiv:2308.10792}, 2023.

\bibitem{umapathi2023med}
Logesh~Kumar Umapathi, Ankit Pal, and Malaikannan Sankarasubbu.
\newblock Med-halt: Medical domain hallucination test for large language models.
\newblock {\em arXiv preprint arXiv:2307.15343}, 2023.

\bibitem{rawte2023survey}
Vipula Rawte, Amit Sheth, and Amitava Das.
\newblock A survey of hallucination in large foundation models.
\newblock {\em arXiv preprint arXiv:2309.05922}, 2023.

\bibitem{ji2023survey}
Ziwei Ji, Nayeon Lee, Rita Frieske, Tiezheng Yu, Dan Su, Yan Xu, Etsuko Ishii, Ye~Jin Bang, Andrea Madotto, and Pascale Fung.
\newblock Survey of hallucination in natural language generation.
\newblock {\em ACM Computing Surveys}, 55(12):1--38, 2023.

\bibitem{LIN2024102795}
Qika Lin, Yifan Zhu, Xin Mei, Ling Huang, Jingying Ma, Kai He, Zhen Peng, Erik Cambria, and Mengling Feng.
\newblock Has multimodal learning delivered universal intelligence in healthcare? a comprehensive survey.
\newblock {\em Information Fusion}, page 102795, 2024.

\bibitem{KRONES2025102690}
Felix Krones, Umar Marikkar, Guy Parsons, Adam Szmul, and Adam Mahdi.
\newblock Review of multimodal machine learning approaches in healthcare.
\newblock {\em Information Fusion}, 114:102690, 2025.

\bibitem{he2023survey}
Kai He, Rui Mao, Qika Lin, Yucheng Ruan, Xiang Lan, Mengling Feng, and Erik Cambria.
\newblock A survey of large language models for healthcare: from data, technology, and applications to accountability and ethics.
\newblock {\em arXiv preprint arXiv:2310.05694}, 2023.

\bibitem{wang2023pre}
Benyou Wang, Qianqian Xie, Jiahuan Pei, Zhihong Chen, Prayag Tiwari, Zhao Li, and Jie Fu.
\newblock Pre-trained language models in biomedical domain: A systematic survey.
\newblock {\em ACM Computing Surveys}, 56(3):1--52, 2023.

\bibitem{thapa2023chatgpt}
Surendrabikram Thapa and Surabhi Adhikari.
\newblock Chatgpt, bard, and large language models for biomedical research: opportunities and pitfalls.
\newblock {\em Annals of Biomedical Engineering}, 51(12):2647--2651, 2023.

\bibitem{omiye2024large}
Jesutofunmi~A Omiye, Haiwen Gui, Shawheen~J Rezaei, James Zou, and Roxana Daneshjou.
\newblock Large language models in medicine: the potentials and pitfalls: a narrative review.
\newblock {\em Annals of Internal Medicine}, 177(2):210--220, 2024.

\bibitem{bhayana2024chatbots}
Rajesh Bhayana.
\newblock Chatbots and large language models in radiology: A practical primer for clinical and research applications.
\newblock {\em Radiology}, 310(1):e232756, 2024.

\bibitem{zhou2024lima}
Chunting Zhou, Pengfei Liu, Puxin Xu, Srinivasan Iyer, Jiao Sun, Yuning Mao, Xuezhe Ma, Avia Efrat, Ping Yu, Lili Yu, et~al.
\newblock Lima: Less is more for alignment.
\newblock {\em Advances in Neural Information Processing Systems}, 36:55006--55021, 2024.

\bibitem{liu2023pre}
Pengfei Liu, Weizhe Yuan, Jinlan Fu, Zhengbao Jiang, Hiroaki Hayashi, and Graham Neubig.
\newblock Pre-train, prompt, and predict: A systematic survey of prompting methods in natural language processing.
\newblock {\em ACM Computing Surveys}, 55(9):1--35, 2023.

\bibitem{liu2005toward}
Huan Liu and Lei Yu.
\newblock Toward integrating feature selection algorithms for classification and clustering.
\newblock {\em IEEE Transactions on Knowledge and Data Engineering}, 17(4):491--502, 2005.

\bibitem{och2004smorgasbord}
Franz~Josef Och, Daniel Gildea, Sanjeev Khudanpur, Anoop Sarkar, Kenji Yamada, Alexander Fraser, Shankar Kumar, Libin Shen, David~A Smith, Katherine Eng, et~al.
\newblock A smorgasbord of features for statistical machine translation.
\newblock In {\em Proceedings of the Human Language Technology Conference of the North American Chapter of the Association for Computational Linguistics: HLT-NAACL 2004}, pages 161--168, 2004.

\bibitem{lecun2015deep}
Yann LeCun, Yoshua Bengio, and Geoffrey Hinton.
\newblock Deep learning.
\newblock {\em Nature}, 521(7553):436--444, 2015.

\bibitem{radford2018improving}
Alec Radford, Karthik Narasimhan, Tim Salimans, Ilya Sutskever, et~al.
\newblock Improving language understanding by generative pre-training.
\newblock 2018.

\bibitem{devlin2018bert}
Jacob Devlin, Ming-Wei Chang, Kenton Lee, and Kristina Toutanova.
\newblock Bert: Pre-training of deep bidirectional transformers for language understanding.
\newblock {\em arXiv preprint arXiv:1810.04805}, 2018.

\bibitem{gu2021domain}
Yu~Gu, Robert Tinn, Hao Cheng, Michael Lucas, Naoto Usuyama, Xiaodong Liu, Tristan Naumann, Jianfeng Gao, and Hoifung Poon.
\newblock Domain-specific language model pretraining for biomedical natural language processing.
\newblock {\em ACM Transactions on Computing for Healthcare (HEALTH)}, 3(1):1--23, 2021.

\bibitem{lee2020biobert}
Jinhyuk Lee, Wonjin Yoon, Sungdong Kim, Donghyeon Kim, Sunkyu Kim, Chan~Ho So, and Jaewoo Kang.
\newblock Biobert: a pre-trained biomedical language representation model for biomedical text mining.
\newblock {\em Bioinformatics}, 36(4):1234--1240, 2020.

\bibitem{radford2019language}
Alec Radford, Jeffrey Wu, Rewon Child, David Luan, Dario Amodei, Ilya Sutskever, et~al.
\newblock Language models are unsupervised multitask learners.
\newblock {\em OpenAI blog}, 1(8):9, 2019.

\bibitem{mesko2023prompt}
Bertalan Mesk{\'o}.
\newblock Prompt engineering as an important emerging skill for medical professionals: tutorial.
\newblock {\em Journal of Medical Internet Research}, 25:e50638, 2023.

\bibitem{zeng2022glm}
Aohan Zeng, Xiao Liu, Zhengxiao Du, Zihan Wang, Hanyu Lai, Ming Ding, Zhuoyi Yang, Yifan Xu, Wendi Zheng, Xiao Xia, et~al.
\newblock Glm-130b: An open bilingual pre-trained model.
\newblock {\em arXiv preprint arXiv:2210.02414}, 2022.

\bibitem{zellers2021merlot}
Rowan Zellers, Ximing Lu, Jack Hessel, Youngjae Yu, Jae~Sung Park, Jize Cao, Ali Farhadi, and Yejin Choi.
\newblock Merlot: Multimodal neural script knowledge models.
\newblock {\em Advances in Neural Information Processing Systems}, 34:23634--23651, 2021.

\bibitem{hendricks2021decoupling}
Lisa~Anne Hendricks, John Mellor, Rosalia Schneider, Jean-Baptiste Alayrac, and Aida Nematzadeh.
\newblock Decoupling the role of data, attention, and losses in multimodal transformers.
\newblock {\em Transactions of the Association for Computational Linguistics}, 9:570--585, 2021.

\bibitem{radford2021learning}
Alec Radford, Jong~Wook Kim, Chris Hallacy, Aditya Ramesh, Gabriel Goh, Sandhini Agarwal, Girish Sastry, Amanda Askell, Pamela Mishkin, Jack Clark, Gretchen Krueger, and Ilya Sutskever.
\newblock Learning transferable visual models from natural language supervision.
\newblock In {\em Proceedings of the 38th International Conference on Machine Learning}, volume 139 of {\em Proceedings of Machine Learning Research}, pages 8748--8763. PMLR, 18--24 Jul 2021.

\bibitem{li2021align}
Junnan Li, Ramprasaath Selvaraju, Akhilesh Gotmare, Shafiq Joty, Caiming Xiong, and Steven Chu~Hong Hoi.
\newblock Align before fuse: Vision and language representation learning with momentum distillation.
\newblock {\em Advances in Neural Information Processing Systems}, 34:9694--9705, 2021.

\bibitem{alayrac2022flamingo}
Jean-Baptiste Alayrac, Jeff Donahue, Pauline Luc, Antoine Miech, Iain Barr, Yana Hasson, Karel Lenc, Arthur Mensch, Katherine Millican, Malcolm Reynolds, et~al.
\newblock Flamingo: a visual language model for few-shot learning.
\newblock {\em Advances in Neural Information Processing Systems}, 35:23716--23736, 2022.

\bibitem{zhang2024mm}
Duzhen Zhang, Yahan Yu, Jiahua Dong, Chenxing Li, Dan Su, Chenhui Chu, and Dong Yu.
\newblock {MM}-{LLM}s: Recent advances in {M}ulti{M}odal large language models.
\newblock In {\em Findings of the Association for Computational Linguistics: ACL 2024}, pages 12401--12430. Association for Computational Linguistics, August 2024.

\bibitem{xie2024medtrinity}
Yunfei Xie, Ce~Zhou, Lang Gao, Juncheng Wu, Xianhang Li, Hong-Yu Zhou, Sheng Liu, Lei Xing, James Zou, Cihang Xie, et~al.
\newblock Medtrinity-25m: A large-scale multimodal dataset with multigranular annotations for medicine.
\newblock {\em arXiv preprint arXiv:2408.02900}, 2024.

\bibitem{alpaca}
Rohan Taori, Ishaan Gulrajani, Tianyi Zhang, Yann Dubois, Xuechen Li, Carlos Guestrin, Percy Liang, and Tatsunori~B. Hashimoto.
\newblock Stanford alpaca: An instruction-following llama model.
\newblock \url{https://github.com/tatsu-lab/stanford_alpaca}, 2023.

\bibitem{raffel2020exploring}
Colin Raffel, Noam Shazeer, Adam Roberts, Katherine Lee, Sharan Narang, Michael Matena, Yanqi Zhou, Wei Li, and Peter~J Liu.
\newblock Exploring the limits of transfer learning with a unified text-to-text transformer.
\newblock {\em Journal of Machine Learning Research}, 21(140):1--67, 2020.

\bibitem{yin2023survey}
Shukang Yin, Chaoyou Fu, Sirui Zhao, Ke~Li, Xing Sun, Tong Xu, and Enhong Chen.
\newblock A survey on multimodal large language models.
\newblock {\em National Science Review}, page nwae403, 11 2024.

\bibitem{singhal2023large}
Karan Singhal, Shekoofeh Azizi, Tao Tu, S~Sara Mahdavi, Jason Wei, Hyung~Won Chung, Nathan Scales, Ajay Tanwani, Heather Cole-Lewis, Stephen Pfohl, et~al.
\newblock Large language models encode clinical knowledge.
\newblock {\em Nature}, 620(7972):172--180, 2023.

\bibitem{xu2023baize}
Canwen Xu, Daya Guo, Nan Duan, and Julian McAuley.
\newblock Baize: An open-source chat model with parameter-efficient tuning on self-chat data.
\newblock In {\em Proceedings of the 2023 Conference on Empirical Methods in Natural Language Processing}, pages 6268--6278. Association for Computational Linguistics, December 2023.

\bibitem{wang2023huatuo}
Haochun Wang, Chi Liu, Nuwa Xi, Zewen Qiang, Sendong Zhao, Bing Qin, and Ting Liu.
\newblock Huatuo: Tuning llama model with chinese medical knowledge.
\newblock {\em arXiv preprint arXiv:2304.06975}, 2023.

\bibitem{han2023medalpaca}
Tianyu Han, Lisa~C Adams, Jens-Michalis Papaioannou, Paul Grundmann, Tom Oberhauser, Alexander L{\"o}ser, Daniel Truhn, and Keno~K Bressem.
\newblock Medalpaca--an open-source collection of medical conversational ai models and training data.
\newblock {\em arXiv preprint arXiv:2304.08247}, 2023.

\bibitem{wu2024pmc}
Chaoyi Wu, Weixiong Lin, Xiaoman Zhang, Ya~Zhang, Weidi Xie, and Yanfeng Wang.
\newblock Pmc-llama: toward building open-source language models for medicine.
\newblock {\em Journal of the American Medical Informatics Association}, 31:1833--1843, 2024.

\bibitem{toma2023clinical}
Augustin Toma, Patrick~R Lawler, Jimmy Ba, Rahul~G Krishnan, Barry~B Rubin, and Bo~Wang.
\newblock Clinical camel: An open-source expert-level medical language model with dialogue-based knowledge encoding.
\newblock {\em arXiv preprint arXiv:2305.12031}, 2023.

\bibitem{zhang2023huatuogpt}
Hongbo Zhang, Junying Chen, Feng Jiang, Fei Yu, Zhihong Chen, Guiming Chen, Jianquan Li, Xiangbo Wu, Zhang Zhiyi, Qingying Xiao, Xiang Wan, Benyou Wang, and Haizhou Li.
\newblock {H}uatuo{GPT}, towards taming language model to be a doctor.
\newblock In {\em Findings of the Association for Computational Linguistics: EMNLP 2023}, pages 10859--10885. Association for Computational Linguistics, December 2023.

\bibitem{peng2023study}
Cheng Peng, Xi~Yang, Aokun Chen, Kaleb~E Smith, Nima PourNejatian, Anthony~B Costa, Cheryl Martin, Mona~G Flores, Ying Zhang, Tanja Magoc, et~al.
\newblock A study of generative large language model for medical research and healthcare.
\newblock {\em NPJ Digital Medicine}, 6(1):210, 2023.

\bibitem{wang2023clinicalgpt}
Guangyu Wang, Guoxing Yang, Zongxin Du, Longjun Fan, and Xiaohu Li.
\newblock Clinicalgpt: large language models finetuned with diverse medical data and comprehensive evaluation.
\newblock {\em arXiv preprint arXiv:2306.09968}, 2023.

\bibitem{yang2024zhongjing}
Songhua Yang, Hanjie Zhao, Senbin Zhu, Guangyu Zhou, Hongfei Xu, Yuxiang Jia, and Hongying Zan.
\newblock Zhongjing: Enhancing the chinese medical capabilities of large language model through expert feedback and real-world multi-turn dialogue.
\newblock In {\em Proceedings of the AAAI Conference on Artificial Intelligence}, volume~38, pages 19368--19376, 2024.

\bibitem{liu2023radiology}
Zhengliang Liu, Yiwei Li, Peng Shu, Aoxiao Zhong, Longtao Yang, Chao Ju, Zihao Wu, Chong Ma, Jie Luo, Cheng Chen, et~al.
\newblock Radiology-llama2: Best-in-class large language model for radiology.
\newblock {\em arXiv preprint arXiv:2309.06419}, 2023.

\bibitem{tan2023medchatzh}
Yang Tan, Mingchen Li, Zijie Huang, Huiqun Yu, and Guisheng Fan.
\newblock Medchatzh: a better medical adviser learns from better instructions.
\newblock {\em arXiv preprint arXiv:2309.01114}, 2023.

\bibitem{shoham2023cpllm}
Ofir~Ben Shoham and Nadav Rappoport.
\newblock Cpllm: Clinical prediction with large language models.
\newblock {\em arXiv preprint arXiv:2309.11295}, 2023.

\bibitem{ye2023qilin}
Qichen Ye, Junling Liu, Dading Chong, Peilin Zhou, Yining Hua, and Andrew Liu.
\newblock Qilin-med: Multi-stage knowledge injection advanced medical large language model.
\newblock {\em arXiv preprint arXiv:2310.09089}, 2023.

\bibitem{zhang2023alpacare}
Xinlu Zhang, Chenxin Tian, Xianjun Yang, Lichang Chen, Zekun Li, and Linda~Ruth Petzold.
\newblock Alpacare: Instruction-tuned large language models for medical application.
\newblock {\em arXiv preprint arXiv:2310.14558}, 2023.

\bibitem{yang2024tcm}
Guoxing Yang, Xiaohong Liu, Jianyu Shi, Zan Wang, and Guangyu Wang.
\newblock Tcm-gpt: Efficient pre-training of large language models for domain adaptation in traditional chinese medicine.
\newblock {\em Computer Methods and Programs in Biomedicine Update}, 6:100158, 2024.

\bibitem{chen2023huatuogpt}
Junying Chen, Xidong Wang, Ke~Ji, Anningzhe Gao, Feng Jiang, Shunian Chen, Hongbo Zhang, Song Dingjie, Wenya Xie, Chuyi Kong, Jianquan Li, Xiang Wan, Haizhou Li, and Benyou Wang.
\newblock Huatuo{GPT}-{II}, one-stage training for medical adaption of {LLM}s.
\newblock In {\em First Conference on Language Modeling}, 2024.

\bibitem{chen2023meditron}
Zeming Chen, Alejandro~Hern{\'a}ndez Cano, Angelika Romanou, Antoine Bonnet, Kyle Matoba, Francesco Salvi, Matteo Pagliardini, Simin Fan, Andreas K{\"o}pf, Amirkeivan Mohtashami, et~al.
\newblock Meditron-70b: Scaling medical pretraining for large language models.
\newblock {\em arXiv preprint arXiv:2311.16079}, 2023.

\bibitem{labrak2024biomistral}
Yanis Labrak, Adrien Bazoge, Emmanuel Morin, Pierre-Antoine Gourraud, Mickael Rouvier, and Richard Dufour.
\newblock Biomistral: A collection of open-source pretrained large language models for medical domains.
\newblock {\em arXiv preprint arXiv:2402.10373}, 2024.

\bibitem{xie2024me}
Qianqian Xie, Qingyu Chen, Aokun Chen, Cheng Peng, Yan Hu, Fongci Lin, Xueqing Peng, Jimin Huang, Jeffrey Zhang, Vipina Keloth, et~al.
\newblock Me llama: Foundation large language models for medical applications.
\newblock {\em arXiv preprint arXiv:2402.12749}, 2024.

\bibitem{wang2024apollo}
Xidong Wang, Nuo Chen, Junyin Chen, Yan Hu, Yidong Wang, Xiangbo Wu, Anningzhe Gao, Xiang Wan, Haizhou Li, and Benyou Wang.
\newblock Apollo: Lightweight multilingual medical llms towards democratizing medical ai to 6b people.
\newblock {\em arXiv preprint arXiv:2403.03640}, 2024.

\bibitem{bolton2024biomedlm}
Elliot Bolton, Abhinav Venigalla, Michihiro Yasunaga, David Hall, Betty Xiong, Tony Lee, Roxana Daneshjou, Jonathan Frankle, Percy Liang, Michael Carbin, et~al.
\newblock Biomedlm: A 2.7 b parameter language model trained on biomedical text.
\newblock {\em arXiv preprint arXiv:2403.18421}, 2024.

\bibitem{yang2024pediatricsgpt}
Dingkang Yang, Jinjie Wei, Dongling Xiao, Shunli Wang, Tong Wu, Gang Li, Mingcheng Li, Shuaibing Wang, Jiawei Chen, Yue Jiang, Qingyao Xu, Ke~Li, Peng Zhai, and Lihua Zhang.
\newblock Pediatrics{GPT}: Large language models as chinese medical assistants for pediatric applications.
\newblock In {\em The Thirty-eighth Annual Conference on Neural Information Processing Systems}, 2024.

\bibitem{xiong2023doctorglm}
Honglin Xiong, Sheng Wang, Yitao Zhu, Zihao Zhao, Yuxiao Liu, Linlin Huang, Qian Wang, and Dinggang Shen.
\newblock Doctorglm: Fine-tuning your chinese doctor is not a herculean task.
\newblock {\em arXiv preprint arXiv:2304.01097}, 2023.

\bibitem{chen2023bianque}
Yirong Chen, Zhenyu Wang, Xiaofen Xing, Zhipei Xu, Kai Fang, Junhong Wang, Sihang Li, Jieling Wu, Qi~Liu, Xiangmin Xu, et~al.
\newblock Bianque: Balancing the questioning and suggestion ability of health llms with multi-turn health conversations polished by chatgpt.
\newblock {\em arXiv preprint arXiv:2310.15896}, 2023.

\bibitem{he2020deberta}
Pengcheng He, Xiaodong Liu, Jianfeng Gao, and Weizhu Chen.
\newblock Deberta: Decoding-enhanced bert with disentangled attention.
\newblock {\em arXiv preprint arXiv:2006.03654}, 2020.

\bibitem{lan2019albert}
Zhenzhong Lan, Mingda Chen, Sebastian Goodman, Kevin Gimpel, Piyush Sharma, and Radu Soricut.
\newblock Albert: A lite bert for self-supervised learning of language representations.
\newblock {\em arXiv preprint arXiv:1909.11942}, 2019.

\bibitem{liu2019roberta}
Yinhan Liu, Myle Ott, Naman Goyal, Jingfei Du, Mandar Joshi, Danqi Chen, Omer Levy, Mike Lewis, Luke Zettlemoyer, and Veselin Stoyanov.
\newblock Roberta: A robustly optimized bert pretraining approach.
\newblock {\em arXiv preprint arXiv:1907.11692}, 2019.

\bibitem{ji2022mentalbert}
Shaoxiong Ji, Tianlin Zhang, Luna Ansari, Jie Fu, Prayag Tiwari, and Erik Cambria.
\newblock {M}ental{BERT}: Publicly available pretrained language models for mental healthcare.
\newblock In {\em Proceedings of the Thirteenth Language Resources and Evaluation Conference}, pages 7184--7190. European Language Resources Association, June 2022.

\bibitem{wang2022language}
Thomas Wang, Adam Roberts, Daniel Hesslow, Teven~Le Scao, Hyung~Won Chung, Iz~Beltagy, Julien Launay, and Colin Raffel.
\newblock What language model architecture and pretraining objective works best for zero-shot generalization?
\newblock In Kamalika Chaudhuri, Stefanie Jegelka, Le~Song, Csaba Szepesvari, Gang Niu, and Sivan Sabato, editors, {\em Proceedings of the 39th International Conference on Machine Learning}, volume 162, pages 22964--22984. PMLR, 17--23 Jul 2022.

\bibitem{dai2022can}
Damai Dai, Yutao Sun, Li~Dong, Yaru Hao, Shuming Ma, Zhifang Sui, and Furu Wei.
\newblock Why can {GPT} learn in-context? language models secretly perform gradient descent as meta-optimizers.
\newblock In {\em Findings of the Association for Computational Linguistics: ACL 2023}, pages 4005--4019. Association for Computational Linguistics, July 2023.

\bibitem{du2021glm}
Zhengxiao Du, Yujie Qian, Xiao Liu, Ming Ding, Jiezhong Qiu, Zhilin Yang, and Jie Tang.
\newblock {GLM}: General language model pretraining with autoregressive blank infilling.
\newblock In {\em Proceedings of the 60th Annual Meeting of the Association for Computational Linguistics}, pages 320--335. Association for Computational Linguistics, May 2022.

\bibitem{shu2023visual}
Chang Shu, Baian Chen, Fangyu Liu, Zihao Fu, Ehsan Shareghi, and Nigel Collier.
\newblock Visual med-alpaca: A parameter-efficient biomedical llm with visual capabilities.
\newblock \url{https://github.com/cambridgeltl/visual-med-alpaca}, 2023.

\bibitem{he2016deep}
Kaiming He, Xiangyu Zhang, Shaoqing Ren, and Jian Sun.
\newblock Deep residual learning for image recognition.
\newblock In {\em 2016 IEEE Conference on Computer Vision and Pattern Recognition}, pages 770--778, 2016.

\bibitem{dosovitskiy2020image}
Alexey Dosovitskiy, Lucas Beyer, Alexander Kolesnikov, Dirk Weissenborn, Xiaohua Zhai, Thomas Unterthiner, Mostafa Dehghani, Matthias Minderer, Georg Heigold, Sylvain Gelly, Jakob Uszkoreit, and Neil Houlsby.
\newblock An image is worth 16x16 words: Transformers for image recognition at scale.
\newblock In {\em International Conference on Learning Representations}, 2021.

\bibitem{chen2023pali3}
Xi~Chen, Xiao Wang, Lucas Beyer, Alexander Kolesnikov, Jialin Wu, Paul Voigtlaender, Basil Mustafa, Sebastian Goodman, Ibrahim Alabdulmohsin, Piotr Padlewski, et~al.
\newblock Pali-3 vision language models: Smaller, faster, stronger.
\newblock {\em arXiv preprint arXiv:2310.09199}, 2023.

\bibitem{hyland2023maira}
Shaury Srivastav, Mercy Ranjit, Fernando P{\'e}rez-Garc{\'\i}a, Kenza Bouzid, Shruthi Bannur, Daniel~C. Castro, Anton Schwaighofer, Harshita Sharma, Maximilian Ilse, Valentina Salvatelli, Sam Bond-Taylor, Fabian Falck, Anja Thieme, Hannah Richardson, Matthew~P. Lungren, Stephanie~L. Hyland, and Javier Alvarez-Valle.
\newblock {MAIRA} at {RRG}24: A specialised large multimodal model for radiology report generation.
\newblock In {\em Proceedings of the 23rd Workshop on Biomedical Natural Language Processing}, pages 597--602. Association for Computational Linguistics, August 2024.

\bibitem{lu2024multimodal}
Ming~Y Lu, Bowen Chen, Drew~FK Williamson, Richard~J Chen, Melissa Zhao, Aaron~K Chow, Kenji Ikemura, Ahrong Kim, Dimitra Pouli, Ankush Patel, et~al.
\newblock A multimodal generative ai copilot for human pathology.
\newblock {\em Nature}, 634:466--473, 2024.

\bibitem{fang2023eva}
Yuxin Fang, Wen Wang, Binhui Xie, Quan Sun, Ledell Wu, Xinggang Wang, Tiejun Huang, Xinlong Wang, and Yue Cao.
\newblock Eva: Exploring the limits of masked visual representation learning at scale.
\newblock In {\em Proceedings of the IEEE/CVF Conference on Computer Vision and Pattern Recognition}, pages 19358--19369, 2023.

\bibitem{vicuna2023}
Wei-Lin Chiang, Zhuohan Li, Zi~Lin, Ying Sheng, Zhanghao Wu, Hao Zhang, Lianmin Zheng, Siyuan Zhuang, Yonghao Zhuang, Joseph~E. Gonzalez, Ion Stoica, and Eric~P. Xing.
\newblock Vicuna: An open-source chatbot impressing gpt-4 with 90\%* chatgpt quality, March 2023.

\bibitem{jiang2023mistral}
Albert~Q Jiang, Alexandre Sablayrolles, Arthur Mensch, Chris Bamford, Devendra~Singh Chaplot, Diego de~las Casas, Florian Bressand, Gianna Lengyel, Guillaume Lample, Lucile Saulnier, et~al.
\newblock Mistral 7b.
\newblock {\em arXiv preprint arXiv:2310.06825}, 2023.

\bibitem{bai2023qwen}
Jinze Bai, Shuai Bai, Yunfei Chu, Zeyu Cui, Kai Dang, Xiaodong Deng, Yang Fan, Wenbin Ge, Yu~Han, Fei Huang, et~al.
\newblock Qwen technical report.
\newblock {\em arXiv preprint arXiv:2309.16609}, 2023.

\bibitem{moor2023med}
Michael Moor, Qian Huang, Shirley Wu, Michihiro Yasunaga, Yash Dalmia, Jure Leskovec, Cyril Zakka, Eduardo~Pontes Reis, and Pranav Rajpurkar.
\newblock Med-flamingo: a multimodal medical few-shot learner.
\newblock In {\em Proceedings of the 3rd Machine Learning for Health Symposium}, volume 225, pages 353--367. PMLR, 10 Dec 2023.

\bibitem{chen2023medblip}
Qiuhui Chen, Xinyue Hu, Zirui Wang, and Yi~Hong.
\newblock Medblip: Bootstrapping language-image pre-training from 3d medical images and texts.
\newblock {\em arXiv preprint arXiv:2305.10799}, 2023.

\bibitem{wang2023XrayGLM}
Rongsheng Wang, Yaofei Duan, Junrong Li, Patrick Pang, and Tao Tan.
\newblock Xrayglm: The first chinese medical multimodal model that chest radiographs summarization.
\newblock \url{https://github.com/WangRongsheng/XrayGLM}, 2023.

\bibitem{yang2023customizing}
Bang Yang, Asif Raza, Yuexian Zou, and Tong Zhang.
\newblock Customizing general-purpose foundation models for medical report generation.
\newblock {\em arXiv preprint arXiv:2306.05642}, 2023.

\bibitem{wu2023towards}
Chaoyi Wu, Xiaoman Zhang, Ya~Zhang, Yanfeng Wang, and Weidi Xie.
\newblock Towards generalist foundation model for radiology.
\newblock {\em arXiv preprint arXiv:2308.02463}, 2023.

\bibitem{chen2024chexagent}
Zhihong Chen, Maya Varma, Jean-Benoit Delbrouck, Magdalini Paschali, Louis Blankemeier, Dave~Van Veen, Jeya Maria~Jose Valanarasu, Alaa Youssef, Joseph~Paul Cohen, Eduardo~Pontes Reis, Emily Tsai, Andrew Johnston, Cameron Olsen, Tanishq~Mathew Abraham, Sergios Gatidis, Akshay~S Chaudhari, and Curtis Langlotz.
\newblock Chexagent: Towards a foundation model for chest x-ray interpretation.
\newblock In {\em AAAI 2024 Spring Symposium on Clinical Foundation Models}, 2024.

\bibitem{van2023open}
Tom Van~Sonsbeek, Mohammad~Mahdi Derakhshani, Ivona Najdenkoska, Cees~GM Snoek, and Marcel Worring.
\newblock Open-ended medical visual question answering through prefix tuning of language models.
\newblock In {\em International Conference on Medical Image Computing and Computer-Assisted Intervention}, pages 726--736. Springer, 2023.

\bibitem{zhang2023pmc}
Xiaoman Zhang, Chaoyi Wu, Ziheng Zhao, Weixiong Lin, Ya~Zhang, Yanfeng Wang, and Weidi Xie.
\newblock Pmc-vqa: Visual instruction tuning for medical visual question answering.
\newblock {\em arXiv preprint arXiv:2305.10415}, 2023.

\bibitem{sun2024pathasst}
Yuxuan Sun, Chenglu Zhu, Sunyi Zheng, Kai Zhang, Lin Sun, Zhongyi Shui, Yunlong Zhang, Honglin Li, and Lin Yang.
\newblock Pathasst: A generative foundation ai assistant towards artificial general intelligence of pathology.
\newblock In {\em Proceedings of the AAAI Conference on Artificial Intelligence}, volume~38, pages 5034--5042, 2024.

\bibitem{thawkar2023xraygpt}
Omkar~Chakradhar Thawakar, Abdelrahman~M. Shaker, Sahal~Shaji Mullappilly, Hisham Cholakkal, Rao~Muhammad Anwer, Salman Khan, Jorma Laaksonen, and Fahad Khan.
\newblock {X}ray{GPT}: Chest radiographs summarization using large medical vision-language models.
\newblock In {\em Proceedings of the 23rd Workshop on Biomedical Natural Language Processing}, pages 440--448. Association for Computational Linguistics, August 2024.

\bibitem{liu2023qilin}
Junling Liu, Ziming Wang, Qichen Ye, Dading Chong, Peilin Zhou, and Yining Hua.
\newblock Qilin-med-vl: Towards chinese large vision-language model for general healthcare.
\newblock {\em arXiv preprint arXiv:2310.17956}, 2023.

\bibitem{he2024pefomed}
Jinlong He, Pengfei Li, Gang Liu, Zixu Zhao, and Shenjun Zhong.
\newblock Pefomed: Parameter efficient fine-tuning on multimodal large language models for medical visual question answering.
\newblock {\em arXiv preprint arXiv:2401.02797}, 2024.

\bibitem{bai2024m3d}
Fan Bai, Yuxin Du, Tiejun Huang, Max Q-H Meng, and Bo~Zhao.
\newblock M3d: Advancing 3d medical image analysis with multi-modal large language models.
\newblock {\em arXiv preprint arXiv:2404.00578}, 2024.

\bibitem{jiang2024moe}
Songtao Jiang, Tuo Zheng, Yan Zhang, Yeying Jin, and Zuozhu Liu.
\newblock Moe-tinymed: Mixture of experts for tiny medical large vision-language models.
\newblock {\em CoRR}, 2024.

\bibitem{bannur2024maira}
Shruthi Bannur, Kenza Bouzid, Daniel~C Castro, Anton Schwaighofer, Sam Bond-Taylor, Maximilian Ilse, Fernando P{\'e}rez-Garc{\'\i}a, Valentina Salvatelli, Harshita Sharma, Felix Meissen, et~al.
\newblock Maira-2: Grounded radiology report generation.
\newblock {\em arXiv preprint arXiv:2406.04449}, 2024.

\bibitem{chen2024huatuogpt}
Junying Chen, Chi Gui, Ruyi Ouyang, Anningzhe Gao, Shunian Chen, Guiming~Hardy Chen, Xidong Wang, Zhenyang Cai, Ke~Ji, Xiang Wan, and Benyou Wang.
\newblock Towards injecting medical visual knowledge into multimodal {LLM}s at scale.
\newblock In {\em Proceedings of the 2024 Conference on Empirical Methods in Natural Language Processing}, pages 7346--7370. Association for Computational Linguistics, November 2024.

\bibitem{alkhaldi2024minigpt}
Asma Alkhaldi, Raneem Alnajim, Layan Alabdullatef, Rawan Alyahya, Jun Chen, Deyao Zhu, Ahmed Alsinan, and Mohamed Elhoseiny.
\newblock Minigpt-med: Large language model as a general interface for radiology diagnosis.
\newblock {\em arXiv preprint arXiv:2407.04106}, 2024.

\bibitem{zhou2024pre}
Juexiao Zhou, Xiaonan He, Liyuan Sun, Jiannan Xu, Xiuying Chen, Yuetan Chu, Longxi Zhou, Xingyu Liao, Bin Zhang, Shawn Afvari, et~al.
\newblock Pre-trained multimodal large language model enhances dermatological diagnosis using skingpt-4.
\newblock {\em Nature Communications}, 15(1):5649, 2024.

\bibitem{zhou4988925sigphi}
Feizhong Zhou, Xingyue Liu, Qiao Zeng, Zhuhan Li, and Hanguang Xiao.
\newblock Sigphi-med: A lightweight vision-language assistant for biomedicine.
\newblock {\em Available at SSRN 4988925}, 2024.

\bibitem{zhao2023chatcad+}
Zihao Zhao, Sheng Wang, Jinchen Gu, Yitao Zhu, Lanzhuju Mei, Zixu Zhuang, Zhiming Cui, Qian Wang, and Dinggang Shen.
\newblock Chatcad+: Toward a universal and reliable interactive cad using llms.
\newblock {\em IEEE Transactions on Medical Imaging}, 43(11):3755--3766, 2024.

\bibitem{gao2023ophglm}
Weihao Gao, Zhuo Deng, Zhiyuan Niu, Fuju Rong, Chucheng Chen, Zheng Gong, Wenze Zhang, Daimin Xiao, Fang Li, Zhenjie Cao, et~al.
\newblock Ophglm: Training an ophthalmology large language-and-vision assistant based on instructions and dialogue.
\newblock {\em arXiv preprint arXiv:2306.12174}, 2023.

\bibitem{jaegle2021perceiver}
Andrew Jaegle, Felix Gimeno, Andy Brock, Oriol Vinyals, Andrew Zisserman, and Joao Carreira.
\newblock Perceiver: General perception with iterative attention.
\newblock In {\em Proceedings of the 38th International Conference on Machine Learning}, volume 139, pages 4651--4664. PMLR, 18--24 Jul 2021.

\bibitem{song2023bridge}
Shezheng Song, Xiaopeng Li, and Shasha Li.
\newblock How to bridge the gap between modalities: A comprehensive survey on multimodal large language model.
\newblock {\em arXiv preprint arXiv:2311.07594}, 2023.

\bibitem{li2023blip}
Junnan Li, Dongxu Li, Silvio Savarese, and Steven Hoi.
\newblock {BLIP}-2: Bootstrapping language-image pre-training with frozen image encoders and large language models.
\newblock In {\em Proceedings of the 40th International Conference on Machine Learning}, volume 202 of {\em Proceedings of Machine Learning Research}, pages 19730--19742. PMLR, 23--29 Jul 2023.

\bibitem{yao2024deco}
Linli Yao, Lei Li, Shuhuai Ren, Lean Wang, Yuanxin Liu, Xu~Sun, and Lu~Hou.
\newblock Deco: Decoupling token compression from semantic abstraction in multimodal large language models.
\newblock {\em arXiv preprint arXiv:2405.20985}, 2024.

\bibitem{liu2023improved}
Haotian Liu, Chunyuan Li, Yuheng Li, and Yong~Jae Lee.
\newblock Improved baselines with visual instruction tuning.
\newblock In {\em 2024 IEEE/CVF Conference on Computer Vision and Pattern Recognition (CVPR)}, pages 26286--26296, 2024.

\bibitem{johnson2016mimic}
Alistair~EW Johnson, Tom~J Pollard, Lu~Shen, Li-wei~H Lehman, Mengling Feng, Mohammad Ghassemi, Benjamin Moody, Peter Szolovits, Leo Anthony~Celi, and Roger~G Mark.
\newblock Mimic-iii, a freely accessible critical care database.
\newblock {\em Scientific Data}, 3(1):1--9, 2016.

\bibitem{johnson2023mimic}
Alistair~EW Johnson, Lucas Bulgarelli, Lu~Shen, Alvin Gayles, Ayad Shammout, Steven Horng, Tom~J Pollard, Sicheng Hao, Benjamin Moody, Brian Gow, et~al.
\newblock Mimic-iv, a freely accessible electronic health record dataset.
\newblock {\em Scientific Data}, 10(1):1, 2023.

\bibitem{herrett2015data}
Emily Herrett, Arlene~M Gallagher, Krishnan Bhaskaran, Harriet Forbes, Rohini Mathur, Tjeerd Van~Staa, and Liam Smeeth.
\newblock Data resource profile: clinical practice research datalink (cprd).
\newblock {\em International Journal of Epidemiology}, 44(3):827--836, 2015.

\bibitem{wang2020cord}
Lucy~Lu Wang, Kyle Lo, Yoganand Chandrasekhar, Russell Reas, Jiangjiang Yang, Douglas Burdick, Darrin Eide, Kathryn Funk, Yannis Katsis, Rodney Kinney, et~al.
\newblock {CORD-19}: The {COVID-19} open research dataset.
\newblock In {\em Proceedings of the 1st Workshop on {NLP} for {COVID-19} at {ACL} 2020}. Association for Computational Linguistics, July 2020.

\bibitem{jin2019pubmedqa}
Qiao Jin, Bhuwan Dhingra, Zhengping Liu, William Cohen, and Xinghua Lu.
\newblock {P}ub{M}ed{QA}: A dataset for biomedical research question answering.
\newblock In {\em Proceedings of the 2019 Conference on Empirical Methods in Natural Language Processing and the 9th International Joint Conference on Natural Language Processing (EMNLP-IJCNLP)}, pages 2567--2577. Association for Computational Linguistics, November 2019.

\bibitem{pal2022medmcqa}
Ankit Pal, Logesh~Kumar Umapathi, and Malaikannan Sankarasubbu.
\newblock Medmcqa: A large-scale multi-subject multi-choice dataset for medical domain question answering.
\newblock In Gerardo Flores, George~H Chen, Tom Pollard, Joyce~C Ho, and Tristan Naumann, editors, {\em Proceedings of the Conference on Health, Inference, and Learning}, volume 174 of {\em Proceedings of Machine Learning Research}, pages 248--260. PMLR, 07--08 Apr 2022.

\bibitem{zhang2018multi}
Sheng Zhang, Xin Zhang, Hui Wang, Lixiang Guo, and Shanshan Liu.
\newblock Multi-scale attentive interaction networks for chinese medical question answer selection.
\newblock {\em IEEE Access}, 6:74061--74071, 2018.

\bibitem{ben2019question}
Asma Ben~Abacha and Dina Demner-Fushman.
\newblock A question-entailment approach to question answering.
\newblock {\em BMC Bioinformatics}, 20:1--23, 2019.

\bibitem{li2023huatuo}
Jianquan Li, Xidong Wang, Xiangbo Wu, Zhiyi Zhang, Xiaolong Xu, Jie Fu, Prayag Tiwari, Xiang Wan, and Benyou Wang.
\newblock Huatuo-26m, a large-scale chinese medical qa dataset.
\newblock {\em arXiv preprint arXiv:2305.01526}, 2023.

\bibitem{zeng2020meddialog}
Guangtao Zeng, Wenmian Yang, Zeqian Ju, Yue Yang, Sicheng Wang, Ruisi Zhang, Meng Zhou, Jiaqi Zeng, Xiangyu Dong, Ruoyu Zhang, et~al.
\newblock Meddialog: Large-scale medical dialogue datasets.
\newblock In {\em Proceedings of the 2020 Conference on Empirical Methods in Natural Language Processing}, pages 9241--9250, 2020.

\bibitem{bodenreider2004unified}
Olivier Bodenreider.
\newblock The unified medical language system (umls): integrating biomedical terminology.
\newblock {\em Nucleic Acids Research}, 32(suppl\_1):D267--D270, 2004.

\bibitem{byambasuren2019preliminary}
Odma Byambasuren, Yunfei Yang, Zhifang Sui, Damai Dai, Baobao Chang, Sujian Li, and Hongying Zan.
\newblock Preliminary study on the construction of chinese medical knowledge graph.
\newblock {\em Journal of Chinese Information Processing}, 33(10):1--9, 2019.

\bibitem{basaldella2020cometa}
Marco Basaldella, Fangyu Liu, Ehsan Shareghi, and Nigel Collier.
\newblock {COMETA}: A corpus for medical entity linking in the social media.
\newblock In Bonnie Webber, Trevor Cohn, Yulan He, and Yang Liu, editors, {\em Proceedings of the 2020 Conference on Empirical Methods in Natural Language Processing (EMNLP)}, pages 3122--3137. Association for Computational Linguistics, November 2020.

\bibitem{lau2018dataset}
Jason~J Lau, Soumya Gayen, Asma Ben~Abacha, and Dina Demner-Fushman.
\newblock A dataset of clinically generated visual questions and answers about radiology images.
\newblock {\em Scientific Data}, 5(1):1--10, 2018.

\bibitem{liu2021slake}
Bo~Liu, Li-Ming Zhan, Li~Xu, Lin Ma, Yan Yang, and Xiao-Ming Wu.
\newblock Slake: A semantically-labeled knowledge-enhanced dataset for medical visual question answering.
\newblock In {\em 2021 IEEE 18th International Symposium on Biomedical Imaging}, pages 1650--1654. IEEE, 2021.

\bibitem{he2020pathvqa}
Xuehai He, Yichen Zhang, Luntian Mou, Eric Xing, and Pengtao Xie.
\newblock Pathvqa: 30000+ questions for medical visual question answering.
\newblock {\em arXiv preprint arXiv:2003.10286}, 2020.

\bibitem{johnson2019mimic}
Alistair~EW Johnson, Tom~J Pollard, Seth~J Berkowitz, Nathaniel~R Greenbaum, Matthew~P Lungren, Chih-ying Deng, Roger~G Mark, and Steven Horng.
\newblock Mimic-cxr, a de-identified publicly available database of chest radiographs with free-text reports.
\newblock {\em Scientific Data}, 6(1):317, 2019.

\bibitem{demner2016preparing}
Dina Demner-Fushman, Marc~D Kohli, Marc~B Rosenman, Sonya~E Shooshan, Laritza Rodriguez, Sameer Antani, George~R Thoma, and Clement~J McDonald.
\newblock Preparing a collection of radiology examinations for distribution and retrieval.
\newblock {\em Journal of the American Medical Informatics Association}, 23(2):304--310, 2016.

\bibitem{irvin2019chexpert}
Jeremy Irvin, Pranav Rajpurkar, Michael Ko, Yifan Yu, Silviana Ciurea-Ilcus, Chris Chute, Henrik Marklund, Behzad Haghgoo, Robyn Ball, Katie Shpanskaya, et~al.
\newblock Chexpert: A large chest radiograph dataset with uncertainty labels and expert comparison.
\newblock In {\em Proceedings of the AAAI conference on artificial intelligence}, volume~33, pages 590--597, 2019.

\bibitem{pelka2018radiology}
Obioma Pelka, Sven Koitka, Johannes R{\"u}ckert, Felix Nensa, and Christoph~M. Friedrich.
\newblock Radiology objects in context (roco): A multimodal image dataset.
\newblock In {\em Intravascular Imaging and Computer Assisted Stenting and Large-Scale Annotation of Biomedical Data and Expert Label Synthesis}, pages 180--189. Springer International Publishing, 2018.

\bibitem{huang2023visual}
Zhi Huang, Federico Bianchi, Mert Yuksekgonul, Thomas~J Montine, and James Zou.
\newblock A visual--language foundation model for pathology image analysis using medical twitter.
\newblock {\em Nature Medicine}, 29(9):2307--2316, 2023.

\bibitem{subramanian2020medicat}
Sanjay Subramanian, Lucy~Lu Wang, Ben Bogin, Sachin Mehta, Madeleine van Zuylen, Sravanthi Parasa, Sameer Singh, Matt Gardner, and Hannaneh Hajishirzi.
\newblock {M}ed{IC}a{T}: A dataset of medical images, captions, and textual references.
\newblock In {\em Findings of the Association for Computational Linguistics: EMNLP 2020}, pages 2112--2120. Association for Computational Linguistics, November 2020.

\bibitem{lin2023pmc}
Weixiong Lin, Ziheng Zhao, Xiaoman Zhang, Chaoyi Wu, Ya~Zhang, Yanfeng Wang, and Weidi Xie.
\newblock Pmc-clip: Contrastive language-image pre-training using biomedical documents.
\newblock In {\em Medical Image Computing and Computer Assisted Intervention"}, pages 525--536. Springer, 2023.

\bibitem{zhang2023large}
Sheng Zhang, Yanbo Xu, Naoto Usuyama, Jaspreet Bagga, Robert Tinn, Sam Preston, Rajesh Rao, Mu~Wei, Naveen Valluri, Cliff Wong, et~al.
\newblock Large-scale domain-specific pretraining for biomedical vision-language processing.
\newblock {\em arXiv preprint arXiv:2303.00915}, 2(3):6, 2023.

\bibitem{tu2024towards_conversational}
Tao Tu, Anil Palepu, Mike Schaekermann, Khaled Saab, Jan Freyberg, Ryutaro Tanno, Amy Wang, Brenna Li, Mohamed Amin, Nenad Tomasev, et~al.
\newblock Towards conversational diagnostic ai.
\newblock {\em arXiv preprint arXiv:2401.05654}, 2024.

\bibitem{yang2022large}
Xi~Yang, Aokun Chen, Nima PourNejatian, Hoo~Chang Shin, Kaleb~E Smith, Christopher Parisien, Colin Compas, Cheryl Martin, Anthony~B Costa, Mona~G Flores, et~al.
\newblock A large language model for electronic health records.
\newblock {\em NPJ Digital Medicine}, 5(1):194, 2022.

\bibitem{tang2023does}
Ruixiang Tang, Xiaotian Han, Xiaoqian Jiang, and Xia Hu.
\newblock Does synthetic data generation of llms help clinical text mining?
\newblock {\em arXiv preprint arXiv:2303.04360}, 2023.

\bibitem{wu2024continual}
Tongtong Wu, Linhao Luo, Yuan-Fang Li, Shirui Pan, Thuy-Trang Vu, and Gholamreza Haffari.
\newblock Continual learning for large language models: A survey.
\newblock {\em arXiv preprint arXiv:2402.01364}, 2024.

\bibitem{wei2021finetuned}
Jason Wei, Maarten Bosma, Vincent Zhao, Kelvin Guu, Adams~Wei Yu, Brian Lester, Nan Du, Andrew~M. Dai, and Quoc~V Le.
\newblock Finetuned language models are zero-shot learners.
\newblock In {\em International Conference on Learning Representations}, 2022.

\bibitem{casper2023open}
Stephen Casper, Xander Davies, Claudia Shi, Thomas~Krendl Gilbert, J{\'e}r{\'e}my Scheurer, Javier Rando, Rachel Freedman, Tomasz Korbak, David Lindner, Pedro Freire, et~al.
\newblock Open problems and fundamental limitations of reinforcement learning from human feedback.
\newblock {\em Transactions on Machine Learning Research}, 2023.
\newblock Survey Certification.

\bibitem{stiennon2020learning}
Nisan Stiennon, Long Ouyang, Jeffrey Wu, Daniel Ziegler, Ryan Lowe, Chelsea Voss, Alec Radford, Dario Amodei, and Paul~F Christiano.
\newblock Learning to summarize with human feedback.
\newblock {\em Advances in Neural Information Processing Systems}, 33:3008--3021, 2020.

\bibitem{bai2022constitutional}
Yuntao Bai, Saurav Kadavath, Sandipan Kundu, Amanda Askell, Jackson Kernion, Andy Jones, Anna Chen, Anna Goldie, Azalia Mirhoseini, Cameron McKinnon, et~al.
\newblock Constitutional ai: Harmlessness from ai feedback.
\newblock {\em arXiv preprint arXiv:2212.08073}, 2022.

\bibitem{rafailov2024direct}
Rafael Rafailov, Archit Sharma, Eric Mitchell, Christopher~D Manning, Stefano Ermon, and Chelsea Finn.
\newblock Direct preference optimization: Your language model is secretly a reward model.
\newblock {\em Advances in Neural Information Processing Systems}, 36, 2024.

\bibitem{deshpande2023toxicity}
Ameet Deshpande, Vishvak Murahari, Tanmay Rajpurohit, Ashwin Kalyan, and Karthik Narasimhan.
\newblock Toxicity in chatgpt: Analyzing persona-assigned language models.
\newblock In {\em Findings of the Association for Computational Linguistics: EMNLP 2023}, pages 1236--1270. Association for Computational Linguistics, December 2023.

\bibitem{papineni2002bleu}
Kishore Papineni, Salim Roukos, Todd Ward, and Wei-Jing Zhu.
\newblock Bleu: a method for automatic evaluation of machine translation.
\newblock In {\em Proceedings of the 40th Annual Meeting on Association for Computational Linguistics}, page 311–318. Association for Computational Linguistics, 2002.

\bibitem{lin2004rouge}
Chin-Yew Lin.
\newblock Rouge: A package for automatic evaluation of summaries.
\newblock In {\em Text Summarization Branches Out}, pages 74--81. Association for Computational Linguistics, 2004.

\bibitem{wu2016google}
Yonghui Wu, Mike Schuster, Zhifeng Chen, Quoc~V Le, Mohammad Norouzi, Wolfgang Macherey, Maxim Krikun, Yuan Cao, Qin Gao, Klaus Macherey, et~al.
\newblock Google's neural machine translation system: Bridging the gap between human and machine translation.
\newblock {\em arXiv preprint arXiv:1609.08144}, 2016.

\bibitem{li2015diversity}
Jiwei Li, Michel Galley, Chris Brockett, Jianfeng Gao, and Bill Dolan.
\newblock A diversity-promoting objective function for neural conversation models.
\newblock In {\em Proceedings of the 2016 Conference of the North {A}merican Chapter of the Association for Computational Linguistics: Human Language Technologies}, pages 110--119. Association for Computational Linguistics, 2016.

\bibitem{vedantam2015cider}
Ramakrishna Vedantam, C.~Lawrence Zitnick, and Devi Parikh.
\newblock Cider: Consensus-based image description evaluation.
\newblock In {\em 2015 IEEE Conference on Computer Vision and Pattern Recognition}, pages 4566--4575, 2015.

\bibitem{zhang2019bertscore}
Tianyi Zhang, Varsha Kishore, Felix Wu, Kilian~Q Weinberger, and Yoav Artzi.
\newblock Bertscore: Evaluating text generation with bert.
\newblock In {\em International Conference on Learning Representations}, 2020.

\bibitem{wang2023chatgpt}
Jiaan Wang, Yunlong Liang, Fandong Meng, Zengkui Sun, Haoxiang Shi, Zhixu Li, Jinan Xu, Jianfeng Qu, and Jie Zhou.
\newblock Is {C}hat{GPT} a good {NLG} evaluator? a preliminary study.
\newblock In {\em Proceedings of the 4th New Frontiers in Summarization Workshop}, pages 1--11. Association for Computational Linguistics, December 2023.

\bibitem{zheng2024judging}
Lianmin Zheng, Wei-Lin Chiang, Ying Sheng, Siyuan Zhuang, Zhanghao Wu, Yonghao Zhuang, Zi~Lin, Zhuohan Li, Dacheng Li, Eric~P. Xing, Hao Zhang, Joseph~E. Gonzalez, and Ion Stoica.
\newblock Judging llm-as-a-judge with mt-bench and chatbot arena.
\newblock In {\em Advances in Neural Information Processing Systems}, 2024.

\bibitem{szolovits1988artificial}
Peter Szolovits, Ramesh~S Patil, and William~B Schwartz.
\newblock Artificial intelligence in medical diagnosis.
\newblock {\em Annals of Internal Medicine}, 108(1):80--87, 1988.

\bibitem{yuan2023advanced}
J~Yuan, P~Bao, Z~Chen, M~Yuan, J~Zhao, J~Pan, Y~Xie, Y~Cao, Y~Wang, Z~Wang, et~al.
\newblock Advanced prompting as a catalyst: Empowering large language models in the management of gastrointestinal cancers.
\newblock {\em The Innovation}, 521, 2023.

\bibitem{zhu2023can}
Lingxuan Zhu, Weiming Mou, and Rui Chen.
\newblock Can the chatgpt and other large language models with internet-connected database solve the questions and concerns of patient with prostate cancer and help democratize medical knowledge?
\newblock {\em Journal of Translational Medicine}, 21(1):269, 2023.

\bibitem{liu2023medical}
Fenglin Liu, Tingting Zhu, Xian Wu, Bang Yang, Chenyu You, Chenyang Wang, Lei Lu, Zhangdaihong Liu, Yefeng Zheng, Xu~Sun, et~al.
\newblock A medical multimodal large language model for future pandemics.
\newblock {\em NPJ Digital Medicine}, 6(1):226, 2023.

\bibitem{lee2023benefits}
Peter Lee, Sebastien Bubeck, and Joseph Petro.
\newblock Benefits, limits, and risks of gpt-4 as an ai chatbot for medicine.
\newblock {\em New England Journal of Medicine}, 388(13):1233--1239, 2023.

\bibitem{ali2023using}
Stephen~R Ali, Thomas~D Dobbs, Hayley~A Hutchings, and Iain~S Whitaker.
\newblock Using chatgpt to write patient clinic letters.
\newblock {\em The Lancet Digital Health}, 5(4):e179--e181, 2023.

\bibitem{patel2023chatgpt}
Sajan~B Patel and Kyle Lam.
\newblock Chatgpt: the future of discharge summaries?
\newblock {\em The Lancet Digital Health}, 5(3):e107--e108, 2023.

\bibitem{clough2024transforming}
Reece Alexander~James Clough, William~Anthony Sparkes, Oliver~Thomas Clough, Joshua~Thomas Sykes, Alexander~Thomas Steventon, and Kate King.
\newblock Transforming healthcare documentation: harnessing the potential of ai to generate discharge summaries.
\newblock {\em BJGP Open}, 2024.

\bibitem{nori2023capabilities}
Harsha Nori, Nicholas King, Scott~Mayer McKinney, Dean Carignan, and Eric Horvitz.
\newblock Capabilities of gpt-4 on medical challenge problems.
\newblock {\em arXiv preprint arXiv:2303.13375}, 2023.

\bibitem{yang2023dawn}
Zhengyuan Yang, Linjie Li, Kevin Lin, Jianfeng Wang, Chung-Ching Lin, Zicheng Liu, and Lijuan Wang.
\newblock The dawn of lmms: Preliminary explorations with gpt-4v (ision).
\newblock {\em arXiv preprint arXiv:2309.17421}, 9(1):1, 2023.

\bibitem{yang2023performance}
Zhichao Yang, Zonghai Yao, Mahbuba Tasmin, Parth Vashisht, Won~Seok Jang, Feiyun Ouyang, Beining Wang, Dan Berlowitz, and Hong Yu.
\newblock Performance of multimodal gpt-4v on usmle with image: Potential for imaging diagnostic support with explanations.
\newblock {\em medRxiv}, pages 2023--10, 2023.

\bibitem{khan2023harnessing}
Sal Khan.
\newblock Harnessing gpt-4 so that all students benefit. a nonprofit approach for equal access.
\newblock {\em Khan Academy Blog}, 2023.

\bibitem{team2023introducing}
Duolingo Team.
\newblock Introducing duolingo max, a learning experience powered by gpt-4.
\newblock {\em Retrieved March}, 15:2023, 2023.

\bibitem{karabacak2023advent}
Mert Karabacak, Burak~Berksu Ozkara, Konstantinos Margetis, Max Wintermark, and Sotirios Bisdas.
\newblock The advent of generative language models in medical education.
\newblock {\em JMIR Medical Education}, 9:e48163, 2023.

\bibitem{lee2024rise}
Hyunsu Lee.
\newblock The rise of chatgpt: Exploring its potential in medical education.
\newblock {\em Anatomical sciences education}, 17(5):926--931, 2024.

\bibitem{lyu2023translating}
Qing Lyu, Josh Tan, Michael~E Zapadka, Janardhana Ponnatapura, Chuang Niu, Kyle~J Myers, Ge~Wang, and Christopher~T Whitlow.
\newblock Translating radiology reports into plain language using chatgpt and gpt-4 with prompt learning: results, limitations, and potential.
\newblock {\em Visual Computing for Industry, Biomedicine, and Art}, 6(1):9, 2023.

\bibitem{han2023explorative}
Zhiyong Han, Fortunato Battaglia, Abinav Udaiyar, Allen Fooks, and Stanley~R Terlecky.
\newblock An explorative assessment of chatgpt as an aid in medical education: Use it with caution.
\newblock {\em Medical Teacher}, pages 1--8, 2023.

\bibitem{ahn2023impending}
Sangzin Ahn.
\newblock The impending impacts of large language models on medical education.
\newblock {\em Korean Journal of Medical Education}, 35(1):103, 2023.

\bibitem{van2023global}
Alastair~C van Heerden, Julia~R Pozuelo, and Brandon~A Kohrt.
\newblock Global mental health services and the impact of artificial intelligence--powered large language models.
\newblock {\em JAMA Psychiatry}, 80(7):662--664, 2023.

\bibitem{de2023benefits}
Munmun De~Choudhury, Sachin~R Pendse, and Neha Kumar.
\newblock Benefits and harms of large language models in digital mental health.
\newblock {\em arXiv preprint arXiv:2311.14693}, 2023.

\bibitem{stock2023tell}
Anna Stock, Stephan Schl{\"o}gl, and Aleksander Groth.
\newblock Tell me, what are you most afraid of? exploring the effects of agent representation on information disclosure in human-chatbot interaction.
\newblock In {\em International Conference on Human-Computer Interaction}, pages 179--191. Springer, 2023.

\bibitem{chaves2021should}
Ana~Paula Chaves and Marco~Aurelio Gerosa.
\newblock How should my chatbot interact? a survey on social characteristics in human--chatbot interaction design.
\newblock {\em International Journal of Human--Computer Interaction}, 37(8):729--758, 2021.

\bibitem{BAI2025102602}
Long Bai, Guankun Wang, Mobarakol Islam, Lalithkumar Seenivasan, An~Wang, and Hongliang Ren.
\newblock Surgical-vqla++: Adversarial contrastive learning for calibrated robust visual question-localized answering in robotic surgery.
\newblock {\em Information Fusion}, 113:102602, 2025.

\bibitem{barua2024innovations}
Ranjit Barua.
\newblock Innovations in minimally invasive surgery: The rise of smart flexible surgical robots.
\newblock In {\em Emerging Technologies for Health Literacy and Medical Practice}, pages 110--131. IGI Global, 2024.

\bibitem{seenivasan2023surgicalgpt}
Lalithkumar Seenivasan, Mobarakol Islam, Gokul Kannan, and Hongliang Ren.
\newblock Surgicalgpt: End-to-end language-vision gpt for visual question answering in surgery.
\newblock In {\em International Conference on Medical Image Computing and Computer-Assisted Intervention}, pages 281--290. Springer, 2023.

\bibitem{cao2023instruction}
Yihan Cao, Yanbin Kang, Chi Wang, and Lichao Sun.
\newblock Instruction mining: Instruction data selection for tuning large language models.
\newblock In {\em First Conference on Language Modeling}, 2024.

\bibitem{chen2023alpagasus}
Lichang Chen, Shiyang Li, Jun Yan, Hai Wang, Kalpa Gunaratna, Vikas Yadav, Zheng Tang, Vijay Srinivasan, Tianyi Zhou, Heng Huang, and Hongxia Jin.
\newblock Alpagasus: Training a better alpaca with fewer data.
\newblock In {\em The Twelfth International Conference on Learning Representations}, 2024.

\bibitem{dhuliawala2023chain}
Shehzaad Dhuliawala, Mojtaba Komeili, Jing Xu, Roberta Raileanu, Xian Li, Asli Celikyilmaz, and Jason Weston.
\newblock Chain-of-verification reduces hallucination in large language models.
\newblock In {\em Findings of the Association for Computational Linguistics}, pages 3563--3578. Association for Computational Linguistics, August 2024.

\bibitem{shuster2021retrieval}
Kurt Shuster, Spencer Poff, Moya Chen, Douwe Kiela, and Jason Weston.
\newblock Retrieval augmentation reduces hallucination in conversation.
\newblock In {\em Findings of the Association for Computational Linguistics: EMNLP 2021}, pages 3784--3803. Association for Computational Linguistics, November 2021.

\bibitem{oquab2023dinov2}
Maxime Oquab, Timoth{\'e}e Darcet, Th{\'e}o Moutakanni, Huy Vo, Marc Szafraniec, Vasil Khalidov, Pierre Fernandez, Daniel Haziza, Francisco Massa, Alaaeldin El-Nouby, et~al.
\newblock {DINO}v2: Learning robust visual features without supervision.
\newblock {\em Transactions on Machine Learning Research}, 2024.

\bibitem{kirillov2023segment}
Alexander Kirillov, Eric Mintun, Nikhila Ravi, Hanzi Mao, Chloe Rolland, Laura Gustafson, Tete Xiao, Spencer Whitehead, Alexander~C Berg, Wan-Yen Lo, et~al.
\newblock Segment anything.
\newblock In {\em Proceedings of the IEEE/CVF International Conference on Computer Vision}, pages 4015--4026, 2023.

\bibitem{zong2024mova}
Zhuofan Zong, Bingqi Ma, Dazhong Shen, Guanglu Song, Hao Shao, Dongzhi Jiang, Hongsheng Li, and Yu~Liu.
\newblock Mo{VA}: Adapting mixture of vision experts to multimodal context.
\newblock In {\em The Thirty-eighth Annual Conference on Neural Information Processing Systems}, 2024.

\bibitem{li2021prefix}
Xiang~Lisa Li and Percy Liang.
\newblock Prefix-tuning: Optimizing continuous prompts for generation.
\newblock In {\em Proceedings of the 59th Annual Meeting of the Association for Computational Linguistics and the 11th International Joint Conference on Natural Language Processing}, pages 4582--4597. Association for Computational Linguistics, August 2021.

\bibitem{hu2021lora}
Edward~J Hu, yelong shen, Phillip Wallis, Zeyuan Allen-Zhu, Yuanzhi Li, Shean Wang, Lu~Wang, and Weizhu Chen.
\newblock Lo{RA}: Low-rank adaptation of large language models.
\newblock In {\em International Conference on Learning Representations}, 2022.

\bibitem{houlsby2019parameter}
Neil Houlsby, Andrei Giurgiu, Stanislaw Jastrzebski, Bruna Morrone, Quentin De~Laroussilhe, Andrea Gesmundo, Mona Attariyan, and Sylvain Gelly.
\newblock Parameter-efficient transfer learning for nlp.
\newblock In {\em International Conference on Machine Learning}, pages 2790--2799. PMLR, 2019.

\bibitem{chu2023mobilevlm}
Xiangxiang Chu, Limeng Qiao, Xinyang Lin, Shuang Xu, Yang Yang, Yiming Hu, Fei Wei, Xinyu Zhang, Bo~Zhang, Xiaolin Wei, et~al.
\newblock Mobilevlm: A fast, reproducible and strong vision language assistant for mobile devices.
\newblock {\em arXiv preprint arXiv:2312.16886}, 2023.

\bibitem{yuan2023tinygpt}
Zhengqing Yuan, Zhaoxu Li, Weiran Huang, Yanfang Ye, and Lichao Sun.
\newblock Tiny{GPT}-v: Efficient multimodal large language model via small backbones.
\newblock In {\em 2nd Workshop on Advancing Neural Network Training: Computational Efficiency, Scalability, and Resource Optimization}, 2024.

\bibitem{peng2023rwkv}
Bo~Peng, Eric Alcaide, Quentin Anthony, Alon Albalak, Samuel Arcadinho, Huanqi Cao, Xin Cheng, Michael Chung, Matteo Grella, Kranthi~Kiran GV, et~al.
\newblock {RWKV}: Reinventing {RNN}s for the transformer era.
\newblock In {\em Findings of the Association for Computational Linguistics: EMNLP 2023}, pages 14048--14077. Association for Computational Linguistics, December 2023.

\bibitem{gu2023mamba}
Albert Gu and Tri Dao.
\newblock Mamba: Linear-time sequence modeling with selective state spaces.
\newblock In {\em First Conference on Language Modeling}, 2024.

\bibitem{zhai2023investigating}
Yuexiang Zhai, Shengbang Tong, Xiao Li, Mu~Cai, Qing Qu, Yong~Jae Lee, and Yi~Ma.
\newblock Investigating the catastrophic forgetting in multimodal large language model fine-tuning.
\newblock In {\em Conference on Parsimony and Learning}, 2023.

\bibitem{zheng2024beyond}
Junhao Zheng, Qianli Ma, Zhen Liu, Binquan Wu, and Huawen Feng.
\newblock Beyond anti-forgetting: Multimodal continual instruction tuning with positive forward transfer.
\newblock {\em arXiv preprint arXiv:2401.09181}, 2024.

\bibitem{yao2023editing}
Yunzhi Yao, Peng Wang, Bozhong Tian, Siyuan Cheng, Zhoubo Li, Shumin Deng, Huajun Chen, and Ningyu Zhang.
\newblock Editing large language models: Problems, methods, and opportunities.
\newblock In {\em Proceedings of the 2023 Conference on Empirical Methods in Natural Language Processing}, pages 10222--10240. Association for Computational Linguistics, December 2023.

\bibitem{huang2023transformer}
Zeyu Huang, Yikang Shen, Xiaofeng Zhang, Jie Zhou, Wenge Rong, and Zhang Xiong.
\newblock Transformer-patcher: One mistake worth one neuron.
\newblock {\em arXiv preprint arXiv:2301.09785}, 2023.

\bibitem{hartvigsen2024aging}
Tom Hartvigsen, Swami Sankaranarayanan, Hamid Palangi, Yoon Kim, and Marzyeh Ghassemi.
\newblock Aging with grace: Lifelong model editing with discrete key-value adaptors.
\newblock In {\em Advances in Neural Information Processing Systems}, volume~36, pages 47934--47959. Curran Associates, Inc., 2023.

\bibitem{meng2022locating}
Kevin Meng, David Bau, Alex Andonian, and Yonatan Belinkov.
\newblock Locating and editing factual associations in gpt.
\newblock {\em Advances in Neural Information Processing Systems}, 35:17359--17372, 2022.

\bibitem{meng2022mass}
Kevin Meng, Arnab~Sen Sharma, Alex~J Andonian, Yonatan Belinkov, and David Bau.
\newblock Mass-editing memory in a transformer.
\newblock In {\em The Eleventh International Conference on Learning Representations}, 2023.

\bibitem{li2024pmet}
Xiaopeng Li, Shasha Li, Shezheng Song, Jing Yang, Jun Ma, and Jie Yu.
\newblock Pmet: Precise model editing in a transformer.
\newblock In {\em Proceedings of the AAAI Conference on Artificial Intelligence}, volume~38, pages 18564--18572, 2024.

\bibitem{carlini2021extracting}
Nicholas Carlini, Florian Tramer, Eric Wallace, Matthew Jagielski, Ariel Herbert-Voss, Katherine Lee, Adam Roberts, Tom Brown, Dawn Song, Ulfar Erlingsson, et~al.
\newblock Extracting training data from large language models.
\newblock In {\em 30th USENIX Security Symposium (USENIX Security 21)}, pages 2633--2650, 2021.

\bibitem{li2023multi}
Haoran Li, Dadi Guo, Wei Fan, Mingshi Xu, Jie Huang, Fanpu Meng, and Yangqiu Song.
\newblock Multi-step jailbreaking privacy attacks on {C}hat{GPT}.
\newblock In {\em Findings of the Association for Computational Linguistics: EMNLP 2023}, pages 4138--4153, December 2023.

\bibitem{turgay2023perturbation}
Safiye Turgay, {\.I}lker {\.I}lter, et~al.
\newblock Perturbation methods for protecting data privacy: A review of techniques and applications.
\newblock {\em Automation and Machine Learning}, 4(2):31--41, 2023.

\bibitem{ferrara2023should}
Emilio Ferrara.
\newblock Should chatgpt be biased? challenges and risks of bias in large language models.
\newblock {\em arXiv preprint arXiv:2304.03738}, 2023.

\bibitem{yang2024unmasking}
Yifan Yang, Xiaoyu Liu, Qiao Jin, Furong Huang, and Zhiyong Lu.
\newblock Unmasking and quantifying racial bias of large language models in medical report generation.
\newblock {\em Nature Medicine}, 4:176, 2024.

\bibitem{kotek2023gender}
Hadas Kotek, Rikker Dockum, and David Sun.
\newblock Gender bias and stereotypes in large language models.
\newblock In {\em Proceedings of The ACM Collective Intelligence Conference}, pages 12--24, 2023.

\bibitem{liu2022quantifying}
Ruibo Liu, Chenyan Jia, Jason Wei, Guangxuan Xu, and Soroush Vosoughi.
\newblock Quantifying and alleviating political bias in language models.
\newblock {\em Artificial Intelligence}, 304:103654, 2022.

\bibitem{lahnala2022mitigating}
Allison Lahnala, Charles Welch, B{\'e}la Neuendorf, and Lucie Flek.
\newblock Mitigating toxic degeneration with empathetic data: Exploring the relationship between toxicity and empathy.
\newblock In {\em Proceedings of the 2022 Conference of the North American Chapter of the Association for Computational Linguistics: Human Language Technologies}, pages 4926--4938. Association for Computational Linguistics, July 2022.

\bibitem{xu2024unleashing}
Minrui Xu, Hongyang Du, Dusit Niyato, Jiawen Kang, Zehui Xiong, Shiwen Mao, Zhu Han, Abbas Jamalipour, Dong~In Kim, Xuemin Shen, Victor C.~M. Leung, and H.~Vincent Poor.
\newblock Unleashing the power of edge-cloud generative ai in mobile networks: A survey of aigc services.
\newblock {\em IEEE Communications Surveys \& Tutorials}, 26(2):1127--1170, 2024.

\bibitem{lin2023pushing}
Zheng Lin, Guanqiao Qu, Qiyuan Chen, Xianhao Chen, Zhe Chen, and Kaibin Huang.
\newblock Pushing large language models to the 6g edge: Vision, challenges, and opportunities.
\newblock {\em arXiv preprint arXiv:2309.16739}, 2023.

\bibitem{kim2024adaptive}
Yubin Kim, Chanwoo Park, Hyewon Jeong, Yik~Siu Chan, Xuhai Xu, Daniel McDuff, Cynthia Breazeal, and Hae~Won Park.
\newblock Adaptive collaboration strategy for llms in medical decision making.
\newblock {\em arXiv preprint arXiv:2404.15155}, 2024.

\bibitem{tang2023medagents}
Xiangru Tang, Anni Zou, Zhuosheng Zhang, Ziming Li, Yilun Zhao, Xingyao Zhang, Arman Cohan, and Mark Gerstein.
\newblock {M}ed{A}gents: Large language models as collaborators for zero-shot medical reasoning.
\newblock In {\em Findings of the Association for Computational Linguistics: ACL 2024}, pages 599--621. Association for Computational Linguistics, August 2024.

\bibitem{chan2024medtsllm}
Nimeesha Chan, Felix Parker, William Bennett, Tianyi Wu, Mung~Yao Jia, James Fackler, and Kimia Ghobadi.
\newblock Medtsllm: Leveraging llms for multimodal medical time series analysis.
\newblock {\em arXiv preprint arXiv:2408.07773}, 2024.

\bibitem{hu2024parallel}
Min Hu, Lei Liu, Xiaohua Wang, Yiming Tang, Jiaoyun Yang, and Ning An.
\newblock Parallel multiscale bridge fusion network for audio–visual automatic depression assessment.
\newblock {\em IEEE Transactions on Computational Social Systems}, 11(5):6830--6842, 2024.

\bibitem{CHEN2024102017}
Jian Chen, Yuzhu Hu, Qifeng Lai, Wei Wang, Junxin Chen, Han Liu, Gautam Srivastava, Ali~Kashif Bashir, and Xiping Hu.
\newblock Iifdd: Intra and inter-modal fusion for depression detection with multi-modal information from internet of medical things.
\newblock {\em Information Fusion}, 102:102017, 2024.

\bibitem{HE202256}
Lang He, Mingyue Niu, Prayag Tiwari, Pekka Marttinen, Rui Su, Jiewei Jiang, Chenguang Guo, Hongyu Wang, Songtao Ding, Zhongmin Wang, Xiaoying Pan, and Wei Dang.
\newblock Deep learning for depression recognition with audiovisual cues: A review.
\newblock {\em Information Fusion}, 80:56--86, 2022.

\end{thebibliography}



\end{document}